\documentclass[10pt,final,a4paper,oneside,onecolumn]{article}

\edef\reptitle{Simulation Assessment Guidelines towards Independent Safety Assurance of Autonomous Vehicles}
\edef\shorttitle{Simulation Assessment Guidelines for AVs}
\def\repauthor{Jim Cherian\\Martin Slavik\\Andrea Piazzoni\\Roshan Vijay\\Mohamed Azhar\\Niels de Boer}

\makeatletter
\def\@maketitle{%
	\newpage
	\null
	\vspace{-8em}
	\begin{center}%
		\begin{tabular}{l c r}
			\multicolumn{3}{c}{\includegraphics[width=0.6\linewidth]{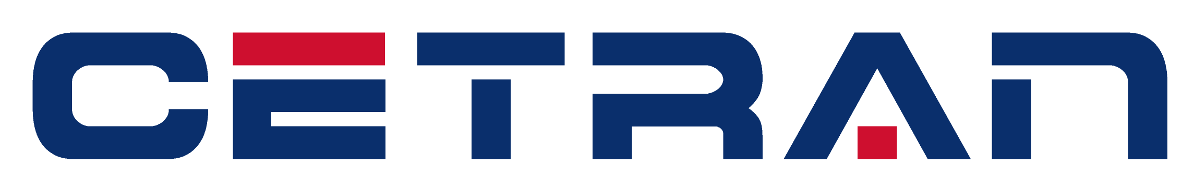}}\vspace{3em} \\
			\includegraphics[width=0.25\linewidth]{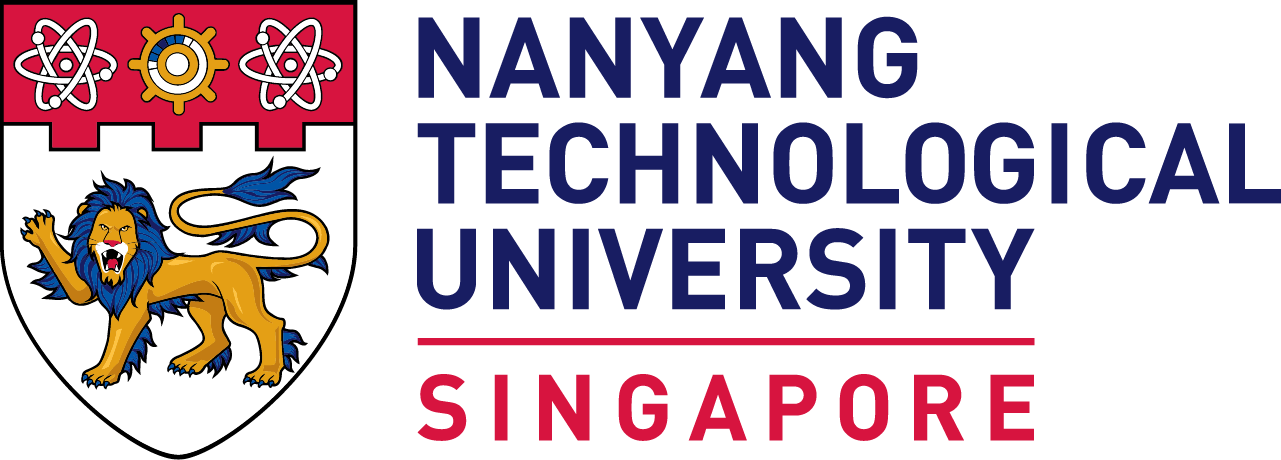} & \hspace{4em} &
			\includegraphics[width=0.3\linewidth]{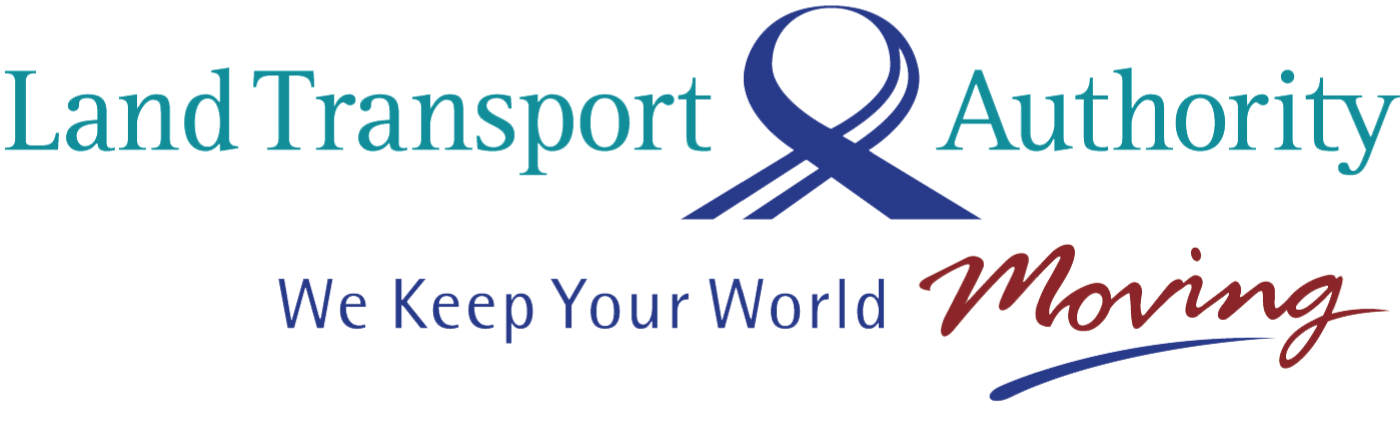}
		\end{tabular}\\
		\vskip 4.5em
		\let \footnote \thanks
		{\LARGE \@title \par}%
		\vskip 1em%
		{\large Version 1.0 from October 2, 2023}%
		\vskip 1.5em%
		{\large
			\lineskip .5em%
			\begin{tabular}[t]{c}%
				Jim Cherian\\
				Martin Slavik\\
				Andrea Piazzoni\\
				Roshan Vijay\\
				Mohamed Azhar\\
				Niels de Boer
			\end{tabular}\par}%
		\begin{tikzpicture}[remember picture,overlay]
		\node[anchor=south west,inner sep=0pt] at (current page.south west)
		{\includegraphics[width=\paperwidth]{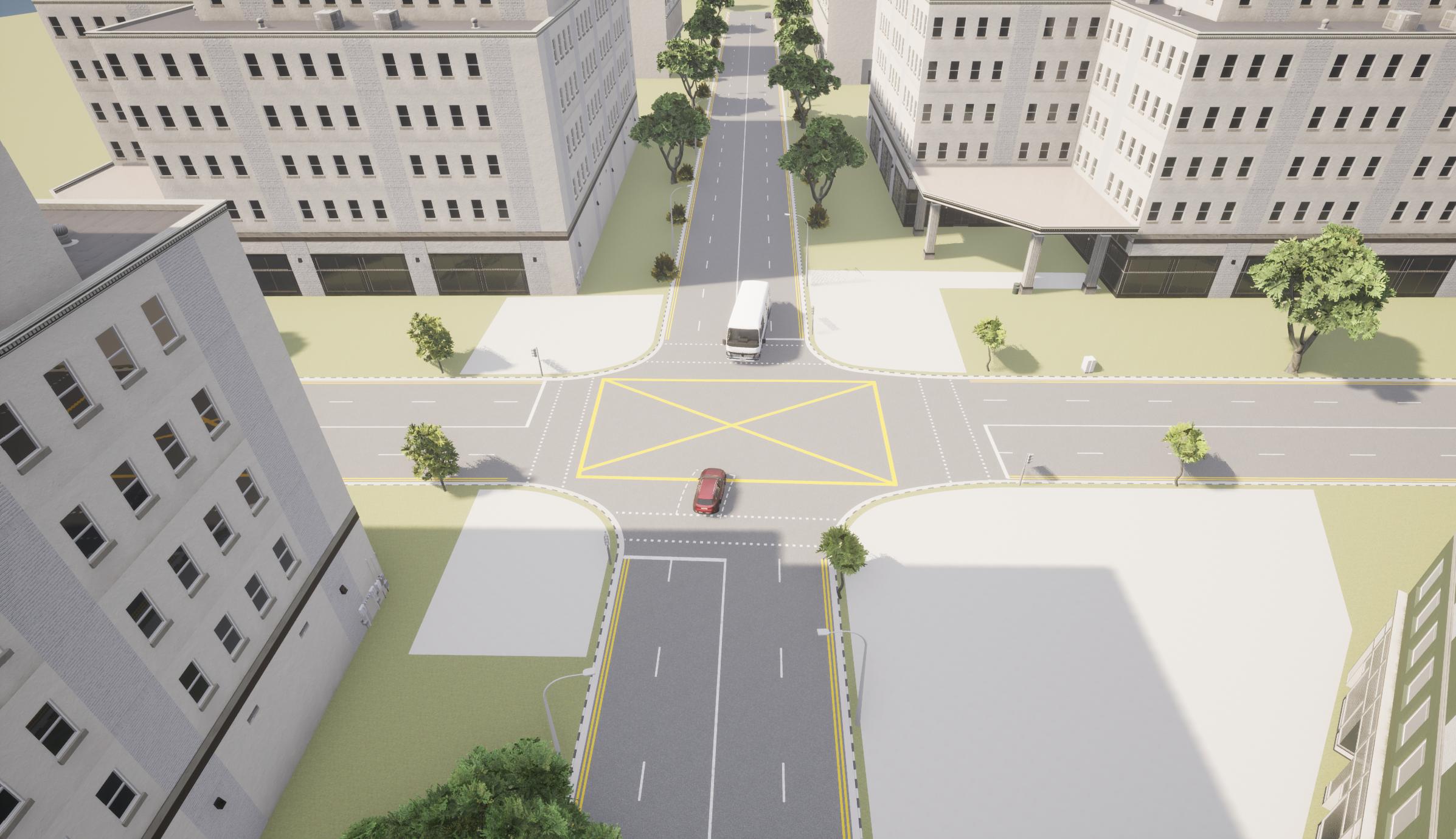}};
		\end{tikzpicture}
	\end{center}%
	\par
	\vskip 1.5em}
\makeatother

\usepackage[utf8]{inputenc}
\usepackage[a4paper,left=3cm,right=2.5cm,top=3cm,bottom=3cm]{geometry} 
\usepackage[T1]{fontenc}                        
\usepackage{lmodern}                            
\usepackage[cmex10]{amsmath}                    
\usepackage{amssymb}                            
\usepackage{amsthm}                             
\usepackage{dsfont}                             
\usepackage{mathrsfs}                           
\usepackage[pdftex]{graphicx}                   
\usepackage{array}                              
\usepackage{upgreek}                            
\usepackage[style=ieee,doi=false,isbn=false,url=false,date=year,backend=biber,mincitenames=2,maxcitenames=2,mincitenames=1,uniquelist=false,uniquename=false,giveninits=true,sorting=nty]{biblatex}
\usepackage{stfloats}                           
\usepackage{float}
\usepackage{multirow}                           
\usepackage{subfig}                             
\usepackage{fancyhdr}                           
\usepackage[official,left]{eurosym}             
\usepackage{appendix}                           
\usepackage{xspace}                             
\usepackage{microtype}							
\usepackage{verbatim}                           
\usepackage[dutch,USenglish]{babel}             
\usepackage{wrapfig}                            
\usepackage{longtable}                          
\usepackage{pgfplots}                           
\usepackage{calc}                               
\usepackage{lscape}								
\usepackage[breaklinks=true,hidelinks,          
            bookmarksnumbered=true]{hyperref}   
\usepackage{vhistory}
\usepackage{makecell}
\usepackage[toc]{glossaries}                    
\usepackage[normalem]{ulem}                     
\usepackage{booktabs}							
\usepackage[bottom]{footmisc}                   
\usepackage{csvsimple}  						
\usepackage{makecell}  						
\usepackage{longtable}
\usepackage{acronym}                            
\usepackage{titlesec}
\usepackage{setspace}
\usepackage{changepage}
\usepackage{xcolor}

\usepackage{enumitem}

\usepackage{refcount}

\usepackage{rotating} 

\fancypagestyle{plain}{\fancyhf{}

	\setlength{\headheight}{17pt}
	\addtolength{\footskip}{-5pt}}

\newlength\figurewidth
\newlength\figureheight
\pgfplotsset{every axis/.append style={
		scaled y ticks=false,
		scaled x ticks=false,
		y tick label style={/pgf/number format/fixed},
		x tick label style={/pgf/number format/fixed},
		legend style={font=\small}},
	compat=1.9}                                 
\usetikzlibrary{arrows.meta}                    
\usetikzlibrary{external}                       

\theoremstyle{plain}    
\theoremstyle{definition}     
\theoremstyle{remark}\newtheorem{remarkenv}{Remark}[section]        
                       {\hfill$\lozenge$\end{remarkenv}}            

\graphicspath{{figures/}} 
\fnbelowfloat                                       
\hypersetup{pdftitle={\reptitle},
            pdfauthor={\repauthor},
            colorlinks=true,
            urlcolor=blue,
        	linkcolor=black,
        	citecolor=black}	               

\newlength\ndist                                
\newlength\nheight                              
\newlength\nwidth                                   
\newlength\nsep                                     
\tikzstyle{block}=[node distance=\ndist,
                   text width=0.84*\ndist,
                   minimum height=\nheight,
                   align=center,
                   font=\footnotesize\sffamily,
                   rounded corners=0.4em]       
                   \tikzstyle{block2}=[node distance=\nwidth+\nsep,
                   text width=0.88\nwidth,
                   minimum height=\nheight,
                   minimum width=\nwidth,
                   align=center,
                   font=\footnotesize\sffamily,
                   fill=TNOyellow,
                   rounded corners=0.4em,
                   anchor=south west]
                   \tikzstyle{block3}=[node distance=\nwidth+\nsep,
                   text width=0.88\nwidth,
                   minimum height=\nheight,
                   minimum width=\nwidth,
                   align=center,
                   font=\footnotesize\sffamily,
                   fill=TNOorange,
                   rounded corners=0.4em,
                   anchor=south west]               
                   \tikzstyle{halfblock}=[node distance=\nwidth+\nsep,
                   text width=0.88\nwidth,
                   minimum height=0.5\nheight-0.5\nsep,
                   minimum width=\nwidth,
                   align=center,
                   font=\footnotesize\sffamily,
                   fill=TNOlightblue,
                   rounded corners=0.4em,
                   anchor=south west]               
                   \tikzstyle{doublehalfblock}=[node distance=2\nwidth+2\nsep,
                   text width=1.88\nwidth,
                   minimum height=0.5\nheight-0.5\nsep,
                   minimum width=2\nwidth+\nsep,
                   align=center,
                   font=\footnotesize\sffamily,
                   fill=TNOlightblue,
                   rounded corners=0.4em,
                   anchor=south west]               
               
\colorlet{pipeBlue}{blue!30}                    
\colorlet{pipeOrange}{orange!30}                
\colorlet{pipeGreen}{green!30}                  
\colorlet{pipeRed}{red!30}                      
\definecolor{TNOred}{RGB}{204,0,0}              
\definecolor{TNOgreen}{RGB}{51,153,51}          
\definecolor{TNOblue}{RGB}{102,153,204}         
\definecolor{TNOdarkblue}{RGB}{51,102,153}      
\definecolor{TNOorange}{RGB}{255,102,0}         
\definecolor{TNOyellow}{RGB}{255,204,51}        
\definecolor{TNOgray}{RGB}{166,166,166}         
\definecolor{TNOlightgray}{RGB}{222,222,231}    

\newcommand\disclaimertext{%
	\begin{center} Disclaimer \end{center} 
	\footnotesize
	\begin{itemize}
        \item This document and the information contained within are intended to assist the developers and testers of AVs in Singapore.
        \item This document has been reviewed by LTA and the information contained within is co-developed by CETRAN and LTA.
	\end{itemize}
}
\newcommand\disclaimernotice{%
	\begin{tikzpicture}[remember picture,overlay]
	\node[anchor=south,yshift=100pt] at (current page.south) {\fbox{\parbox{\dimexpr\textwidth-\fboxsep-\fboxrule\relax}{\disclaimertext}}};
	\end{tikzpicture}%
}

\newcommand{\tabitem}{~~\llap{\textbullet}~}

\bibliography{bibliography}

\begin{document}

\title{\Huge\textbf{\reptitle}}
\author{}

\maketitle
\vfill

\clearpage

\setlength\ndist{5em}   
\setlength\nheight{8em} 
\selectlanguage{USenglish}
\pagenumbering{roman}

\vfill

\clearpage

\tableofcontents
\disclaimernotice

\cleardoublepage

\pagenumbering{arabic}

\section{Introduction}

This document describes a major component of the overall \acs{cetran} Safety Assessment Framework for Autonomous Vehicles.

This document is primarily intended to help the developers of \acp{av} in Singapore to prepare their software simulations and provide recommendations that can ensure their readiness for independent assessment of their virtual simulation results according to the Milestone-testing framework adopted by the \textit{assessor} and the local \textit{authority} in Singapore, namely, \ac{cetran} and \ac{lta} respectively.

The main motivation behind the simulation assessment exercise is to ensure and gather sufficient confidence that:
\begin{itemize}
	\item the applicant has sufficient technical capability (or the access to such capability) towards implementing and achieving a meaningful \textit{virtual} testing of their \ac{av}, or at least their \ac{ads} in a \ac{sil} configuration, for \ac{vv} purposes.
	\item the applicant has implemented the \ac{vtt} and integrated their \ac{av} and/or \ac{ads} to it, in such a way that:
	\begin{itemize}
		\item the fidelity \cite{SISO_FISG_Report_1999} of the \ac{vtt} has been, is, and can be ascertained 
		\item and that the virtual testing results can be considered valid and representative of the real system as if it were operating in its real-world \ac{odd} under similar test conditions
	\end{itemize} 
	\item the applicant has established a mature internal \ac{vnv} process that involves meaningful \textit{virtual} testing of their \ac{av} or at least their \ac{ads} in a \ac{sil} configuration, that complements their physical testing and other \ac{vnv} methods, and as per documentary evidence produced during a review of quality processes documentation conducted prior to this simulation assessment
	\item the behavioral safety of the applicant's \ac{av} or at least their \ac{ads} in a \ac{sil} configuration, especially under critical driving situations or any extreme cases that are either practically difficult or too risky to be orchestrated and executed through physical testing methods, is sufficient enough that the \ac{av} can be allowed to drive on public roads; at least, with an onboard safety driver under \ac{m2} trial conditions or without any safety-related involvement from a safety operator under \ac{m3} trial conditions.	
\end{itemize}

In the rest of this document, we may use the following set of terms interchangeably with the same meaning.
\begin{itemize}
	\item \ac{vut} or \ac{av} or ego vehicle
	\item Simulation or Virtual testing or Simulation testing or Virtual simulation or Virtual test execution
\end{itemize}

In particular, this document offers a set of guidelines that describe the following aspects of virtual testing as applicable to an \ac{av} developer that develops the automated vehicles:
\begin{itemize}
	\item Overall simulation assessment work flow
	\item General capability requirements expected for a \ac{vtt}
	\item Description of scenarios and test case parameters, to conduct the virtual testing 
	\item The data format for recording the virtual testing results
	\item Salient features of the assessment procedure, including rules and metrics
\end{itemize}

\begin{figure}[h]
	\centering
	\includegraphics[scale=0.45]{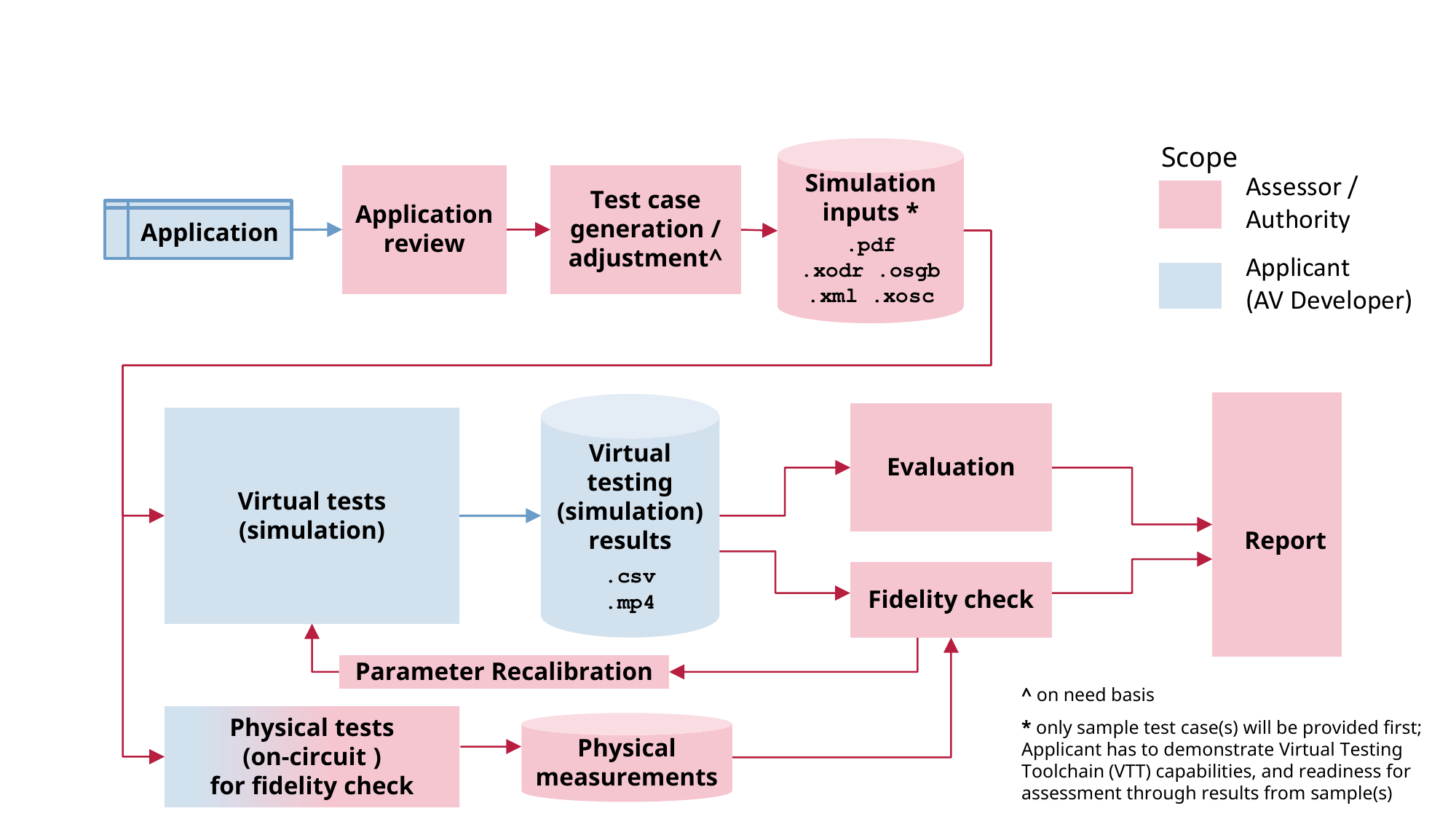}
	\caption{Overview of the Simulation Assessment Process}
	\label{fig:virtual_test_overall_workflow}
\end{figure}

\begin{sidewaysfigure}[h]
	\centering
	\includegraphics[width=\linewidth]{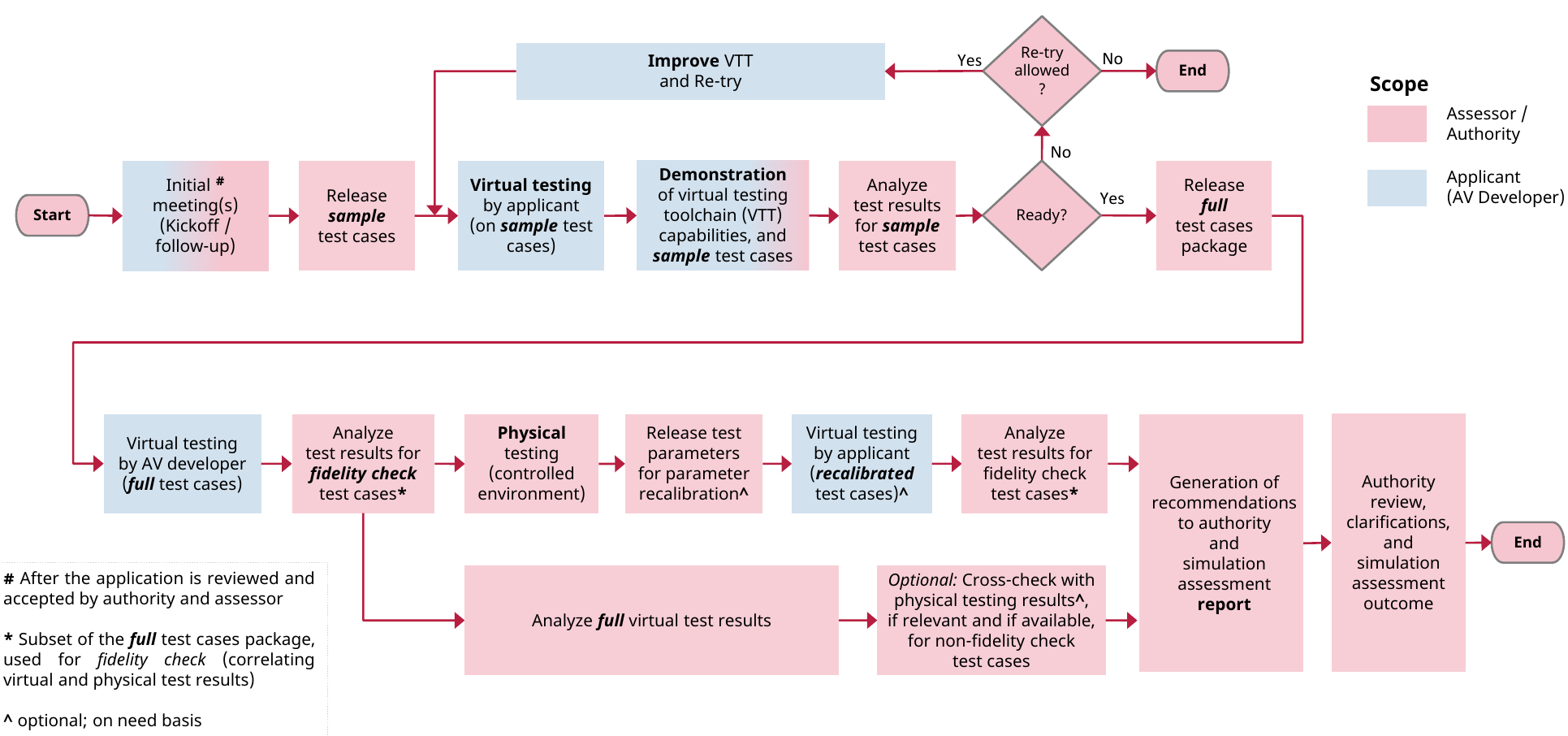}
	\caption{Flow chart showing the flow of major steps and events involved the Simulation Assessment process}
	\label{fig:virtual_test_reference_flow_chart}
\end{sidewaysfigure}

An overview of the simulation assessment process is depicted in Fig.~\ref{fig:virtual_test_overall_workflow}.


In order to conduct virtual testing (simulation) of the \ac{ads} in a virtual environment, the applicant can refer to the `simulation inputs package' provided by the assessor. 
This package typically consists of the following items:

\begin{itemize}
	\item A \textbf{Simulation Guidelines document} (this document) that includes the following:
	\begin{itemize}
		\item Description of the overall assessment pipeline (refer to this document)
		\item General capability requirements for a \ac{vtt} (refer to this document)		
		\item Working example(s) of scenarios (refer to the attachments)
		\item Format of simulation results for independent evaluation and fidelity check (refer to this document)
	\end{itemize}
	\item A \textbf{set of test cases} grouped by scenarios, each consisting of:
	\begin{itemize}
		\item Test case documentation, containing a description of the static and dynamic aspects of the test case
		\item OpenDRIVE\textsuperscript{\textregistered} file for the static environment around the \ac{av}
		\item \textit{ScenarioXML} file for the dynamic environment representing the instance of the scenario under test
		\item Graphical Database file to visualize the static  environment (in 3D)
	\end{itemize}
	
\end{itemize}

Either upon receiving the application or even during any inquiries made prior to that, the assessor shall provide the applicant (AV developer) the latest applicable generic simulation guidelines document (such as, this document) so that they can be better prepared for the simulation assessment.

After the application is reviewed and accepted, the assessor shall issue the AV developer with a few sample test case(s) as applicable to the particular \ac{vut} and to prove readiness of the virtual testing. However, this step will usually require a \ac{nda} to be formally signed beforehand between the applicant, assessor and/or the authority.
The applicant is expected to implement the sample test case(s) on their \ac{vtt} by following the procedure described in this document, and any additional detailed guidelines mentioned in the specific test case documents.
The virtual test results from such sample test case(s) will then have to be submitted to the assessor, in order to demonstrate the readiness to proceed with the full virtual simulation assessment. The results will be used to determine if the simulation results from the AV developer meets the assessor's requirements and includes all necessary parameters to carry out a successful evaluation. If the developer submits unacceptable results for the sample test cases, the assessor will provide the necessary feedback and inform the developer of the problems in the results. After this, if the developer requires more than 2 weeks to resolve the issues, the assessor may not allow the developer to re-try due to time constraints.

The applicant is also expected to demonstrate the capabilities of their \ac{vtt}, both for the execution of the sample test cases, as well as for their \ac{vtp} process established for their internal \ac{vnv} activities.
In addition, documentary evidence to show that the applicant has established a meaningful \ac{vtt} and virtual testing process to run virtual simulations regularly along with physical testing, as the \ac{av} is being developed/trialled, has to be checked separately through an independent document review procedure.

Once the authority and assessor finds the initial test results to be satisfactory (thus indicating readiness), the assessor can issue the full test case package with all applicable test cases.
Once the full test case package is provided, it is recommended that the fully generated results are submitted to the assessor within a relatively short but pre-defined time period (e.g., 2 weeks) so that the actual assessment (analysis and evaluation of virtual testing results) and fidelity check can be performed.
Along with the actual results, the applicant is also expected to submit a declaration form with details of the version(s) of the hardware/software/tools used to generate the virtual testing results, dates etc. and some information required for conducting fidelity check and evaluation.
 
As part of the assessment, the AV developer is also encouraged to demonstrate (to the authority and assessor) their actual virtual testing capabilities and any salient features of their virtual testing process used for internal verification and validation of the ADS during development.
This could be done through live demonstrations of their \ac{vtt} as well as by means of showcasing the relevant internal verification and validation reports.

The assessor will analyze the virtual testing results and evaluate them against both pre-defined and well-established objective metrics and subjective assessment by safety experts against the standard driving rules and also the latest regulations applicable to \acp{av} if any (e.g., \cite{tr68basicbehaviour,tr68safety}).
The detailed simulation assessment findings will be discussed between assessor and authority. The simulation assessment report produced will be used to recommend the overall simulation assessment outcome to the authority.

The general flow of major steps and events during a typical simulation assessment is illustrated in \autoref{fig:virtual_test_reference_flow_chart}.

\newpage

\section{Scenario-based virtual testing}
\label{sec:scenario_based_virtual_testing}

Virtual testing generally employs a \acf{vtt} to \textit{virtually} simulate scenarios involving an \acf{ads} in a closed loop feedback scheme. The ADS can be tested at different levels: model(s), software, hardware and/or full vehicle in the simulation loop.
With regard to the scope of scenario-based virtual testing within a simulation assessment for independent safety assurance, we can generally consider that ADS Software in the loop (SiL) testing is sufficient; however, Hardware in the loop (HiL) or Vehicle in the loop (ViL) may also be considered acceptable.

In general, scenario-based virtual testing allows the tester to execute a wide variety of \textit{scenario}s, which are often (but not necessarily) considered inefficient, risky, and/or infeasible to be tested physically.
This can include many rare scenarios that may not occur very frequently in real life (although their severity can be high).

Each scenario can cover multiple scenario categories \cite{degelder2020scenariocategories}, with a diversity of activity parameter sets per class. 
In fact, each \textit{scenario category} can be considered quite generic by itself.
Furthermore, each scenario can be parameterized to form concrete \textit{test case}s. 
Each test case includes a set of parameters that are assigned specific values.
The combinations of applicable parameters adds on to the diversity and volume of the tests required to cover the desired safety goals.
Many of these scenarios may also involve a high level of risk, making physical tests infeasible.
Due to the above reasons, it is imperative to test these variety of scenarios through virtual testing.



In order to perform assessment for independent safety assurance of the \ac{av}, a subset of the relevant scenarios are to be selected through a selection process, and tested using appropriate methods, that primarily includes virtual testing and physical testing. 
If direct testing is infeasible, a demonstration by the applicant may be considered as an alternative, subject to mutual agreement between the authority, assessor and the applicant.
Other alternative approaches such as formal verification or system design/implementation review may also be adopted when absolutely necessary, although these are difficult to achieve in an independent safety assurance context due to the deeper intellectual property right and confidentiality protection aspects involved.

\begin{figure}[h]
	\centering
	\includegraphics[scale=0.5]{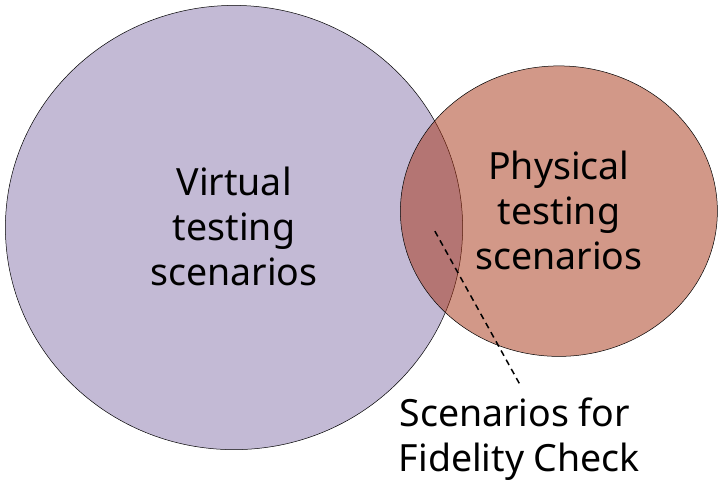}
	\caption{Diagram shows the relationship between the set of scenarios for physical and virtual testing, as well the common ones used to check the fidelity of the virtual testing toolchain}
	\label{fig:VirtPhys}
\end{figure}

For the purpose of a simulation assessment for independent safety assurance, a subset of the relevant scenarios are to be selected and tested virtually, by making use of a \acf{vtt}.
The actual procedure\footnote{\label{footnotetestgenprocess}This process may be internal to the assessor and/or authority} of selecting the subset of relevant scenarios for assessment, can consider various factors or selection criteria. This may include the general risk or challenges that the \ac{av} may be exposed to in the given scenario, feasibility of execution, adherence to the ODD stated by the applicant, and much more.
The selected set of scenarios can also include some common scenarios that are used to ascertain the validity of the \ac{vtt} through a Fidelity Check process, and these scenarios will have to be tested both physically and virtually.
Figure.~\ref{fig:VirtPhys} illustrates the overlap of the physical and virtual testing scenarios in a generic setting.
However, the actual and specific set of concrete test cases with concrete parameters, that are used in a particular assessment are selected through a test case generation process\footnotemark[\getrefnumber{footnotetestgenprocess}].


\begin{table}[h]
	\setlength\nwidth{5.5em}    
	\setlength\nheight{6em}     
	\centering
	\caption{Scenarios as simulation inputs}
	\label{tab:SimInputs}
	\begin{tabular}{>{\centering}m{\nwidth}m{50mm}m{\textwidth - 50mm - \nwidth - 6\tabcolsep}}
		\hline\hline
		\textbf{Name} & \textbf{Description} & \textbf{Remarks}\tabularnewline
		\hline\smallskip
		\tikz{\node[block] {\includegraphics[width=2.5em]{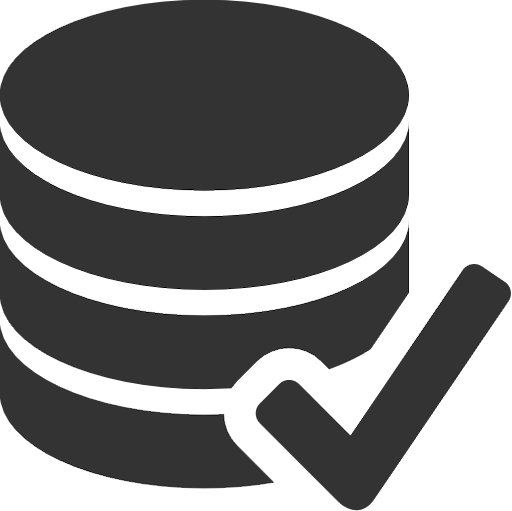}\\Set of scenarios for virtual testing}}
		& The set of scenarios employed for virtual assessment of the safety and performance of the \ac{av}.
		&  Complexity and risk associated with these scenarios can vary from easy to very challenging, based on variations of parameter values. 
		Some scenarios may have to be performed separately in different special operation modes wherever applicable.
		\tabularnewline
		\hline\smallskip
		\tikz{\node[block] {\includegraphics[width=2.5em]{scenario_verification.png}\\Set of scenarios for fidelity check};}
		& Common set of scenarios for both virtual and physical testing
		& This set of scenarios is used for a Fidelity Check process. The validation will be done by looking at the consistency between the ADS performance results from the test in virtual world (virtual tests) and real world (physical tests) under a controlled environment.
		\tabularnewline
		\hline\hline
	\end{tabular}
\end{table}

\clearpage
\subsection{Virtual testing process with a reference virtual testing toolchain}
\label{sec:virtual testing process}


\begin{figure}
	\centering
	\includegraphics[width=\columnwidth]{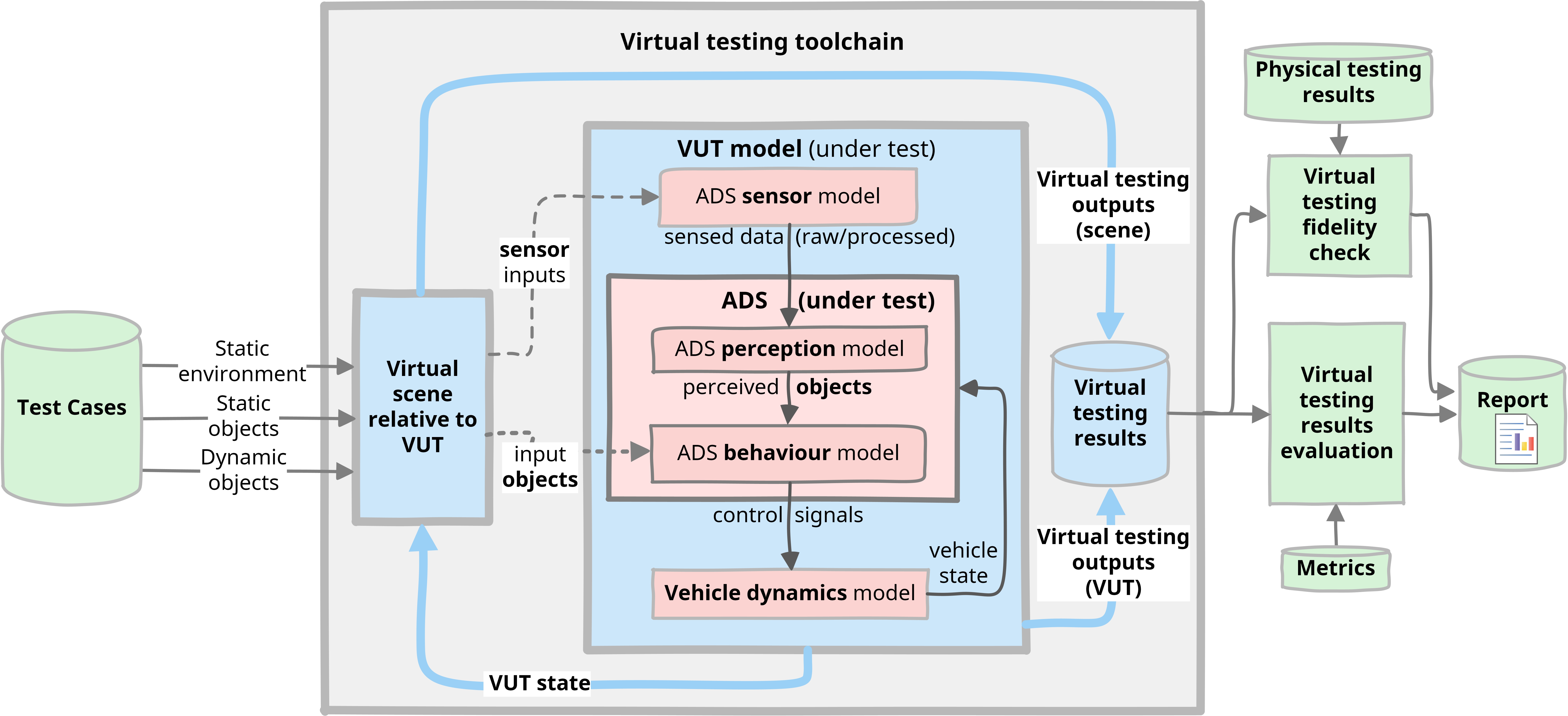}
	\caption{Overview of a reference virtual testing process, as recommended to implement the virtual testing in the context of the simulation assessment methodology.}
	\label{fig:simulation setup}
\end{figure}

The \autoref{fig:simulation setup} illustrates some of the structural and functional aspects of the virtual testing process recommended to be followed by the applicant that shall be assessed by the assessor.

The concrete test cases to be used for this process are to be generated, selected, and verified based on various inputs, such as described in the previous section \ref{sec:scenario_based_virtual_testing}. 

For the internal \acf{vnv}, the applicant is expected to generate a sufficient number of test cases to achieve an exhaustive test suite that can fulfill their \ac{vnv} in a comprehensive manner.
For the independent assessment by assessor, these test cases shall be additionally provided by the assessor and/or authority. 
Optionally, the assessor may also select a subset of the existing test cases from the applicant's \ac{vnv} test suite, if the assessor can be allowed to access these to perform a spot check with selected samples.

The blue and red blocks in \autoref{fig:simulation setup} show a possible \acf{vtt} that the applicant can use for deriving the resulting behaviour of the \ac{av}. 
For each test, the applicant is requested to provide the corresponding response of the \ac{av}; this shall include results from tests performed on the virtual testing toolchain and optionally, from physical tests conducted by the applicant.
Furthermore, as part of the independent safety assessment process, a fidelity check is performed to gauge the fidelity of the \ac{vtt} and the validity of the virtual testing results generated, before the complete set of virtual testing results can be evaluated.

The applicant is expected to simulate realistic perception capabilities in the virtual environment.
As illustrated in \autoref{fig:simulation setup}, this could be achieved through either a realistic sensor simulation that generates sensed data (which may be in raw, processed, or both forms depending on the actual sensors used), or a representative perception output such as a static/dynamic object list.
In the latter case, the object list may be generated by a tool and/or human expert(s); however, this must be validated through a perception validation procedure which has to be separately demonstrated to the assessor. 
This step is necessary because the perception is a largely non-deterministic component of the \ac{ads} that implements the crucial \ac{oedr} functionality, which may require some special tests. 
Such special tests are intended to prove and/or to prove confidence that the perception is robust (meets \acs{sotif} requirements) as an independent component, as a part of the fully integrated \ac{ads}, and of the \ac{av} as a complete system.

\newpage

\section{Simulation inputs}


In order to virtually simulate a particular test case representing an instance of a scenario, the following three files will be provided:
\begin{enumerate}
	\item OpenDRIVE\textsuperscript{\textregistered} - defines static environment and road network
	\item \textit{ScenarioXML} - adds dynamic objects (including stationary obstacles) and actors (or players) to the defined static environment
	\item OSGB graphical database - enables the graphical (3D) visualization of the static and dynamic environment
\end{enumerate}

\subsection{Static environment preparation}  

OpenDRIVE\textsuperscript{\textregistered} is a well defined and openly described format that defines the complete static environment which mainly consists of the road network. 
The typical file extension is \texttt{*.xodr}. This file uses XML syntax to define items, so the *.xodr files can be read or edited by any text editor.\\

The OpenDRIVE\textsuperscript{\textregistered} file format provides the following features and much more:

\begin{itemize}
	\item XML format
	\item Hierarchical structure
	\item Analytical definition of road geometry (plane elements, elevation, cross-fall, lane width etc.)
	\item Various types of lanes
	\item Junctions including priorities
	\item Logical inter-connection of lanes
	\item Signs and signals including dependencies
	\item Road and road-side objects
\end{itemize}

All the *.xodr files, that are provided for AV simulation assessment, are defined in the 1.4H version\footnote{The latest version of OpenDRIVE\textsuperscript{\textregistered} format as of this document is 1.7 (released 03 Aug 2021). However, since tool support to version 1.7 is very limited, we still adopt version \textbf{1.4H}, primarily to facilitate general tool compatibility. This is because the version 1.4H is widely supported by a large number of AV simulation tools and road environment design tools.} of OpenDRIVE\textsuperscript{\textregistered} format. The versions shall be progressively updated to follow the newest releases of the format after due deliberation. Complete description of the format and its versions are available at \url{http://www.OpenDRIVE.org/}.

Furthermore, the graphical database is provided in osgb format, which is is primarily a binary compiled open scene graph, with a 3D model of the environment that includes all textures that can be used to visualize the road network and 3D environment.

\subsection{Test case preparation}  

The applicant can implement the test cases in their \ac{vtt} primarily by referring to the concrete scenario and parameters described in the test cases document\footnote{This will usually require an \ac{nda} to be signed between the applicant, assessor and/or the authority.\label{fn-nda}} that will be provided separately during the course of assessment.
In addition, for convenience, the assessor can provide the concrete test cases described in a standard scenario description file format that can be used by a simulator tool to play or execute the scenario.

Currently, it is recommended that, as part of the test case package, the concrete test cases are provided to applicants in an XML-based scenario description format, namely \textit{ScenarioXML}.
In later editions of the assessment, in future, the assessor may upgrade to a more open domain-specific language and file format such as OpenSCENARIO\textsuperscript{\textregistered} instead of \textit{ScenarioXML}, after due deliberation. 

The \textit{ScenarioXML} file is used to represent the dynamic aspects of the test case by adding dynamic items and actors (players) to the defined static environment. 
The \textit{ScenarioXML} file uses standard \ac{xml} syntax with a proprietary schema\footnote{Supported by commercial tools such as Virtual Test Drive (VTD) from VIRES Simulationstechnologie GmbH. of Hexagon/MSC group}, that is used to describe dynamic objects and actors in the virtual environment. The file extension used is \texttt{*.xml}.

To define a test case in such a file format, the static environment (road network) has to be first described in OpenDRIVE\textsuperscript{\textregistered} format and provided as an input. 
Once the road network is available, \textit{ScenarioXML} defines values of all the dynamic parameters needed to simulate a specific scenario. 
This includes trajectories, velocities, weather condition, triggering points for TSVs and VRUs. 
It also defines types of TSV or VRU, behaviour of drivers etc.
Exact set of parameters and values are individually specified for each test case.
More information about the format may be obtained at 
\url{https://www.asam.net/standards/detail/opendrive/}.

In any case if the \ac{vtt} used by the applicant does not support the provided test case format, the corresponding test cases can be implemented by using the concrete scenario and parameters described in the test cases document\footnotemark[\getrefnumber{fn-nda}] 
that will be provided separately during the course of assessment.

\begin{table}[h]
	\setlength\nwidth{5.5em}    
	\setlength\nheight{6em}     
	\centering
	\caption{Scenarios preparation}
	\label{tab:ScenarioPreparation}
	\begin{tabular}{>{\centering}m{\nwidth}m{52mm}m{\textwidth - 52mm - \nwidth - 6\tabcolsep}}
		\hline\hline
		\textbf{Type} & \textbf{Description} & \textbf{Remarks}\tabularnewline
		\hline\smallskip
		\tikz{\node[block2]{\includegraphics[width=4em]{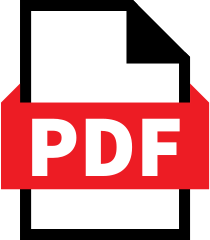}\\};}
		& Test case documentation PDF file containing the details of each test case for implementation
		& Complete detailed description of the test case including the scenario, the environment, and the concrete parameters for actors. Applicant can refer to this and newly implement the test case in their \ac{vtt} with their own static environment and scenario editing tool, but with the same parameters as specified in the document. \tabularnewline
		\hline\smallskip
		\tikz{\node[block2]{\includegraphics[width=5em]{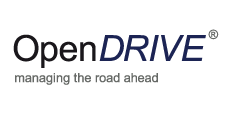}\\};}
		& OpenDRIVE\textsuperscript{\textregistered} format file for each test case
		& Complete description of the relevant static road environment required by the test case\tabularnewline
		\hline\smallskip
		\tikz{\node[block2]{\includegraphics[width=5em]{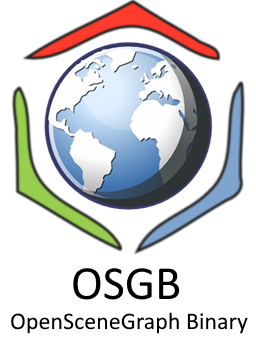}\\};}
		& OpenSceneGraph Binary (OSGB) format file which describes the 3D environment applicable for each test case and references the corresponding OpenDRIVE\textsuperscript{\textregistered} file for its static road environment
		& Complete description of the relevant static road environment required by the test case. 
		OSGB is a serialized binary representation of the OpenSceneGraph 3D model, with all textures contained within one standalone file.
		\tabularnewline
		\hline\smallskip		
		\tikz{\node[block2]{\textit{ScenarioXML}\\};}
		& \textit{ScenarioXML} format file which independently defines each test case and references the  OpenDRIVE\textsuperscript{\textregistered} and OSGB file for its static road environment
		& Description of the dynamic objects and actors (players) in the scenario. 
		This file contains the exact values of parameters required to virtually simulate the scenario. In future, OpenSCENARIO\textsuperscript{\textregistered} (*.xosc) could be used instead of this. \tabularnewline
		\hline\hline

	\end{tabular}
\end{table}

\newpage

\section{Simulation results}

Once the virtual tests results are generated, they should be submitted in a specified format as described in this section for the evaluation by assessor and authority.

\subsection{Results data format}
\label{ssec:results_format}

It is recommended that the virtual tests results data shall be provided in a flat .csv file or a package of distributed .csv files \footnote{other file formats may be possible, but the schema is not available currently. This can be discussed and provided separately, if necessary}. 

The results data shall contain the following information about the \ac{vut} and corresponding dynamic environment around the \ac{vut}, at each simulation step with the corresponding time-stamps.
This shall be recorded either into one .csv file (single flat file format) or folder of .csv files (distributed file format) per test case run.
The specific parameters that should be included in the .csv file are described in Appendix section~\ref{app:ssec:results_format}.

The results file or folder shall be named as \texttt{results\_<testcase\_id>\_r<run\_id>.csv} according to the test case id \texttt{<testcase\_id>} and test case run number \texttt{<run\_id>} (with values 1 to $n$, where $n$ is defined separately).
Alternatively, a distributed file format, where the same data is distributed into separate .csv files each recording the status of VUT, actors, obstacles and traffic lights separately, is also possible. More details can be found in the Appendix section~\ref{app:ssec:csv_file_contents_distributed}.

Sample reference results data files in .csv format are available for download and described under the Appendix \ref{app:ssec:results_format}.
For example, a flat .csv file corresponding to test case id \texttt{M2-CL4-S-TST-05-01} and run id \texttt{09}) is to be named as \texttt{results\_M2-CL4-S-TST-05-01\_r09.csv} and formatted as per the recommended result parameters.
The Appendix also provides examples for different VUT behaviors under the same scenario and some insights on how they may be evaluated.

A summary of the parameters that should be included in the .csv file are listed below (and details in Appendix section~\ref{app:ssec:results_format}).

\begin{itemize}
	\item Dynamic information about the VUT, at each simulation step, with the corresponding timestamp
	\begin{itemize}
		\item VUT position (in WGS84 coordinate system) at \ac{cg}
		\item VUT travelled distance
		\item VUT velocity
		\item VUT acceleration (lateral, longitudinal) at \ac{cg}
		\item VUT yaw, pitch, roll rates at \ac{cg}
		\item VUT heading angle
		\item VUT indicator lights status for direction (left/right turn at both front and back of the VUT), brake, reverse and hazard
		\item VUT throttle/brake level
		\item VUT steering wheel angle
		\item VUT drive status (autonomous mode, manual mode or tele-operation mode)\footnote{If the VUT operation cannot be fully described by these 3 modes alone, then the AV developer is expected to propose additional modes and inform the assessor before providing the simulation results.}
		\item VUT special operation status (normal operation mode, environmental service mode, or any other special operation mode)\footnote{If the VUT has a special operation mode, the current status (e.g., whether environmental service mode or non-environmental service mode is active) shall be recorded. If this status cannot be logged from VTT, then the AV developer must propose any alternative before submitting the simulation results.}
		
	\end{itemize}

	\item Information on each static obstacle (e.g., construction cones, carton, fallen tree branch etc.), chiefly based on ground-truth, at each simulation step with timestamp
	\begin{itemize}
		\item Nearest temporal distance\footnote{\textbf{Nearest temporal distance }is defined as the time taken by \acs{av} to reach the closest point on the obstacle,  object or actor as the case may be\label{fn-temporal-distance}} as estimated from the VUT
		\begin{itemize}
			\item This could be the \ac{ttc}, if the VUT and obstacle are on a collision course
		\end{itemize}
		\item Type of obstacle
		\item Position (in WGS84 coordinate system) at center of the obstacle
			\item Bounding polygon (ground truth)
			\item Bounding polygon (perceived), typically based on the output of ADS perception
	\end{itemize}

	\item Information on each dynamic object or actor (such as a TSV or VRU), chiefly based on ground-truth, at each simulation step with timestamp
	\begin{itemize}
		\item Nearest temporal distance\footnotemark[\getrefnumber{fn-temporal-distance}] as estimated from the VUT
		\begin{itemize}
			\item This could be the \ac{ttc}, if the VUT and actor are on a collision course, such as during vehicle following.
		\end{itemize}		
		\item Type of actor (TSV or VRU)
		\item Position at the geometric center (in WGS84 coordinate system, or alternatively in \ac{vcs} relative to VUT position. For details, see VCS description and data format in Section~\ref{typedef:pos_vcs} and Table~\ref{tab:results_csv_format_actors_true}
		)
		\item Bounding box (ground truth)
		\item Bounding box (perceived), typically based on the output of ADS perception
		\item Speed
		\item Velocity (lateral, longitudinal) at \ac{cg}
		\item Acceleration (lateral, longitudinal) at \ac{cg}
		\item Heading angle (w.r.t. geographic North)		
	\end{itemize}

	\item Information on each traffic light controller, based on ground-truth, at each simulation step with timestamp
	\begin{itemize}
		\item ID of the traffic light controller
		\item Current phase of the traffic light controller (e.g., \texttt{go}, \texttt{stop}, \texttt{go\_exclusive})
	\end{itemize}
\end{itemize}

The data to be included into the .csv file has to be logged at a certain expected minimum frequency $f$ (e.g., 10 Hz). In other words, at least $f$ data records ($f$ rows in the .csv file) recorded at equally spaced time intervals are expected, for 1 second of simulation results data.

To ensure consistency of vehicle behavior in the simulation environment, every test case has to be simulated $n$ times (e.g., $n$ = 10) with the same set of parameters. 
The value of $n$ can be individually changed for selected test cases and such exceptions, if any, shall be defined in respective test case description.
Additional information (such as screenshot images) about the simulation can be required for specific test cases. Such information can be attached together with the corresponding results file wherever specified.

\subsection{Video recording of virtual test runs}

To help the assessor and authority\footnote{In special cases, the Traffic Police may also be invited to review the videos on need basis.} to better understand and assess the simulation, the applicant (AV developer) is required to submit a video of each test case that is run in the simulation environment.

\subsubsection{Video contents and layout}

The video should contain below information:
\begin{itemize}
	\item an elevated view of the relevant static environment at a perspective that covers all the relevant actors of the scenario under test
	\item the relevant features of the environment especially the road features (e.g., lane markings, stop lines, give-way lines, broken / dashed / solid / zig-zag lines, yellow box, pedestrian crossings, bus stop/bay markings, other road markings, kerb lines, entry/exit points of incoming/outgoing minor roads, etc.)
	\item the current speed and heading of the VUT
	\item the current speed (and optionally, the heading) of actors
\end{itemize}
Such information will be useful to analyze the dynamic behavior of the VUT in a given scenario involving actors and also for easily locating the vehicle / actors visually during manual video analysis by human eyes.

\begin{figure}[h]
	\centering
	\includegraphics[width=1.0\linewidth]{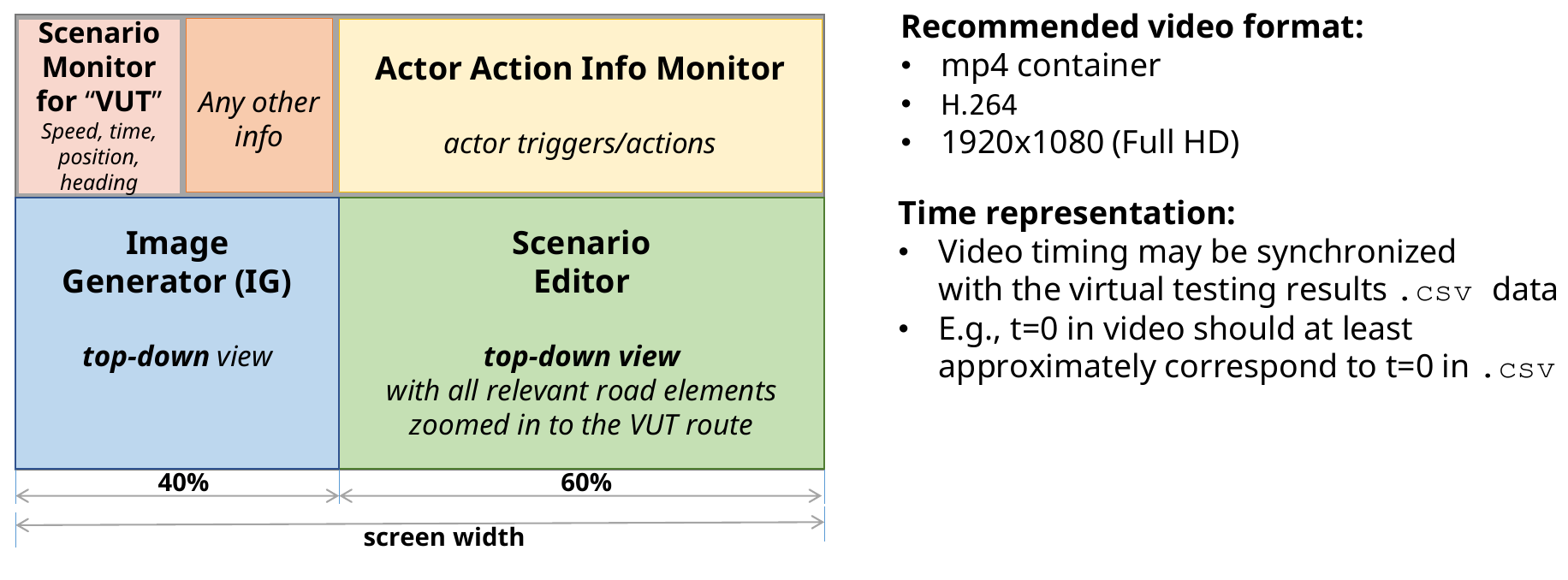}
	\caption{Illustration of how the video content may be arranged in a meaningful layout}
	\label{fig:results_video_format_illustration}
\end{figure}

\begin{figure}[h]
	\centering
	\includegraphics[width=1.0\linewidth]{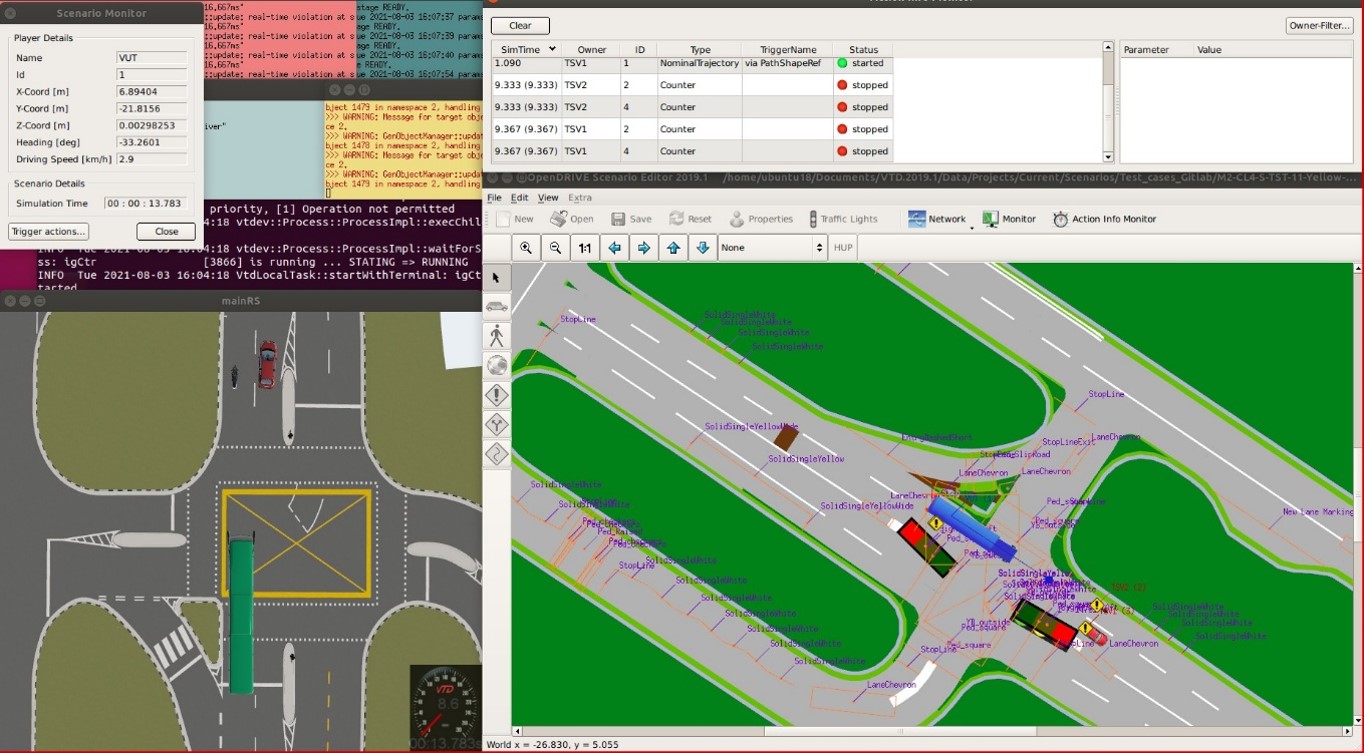}
	\caption{An example of the video layout, using VIRES VTD simulator}
	\label{fig:results_video_format_example}
\end{figure}

Depending on the tool capability, such information may be directly overlaid in the main simulator window, or displayed in separate window(s). When displayed in separate windows, they must be re-arranged and positioned such that they are all included in the same frame of a recorded video. An illustration of the layout can be found in Figure~\ref{fig:results_video_format_illustration} and an example generated using VIRES VTD simulator can be found in Figure~\ref{fig:results_video_format_example}.

For each test case, only 1 video may be sufficient but the corresponding test case run number (out of the $n$ times it was executed) must be specified for ease of comparison. 
However, additional videos may be specifically requested for any particular runs, if this is required for deeper analysis of issues seen in those specific runs.

\subsubsection{Video output format}

The recommended output video format to be submitted is as follows
\begin{itemize}
	\item Video container: mp4
	\item Codec: H.264
	\item Resolution: At least 1920x1080 (Full HD)
\end{itemize}

The video file should be named in the same way as for the corresponding test case run, but with the extension \texttt{.mp4}.
For example, the video file named \texttt{results\_M2-CL4-S-TST-05-01\_r09.mp4} corresponds to the results data file \texttt{results\_M2-CL4-S-TST-05-01\_r09.csv}.

\subsubsection{Time representation}
Video timing has to be synchronized with the virtual testing results .csv data.
E.g., $t=0.000 s$ in video should at least approximately correspond to $t=0.000 s$ in the .csv results data file(s) for the corresponding test case run.

\subsection{Data format correctness and integrity check using samples}

In order to verify the data integrity of the .csv files (such as the correctness, completeness and compliance to the data format guidelines), it is recommended that the AV developer could initially submit the sample results from 1 complete run each for a few (e.g., 2-3) selected test cases involving multiple dynamic actors; these sample test cases could be selected through mutual discussion. 
If the VUT has a special operation mode applicable, then, separate results for the each of the different possible special operation mode values may be required separately.
After the integrity of these samples have been successfully verified by the assessor, the applicant can proceed to generate the remaining results and submit them for a formal evaluation.

\subsection{Simulation results declaration form}

While submitting the results and videos to the assessor, the applicant is also expected to submit a declaration form\footnote{The declaration form template shall be provided to the applicant after the application review process is completed and together with the simulation inputs package.}. This form contains details of the version(s) of the hardware/software/tools that the applicant has used to generate the virtual testing results, the dates of test execution and result generation etc. 
Furthermore, it contains some additional questions on the information required for preparation and conduct of fidelity check and  evaluation of the simulation results.
In particular, this will involve questions to gather details on how the applicant intends to drive the \ac{av} during vehicle dynamics fidelity check as part of physical on-circuit testing.
The versions used for virtual testing, as declared in this form, will be compared against the corresponding versions used during on-circuit testing.

\subsection{Consistency of VUT between virtual and physical testing}

It is generally expected that the applicant will ensure that the same version of system, hardware, software, configuration and/or map will be used for both virtual and physical testing of the VUT.
In case if there are differences, the key differences must be stated and they must be justified with proper causes, reasoning and impact analysis.
\newpage

\section{Assessment procedure and Metrics}
\label{sec:eval}

After the virtual simulation results are submitted to the assessor, the assessor shall perform the following: 
\begin{itemize}
	\item an exhaustive \textit{simulation results evaluation} procedure to evaluate the behavior of the \ac{av} as per these simulation results
	\item a \textit{fidelity check} proceudre to validate these simulation results (at least a part of them) against corresponding physical test results in comparable test conditions.
\end{itemize}

\subsection{Evaluation of simulation results}
\label{ssec:eval_sim_results}
The simulation results will be evaluated against the expected safe behavior for each test case, in accordance to the general driving rules in Singapore. 
In particular, the below listed key performance \textit{metrics} (a.k.a. Key Performance Indicators or KPI) shall be used to achieve a significantly objective evaluation wherever possible.

\subsubsection{Adherence to general driving rules}
	\label{sssec:eval_general_driving_rules}
	In general, the AV is expected to adhere to the established rules in the jurisdiction for road vehicles driving on public roads. For \acp{av} driving in Singapore, they are expected to comply with the latest  Road Traffic Act \cite{road_traffic_act} and the Basic and Final Theory of Driving handbook (refer to the latest edition released by the Traffic Police) \cite{trafficpolice2017handbook, trafficpolicebook2010}.
	
	The simulation evaluation process is expected to check whether the AV exhibits any significant non-compliance (deviations and/or violations) of these rules, that could have been avoided through the prior actions of the AV.
	In particular, the severity and exposure (frequency of occurrence) of such non-compliance may be considered in making a judgement.

\subsubsection{Metrics to evaluate simulation results}
\label{sssec:eval_KPI}
%

\textbf{Metrics for safety of \acs{tsv}s/\acs{vru}s\\}
The metrics listed here will be used to evaluate whether the \ac{vut} behavior will ensure the safety of vulnerable road users (\ac{vru}s such as pedestrians or cyclists) and/or other on-road vehicles (such as stationary or incoming \ac{tsv}s) for each scenario.
\begin{itemize}
	\item Lateral Clearance
	\item Longitudinal Clearance
	\item Nearest Temporal distance\footnote{time taken by \acs{av} to reach the closest point on the \acs{tsv}}
	
\end{itemize}

%
%

{\textbf{General threshold values of metrics\\}
In general, the \ac{vut} is expected to maintain the values of the metrics to remain under certain threshold values (for e.g., refer Table~\ref{tab:KPI_values_general} for the thresholds defined) throughout driving. 
These threshold values are based on the rules specified for vehicles driving in Singapore according to the Basic and Final Theory of Driving handbook (refer to the latest edition released by the Singapore Traffic Police \cite{trafficpolice2017handbook,trafficpolicebook2010}).
However, special-purpose exceptions may be made on these threshold values for specific scenarios, as defined and documented for each test case.
%

\begin{table}[htb]
	\vspace{3ex}
	\centering
	\caption{Description of some metrics used and their general threshold values}
	\label{tab:KPI_values_general}	
	\begin{tabular}{|l|l|p{0.5\textwidth}|}
		\hline
		\textbf{Metric} & \textbf{Threshold} & \textbf{Context} \\ \hline
		\multirow{3}{*}{
			\makecell[l]{Lateral Clearance \\(Refer Fig.~\ref{fig:exclusion_zone_VUT})}} 
		& $\geq$ 0.5m      &
		\makecell[l]{
			\tabitem static obstacle\\
		}
		\\ \cline{2-3} 
		& $\geq$ 1m        & 
		\makecell[l]{
			\tabitem stopped or parked vehicle\\
			\tabitem pedestrian facing traffic
		}
		\\ \cline{2-3} 
		& $\geq$ 1.5m        & 
		\makecell[l]{
			\tabitem moving TSV\\
			\tabitem pedestrian facing away from traffic\\
			\tabitem cyclist\\				
			\tabitem PMD rider
		}		           
		\\ \hline
		\makecell[l]{Longitudinal Clearance \\(Refer Fig.~\ref{fig:exclusion_zone_VUT})} &  $\geq$ 2m                 &
		\makecell[l]{ 
			\tabitem any road user (TSV/VRU) ahead of VUT\\
			\tabitem any obstacle ahead of VUT
		}
		\\ \hline
	\end{tabular}
\end{table}

With regard to the lateral and longitudinal clearance thresholds specified in the Table~\ref{tab:KPI_values_general}, the VUT is expected to maintain an Exclusion Zone as defined below and as represented in Fig.~\ref{fig:exclusion_zone_VUT} during normal driving, even when there are no obstacles. 

{\textbf{Exclusion Zone\\}
	An \textit{exclusion zone} is defined around the autonomous vehicle (see Fig.~\ref{fig:exclusion_zone_VUT}). 
	No object (car, pedestrian etc.) or obstacle should enter this zone \textit{due to the actions of the autonomous vehicle} when it is in operation. (The exclusion zone does not apply to fixed infrastructure on the road/kerbside e.g., lampposts, signposts, pillars, trees, traffic lights, traffic light controller boxes, kerbs etc).

\begin{figure}[htb]
	\centering
	\includegraphics[width=\linewidth]{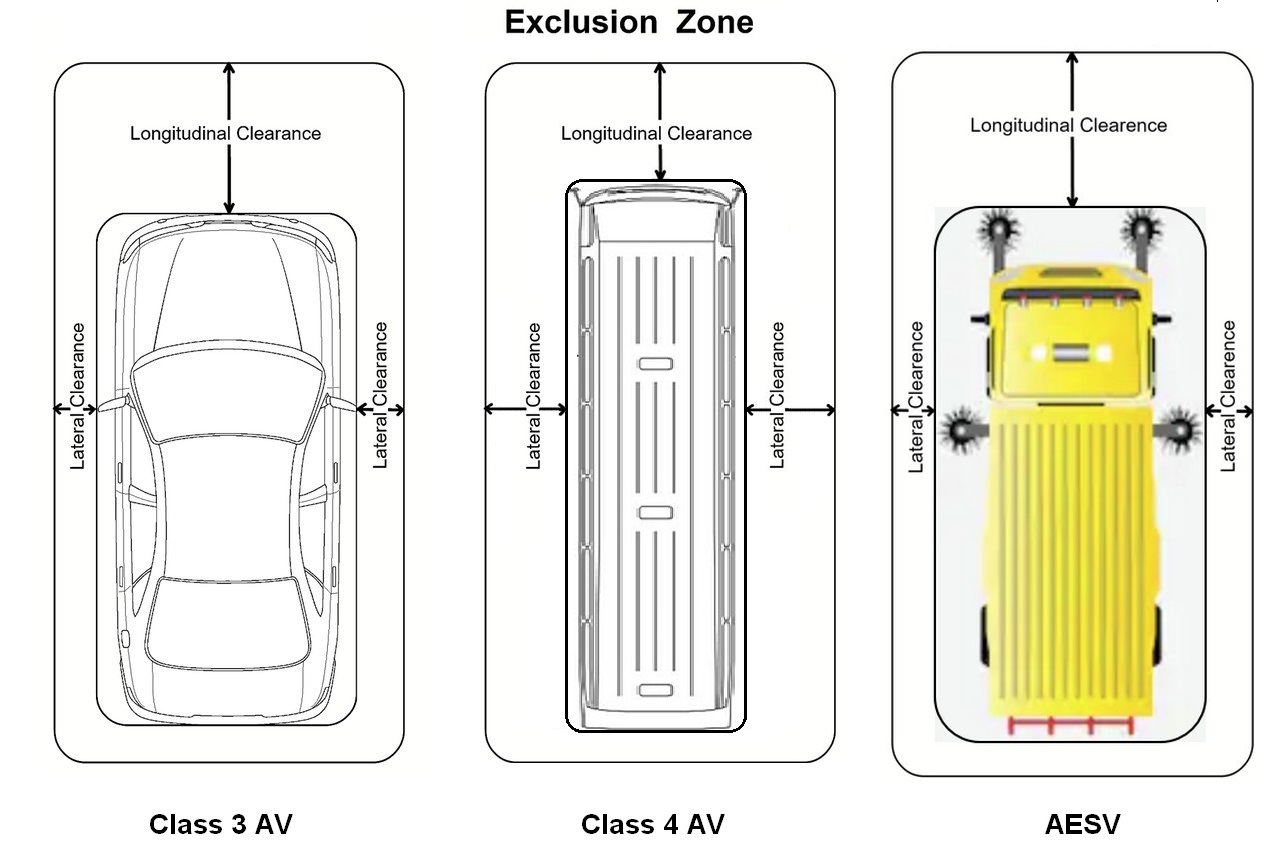}
	\caption{Exclusion zones defined around the VUT for various VUT types: (a) Class 3 AV e.g., an automated passenger car or taxi, (b) Class 4 AV, e.g., an automated  bus or truck (c) \ac{aesv} which may be either Class 3 or Class 4, e.g., an automated road sweeper vehicle. Even for the same VUT type, different exclusion zones may be applicable when evaluating the performance of VUT, depending on the context, as the VUT executes various driving tasks as per Table \ref{tab:KPI_values_general}. For a formal definition of the vehicle types as applicable to Singapore road traffic, please refer to the Road Traffic Act Rules \cite{road_traffic_act}.}
	\label{fig:exclusion_zone_VUT}
\end{figure}

The exclusion zone applicable may be different depending on the type of the \ac{vut}. E.g., whether it is an automated Class 3 passenger taxi, or a Class 4 automated bus, or an automated road sweeper.
For a formal definition of such vehicle types, one may refer to the Singapore Road Traffic Act (Chapter 276) Road Traffic Rules item \#19 under the section titled ``Classification of motor vehicles'' \cite{road_traffic_act}.

%
For example, a lateral clearance of 0.5m must be maintained from any static obstacle on or along the road, 
and a longitudinal clearance of 2m must be maintained from the road user(s) in front (always).
The clearances are measured between the outer bounds of the VUT and the obstacle. As illustrated in Fig.~\ref{fig:exclusion_zone_VUT}, this considers the outermost protruding parts. For example, the brushes for an AESV are considered to be within the bounds of the AESV and must not be included in the clearances calculated.

Note that the specified clearance threshold values may not be strictly applicable for certain types of vehicles under specific exceptional conditions. Some examples are listed below:
\begin{itemize}
	\item Both lateral and longitudinal clearance requirement may not be strictly applicable: 
	\begin{itemize}
		\item under situations when another road user (such as a VRU or TSV) enters into the exclusion zone of the VUT. 
		For example, when a motorcycle overtakes the VUT by driving near the lane marker and between two lanes (in this case, the motorcycle may enter within the 1.5m lateral clearance threshold of the VUT). 
		
	\end{itemize}
\end{itemize}

	Thresholds for such cases, vehicle types, their special operational modes, and/or exceptional situations may be defined individually for the applicable test cases if and when necessary.

{\textbf{Optional Metrics for future consideration\\}

Although they may not be directly used in current assessments, the below metrics have been short-listed for future consideration, as they have significant potential for an independent assessment of the behavioral safety of an \ac{av} in virtual simulation.

\begin{itemize}
	\item Longitudinal/lateral acceleration \cite{sae2021metrics}
	\item Longitudinal/lateral jerk	\cite{sae2021metrics} 
	\item OEDR reaction time \cite{sae2021metrics}
\end{itemize}

%

\clearpage

\subsection{Fidelity Check: Validation of the virtual simulation results}
\label{ssec:validation_physical}
%
In addition to the evaluation described above, a subset of the test cases shall be used to validate the \ac{vtt} by following a Fidelity check procedure that involves physical testing. 

The fidelity check procedure involves a detailed sanity check by performing a statistical comparison of the results obtained from physical tests\footnote{These physical test results will be obtained from a \ac{gnss} / \ac{ins}-based measurement system which is to be externally mounted on the \ac{vut} during physical testing.} of this subset against their corresponding simulation results.
The key idea behind this procedure is to check for evidence of general \textit{consistency} between virtual testing results and real world performance under the same (or similar) conditions in a controlled environment. 
In other words, this is to check whether the virtual testing toolchain can be considered to be sufficiently representative of the real \ac{av} driving in the real world.

\textit{Any} of the test cases from the simulation test case package may be used for the Fidelity check. 
This selection of appropriate test cases will be performed by the assessor, using an internal process. 
In general, after the applicant has submitted the virtual testing results for the test package, a subset of the test cases will be used to enact physical circuit-based tests for the purpose of checking fidelity. 
This can include a few test cases that are specifically designed for fidelity check, with a focus on the general representativeness of the AV behavior and vehicle dynamics.

If the virtual simulation is done only one-time and only before the actual physical tests, the Fidelity check may not be effective in confirming the validity of the virtual simulation. 
This is because, the actual real-world test parameters used in physical tests may not always be repeatable due to practical challenges and constraints.
Therefore, we intend to follow the below procedure:
\begin{itemize}
	\item After the on-circuit physical test, the assessor may request the applicant to \textit{recalibrate} the test parameters of some of the already implemented virtual test cases. This can help to match the parameters actually used during the circuit test. 
	\item For these recalibrated test cases, the applicant is expected to re-submit the virtual test results to assessor.
	\item If there are any major differences in AV performance/behaviour between the original test results and the resubmitted post-recalibration results, the applicant is expected to provide justifications.
\end{itemize}


\textit{Recalibration} of test parameters is expected to be done only for a small subset of test cases that are already implemented and selected for fidelity check, e.g., only around 20-25\% of the total number of already implemented test cases. Therefore, the additional effort for this recalibration is not expected to be significant.

\newpage

\section{Virtual Testing Capability Requirements}
\label{sec:req_vtt}

While applying for the simulation assessment, the applicant is expected to declare the known capabilities of the \acf{vtt} that is (or, can be) used for verification and validation of the safety of their \ac{av}, and state the known limitations if any.

These general virtual testing capability requirements are summarized below in \autoref{tab:req_vtt}.

In general, these requirements are listed for two different scopes of simulation assessments, each of which are intended to assess two broadly different levels of capability and maturity that one would expect from the \ac{av}. 
\begin{enumerate}
	\item \textbf{With onboard \ac{so}}: for assessing \acp{av} equipped with technology which is still under active development, and will therefore require an onboard \ac{so} to safely conduct the AV trials even within a given \ac{odd}. 
	\item\textbf{Without onboard \ac{so}}: for assessing \acp{av} equipped with relatively well-developed technology such that for a given \ac{odd}, the \ac{av} would be expected to perceive its environment and drive autonomously and mitigate failures (transition to \ac{mrc} or perform any \ac{mrm}) without any onboard \ac{so}.
\end{enumerate}

In the context of the existing safety assessment framework established in Singapore for \acp{av}, the two types of assessments mentioned above may correspond to the \acf{m2} and \acf{m3} assessments respectively.


\begin{longtable}{>{\raggedright}p{3cm}>{\raggedright\arraybackslash} p{7.75cm} p{1.75cm} p{2cm}}
	\caption{General Virtual Testing Capability Requirements}
	\label{tab:req_vtt}
	\\ 
	\toprule			
	\bfseries Capability \mbox{requirement item} & \bfseries Description 
	& \bfseries Assessed? (with onboard \acs{so})  
	& \bfseries Assessed? (\mbox{without} onboard \acs{so}) 
	\\ 
	\midrule
	\endhead 
	\textsf{Scenario creation} & The VTT is able to create urban driving and/or traffic scenarios that are necessary to test the capability of the \ac{av} to drive safely within the specified ODD. & Yes & Yes \\
	\midrule
	\textsf{Static environment \mbox{models}} & The VTT is able to model (or import existing models such as in OpenDRIVE or any standard HD map format) the static environment, which encompasses the road network, road furniture, traffic signals and any other static (permanent) fixtures on the road and also along the road wherever these are relevant to achieve simulation results of a reasonably good fidelity. & Yes & Yes\\
	\midrule
	\textsf{Dynamic environment models (traffic)} & The VTT is able to model the dynamic aspects of the environment, wherever relevant, and to the level of detail as required for specific test cases. This primarily includes behavioural models of other traffic participants (actors). & Yes & Yes \\
	\midrule
	\textsf{Dynamic environment models (sensory conditions)} & The VTT is able to additionally model the sensory conditions of the environment such as lighting and weather (e.g., precipitation, wind). & \textcolor{red}{No} & Yes \\
	\midrule
	\textsf{Sensing \& perception models (objects)} & The VTT is able to realistically model the detection and perception of other traffic participants (actors) as objects with dynamic properties (such as position, velocity, orientation, and optionally, their type), based on simulation ground truth and the real-world AV capability and limitations. This is required to achieve an approximate virtual representation of the AV’s limitations in object perception, detection and tracking, including but not limited to detection range, handling of occlusions and tracking loss. & Yes & Yes \\
	\midrule
	\textsf{Sensing \& perception models (raw sensors)} & The VTT is able to accurately model the sensors used in the actual AV and their raw sensor data output that the ADS can consume. 
	If this is not feasible, an object-based approximation of the AV capability and limitations in object perception, detection and tracking can be used as an alternative – please refer to the requirement item Sensing \& perception models (objects).
	& \textcolor{red}{No} & Yes \\
	\midrule
	\textsf{Vehicle dynamics models} & The VTT is able to faithfully reproduce the actual dynamics of the AV during its operation. Simple kinematic models are not acceptable. & Yes & Yes  \\
	\midrule
	\textsf{External controllability} & The VTT (either by itself, or in any special-purpose configurations and/or co-simulations if necessary) is able to configure the ADS to disengage from controlling the motion of the virtual vehicle model and place in a neutral gear, wherever required for the purpose of conducting special verification tests, such as coast down tests\footnote{For a generic description of coast down tests and other special tests for vehicle dynamics, refer: Záhorský, Jakub. Master Thesis. "\href{https://dspace.cvut.cz/bitstream/handle/10467/80015/F2-DP-2018-Zahorsky-Jakub-Thesis_Final.pdf}{Development of a virtual car model and subsequent physical validation}”. Czech Technical University in Prague. 2019.}  for vehicle dynamics fidelity check. \cite{Zahorsky2019} Evidence of such tests (e.g., test reports) may be requested during the assessment. & Yes & Yes \\
	\midrule
	\textsf{Automation of simulation runs} & The VTT is able to automatically run a pool of test cases, at least in a batch (sequential) execution mode. The creation of parametrized test cases in an appropriate executable format (compatible with the VTT) such as Scenario XML or OpenSCENARIO or any other equivalent format, may be done manually. However, the batched execution of test cases and logging of test results are expected to be automated, in order to help achieve an efficient and robust virtual testing process. & Yes & Yes\\
	\midrule
	\textsf{Automated logging of test results} & The VTT is able to automatically log (with additional post-processing, if needed) the simulation results in a prescribed results data format as .CSV files.  & Yes & Yes \\
	\bottomrule
	
\end{longtable}

\newpage

\section{FAQ}
\label{sec:faq}
	

\begin{description}[style=nextline]
	
	\item[What is the scope of virtual testing expected for assessment?] \hfill \\ 
	\begin{itemize}		
		\item The virtual tests are mainly intended to check the AV behavioral safety.
		\begin{itemize}
			\item The representativeness and fidelity of VTT will be checked using physical on-circuit tests.
			\item Basic vehicle dynamics, as required for good representativeness, may also be checked
		\end{itemize}

		\item We expect Software in the Loop (SiL) testing setup with actual ADS software integrated with the Virtual Testing Toolchain (VTT)
		
		\item The actual ADS software has to be in the simulation loop in its entirety.
		\\
		However, the below exceptions could be applicable, in case of assessments of AVs where safety driver is allowed to be onboard the AV and is available as a human fallback:
		\begin{itemize}
			\item ADS Sensors and Perception components may be bypassed, and replaced with an object list based on virtual ground truth + error models (if any)
			\item ADS Localization function may not be activated, and may be replaced with location updates from virtual simulation ground-truth
		\end{itemize}
		\item Model in the Loop (MiL) testing is not sufficient and is not acceptable.
		\item The Virtual testing test cases are, by default, expected to be executed only on a virtual model of a test track where corresponding physical tests can also be conducted for a fidelity check. 
		However, exceptions may be made based on the vehicle type, ODD, context, and any other deployment aspects.
	\end{itemize}

	\item[What are the key requirements for the Virtual testing toolchain (VTT)?] \hfill \\
	
	For details, refer to the requirements listed under Section~\ref{sec:req_vtt} of this document. A summary can be found below:	
	\item The capability to implement scripted scenarios with some pre-defined triggers for activating/deactivating special behavior of actors (other road users) in the recommended virtual static environment (e.g., the test track)
	\begin{itemize}
		\item test cases containing variations and parameters of the scenarios are be provided as a document. Optionally, this may also be provided as an executable scenario files, in a mature and open scenario description format such as OpenSCENARIO\textsuperscript{\textregistered}, or a format compatible with applicant's \ac{vtt} if this is feasible to the asesssor
		\item a reference virtual static environment (e.g., test track) may be provided as an OpenDRIVE\textsuperscript{\textregistered} file.
	\end{itemize}

	\item The capability to log some detailed data (in .csv format) as a time-series of the dynamic state of the vehicle under test (VUT) and other actors in each scene (time-synchronized), for each test case
	\begin{itemize}
		\item Details of the results data format is available in the Appendix section~\ref{app:ssec:results_format} of this document.
	\end{itemize}

	\item Test automation (test execution) and logging capabilities
	\begin{itemize}
		\item For assessment, typically at least 10 runs of same test case are required. There can be several 10s of test cases, depending on each application and ODD. Therefore, automation is very important.
	\end{itemize}
	\vfill 

	\item[Is there any specific tool(s) recommended for the independent assessment?] \hfill \\ 
	\begin{itemize}		
		\item There is no particular simulation tool mandated or recommended per se.
		\begin{itemize}
			\item Any good tools may be used - provided good fidelity of the software-in-the-loop testing system is ensured, after integration with ADS and considering the ODD.
			\item Ultimately, the choice of virtual testing toolchain (VTT) is to be made by the AV developer.
		\end{itemize}
		\item AV developer can select/configure/re-engineer appropriate tool(s) 
		\begin{itemize}
			\item The tools should be chosen such that they integrate well with the ADS and can meet the purpose of virtually verifying/validating the behaviour of ADS/AV in its ODD, 
			\item The tools should offer a reasonably good fidelity such that it can be considered as representative of the real-world system.
		\end{itemize}
		\item The VTT requirements described in this document under Section~\ref{sec:req_vtt} are helpful for applicant to make this decision. Furthermore, compliance to these requirements are expected to be stated by the applicant, within the application form submitted for the independent assessment by the assessor (such as CETRAN in Singapore).
					
	\end{itemize}

\end{description}

\cleardoublepage
\section*{Acronyms}
\addcontentsline{toc}{section}{Acronyms}
\begin{acronym}[AAAAAAAA]
	\acro{ads}[ADS]{Automated Driving System}
	\acrodefplural{ads}[ADSs]{Automated Driving Systems}
	\acroindefinite{ads}{an}{an}
	\acro{asil}[ASIL]{Automotive Safety Integrity Level}
	\acroindefinite{asil}{an}{an}
	\acro{av}[AV]{Autonomous Vehicle}
	\acrodefplural{av}[AVs]{Autonomous Vehicles}
	\acroindefinite{av}{an}{an}
	\acro{aesv}[AESV]{Autonomous Environmental Service Vehicle}
	\acrodefplural{aesv}[AESVs]{Autonomous Environmental Servicing Vehicles}
	\acroindefinite{aesv}{an}{an}
%
	\acro{cetran}[CETRAN]{Centre of Excellence for Testing \& Research of AVs - NTU}
	\acro{cg}[CoG]{Centre of Gravity}
	\acro{ddt}[DDT]{Dynamic Driving Task}
	\acro{fusa}[FuSa]{Functional Safety}
	\acro{gnss}[GNSS]{Global Navigation Satellite System}
	\acro{hara}[HARA]{Hazard Analysis and Risk Assessment}
	\acro{hil}[HiL]{Hardware-in-the Loop Testing}
	\acro{hmi}[HMI]{Human Machine Interface}
	\acro{ins}[INS]{Inertial Navigation System}
	\acro{kpi}[KPI]{Key Performance Indicator}	
	\acro{lta}[LTA]{Land Transport Authority}		
	\acro{m1}[M1]{Milestone 1}
	\acroindefinite{m1}{a}{an}	
	\acro{m2}[M2]{Milestone 2}	
	\acroindefinite{m2}{a}{an}	
	\acro{m3}[M3]{Milestone 3}
	\acroindefinite{m3}{a}{an}
	\acro{mrc}[MRC]{Minimal Risk Condition}
	\acroindefinite{mrc}{a}{an}
	\acro{mrm}[MRM]{Minimal Risk Maneuver}
	\acroindefinite{mrm}{a}{an}	
	\acro{mil}[MiL]{Model-in-the Loop Testing}
	\acro{nda}[NDA]{Non-Disclosure Agreement}	
	\acro{ntu}[NTU]{Nanyang Technological University}
	\acro{odd}[ODD]{Operational Design Domain}
	\acroindefinite{odd}{an}{an}
	\acrodefplural{odd}[ODD]{Operational Design Domains}
	\acro{oedr}[OEDR]{Object and Event Detection and Response}	
	\acro{pmd}[PMD]{Personal Mobility Device}	
	\acro{so}[SO]{Safety Operator}
	\acro{sotif}[SOTIF]{Safety Of The Intended Functionality}
	\acro{sae}[SAE]{Society of Automotive Engineers International}	
	\acro{sil}[SiL]{Software-in-the Loop Testing}
	\acro{ttc}[TTC]{Time to Collision}
	\acro{tsv}[TSV]{Traffic Simulation Vehicle}	
	\acro{vcs}[VCS]{Vehicle Coordinate System}
	\acro{vnv}[V\&V]{Verification and Validation}
	\acro{vru}[VRU]{Vulnerable Road User}	
	\acro{vtt}[VTT]{Virtual Testing Toolchain}
	\acro{vtp}[VTP]{Virtual Testing Process}
	\acro{vv}[VV]{Virtual Validation}
	\acro{vut}[VUT]{Vehicle Under Test}
	\acro{xml}[XML]{Extensible Markup Language}	
\end{acronym}
\cleardoublepage

\cleardoublepage
\printbibliography


\clearpage
\appendix                                               
\appendixpage                                           
\addappheadtotoc                                        
\renewcommand{\thefigure}{\thesection.\arabic{figure}}  
\renewcommand{\thetable}{\thesection.\arabic{table}}    

\section{Example of a scenario and corresponding assessment}
\label{app:sec:scenario_eg}

In this section, a scenario ``Overtaking stopped vehicle'' is selected as an example to demonstrate the simulation assessment methodology described above.


\subsection{Scenario description: VUT overtakes a Stationary TSV}
\label{app:ssec:scen_desc}
Initially, the VUT follows the road. Upon detecting a stationary TSV in front, the VUT begins an overtaking manoeuvre at an appropriate distance before the stationary TSV (change the lane to the right). 
The VUT returns to the original lane after the end of the overtaking manoeuvre.
Fig.~\ref{fig:5.1}

\begin{figure}[ht]
	\centering
	\includegraphics[width=\textwidth]{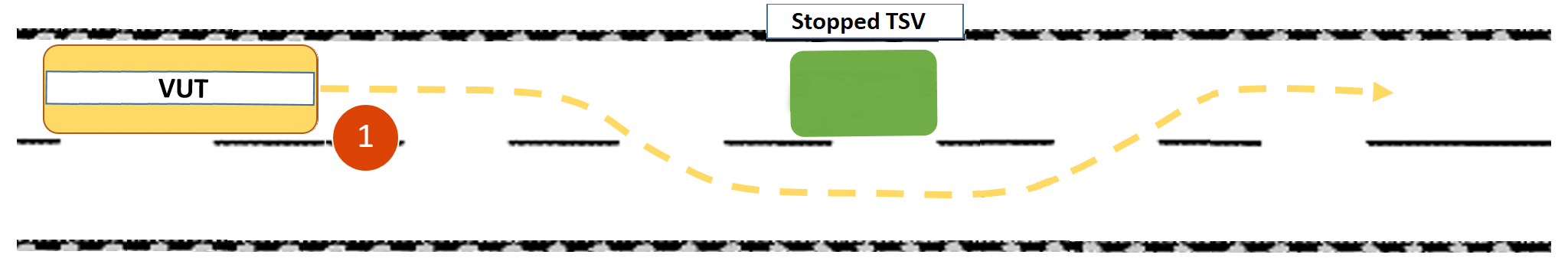}
	\caption{Scenario description diagram}
	\vspace{-1ex}
	\label{fig:5.1}
\end{figure}

\subsection{Test case}
\label{app:ssec:test_case}

This subsection describes the final set of parameters to create a concrete test case. The parameters are carefully chosen considering that the corresponding physical test can be performed on CETRAN test track. 
All relevant parameters are included within the provided OpenDRIVE\textsuperscript{\textregistered} and \textit{ScenarioXML} files.

Note: the illustrations in the figure(s) are not to scale.

\begin{figure}[ht]
	\vspace{-1ex}
	\centering
	\includegraphics[width=\textwidth]{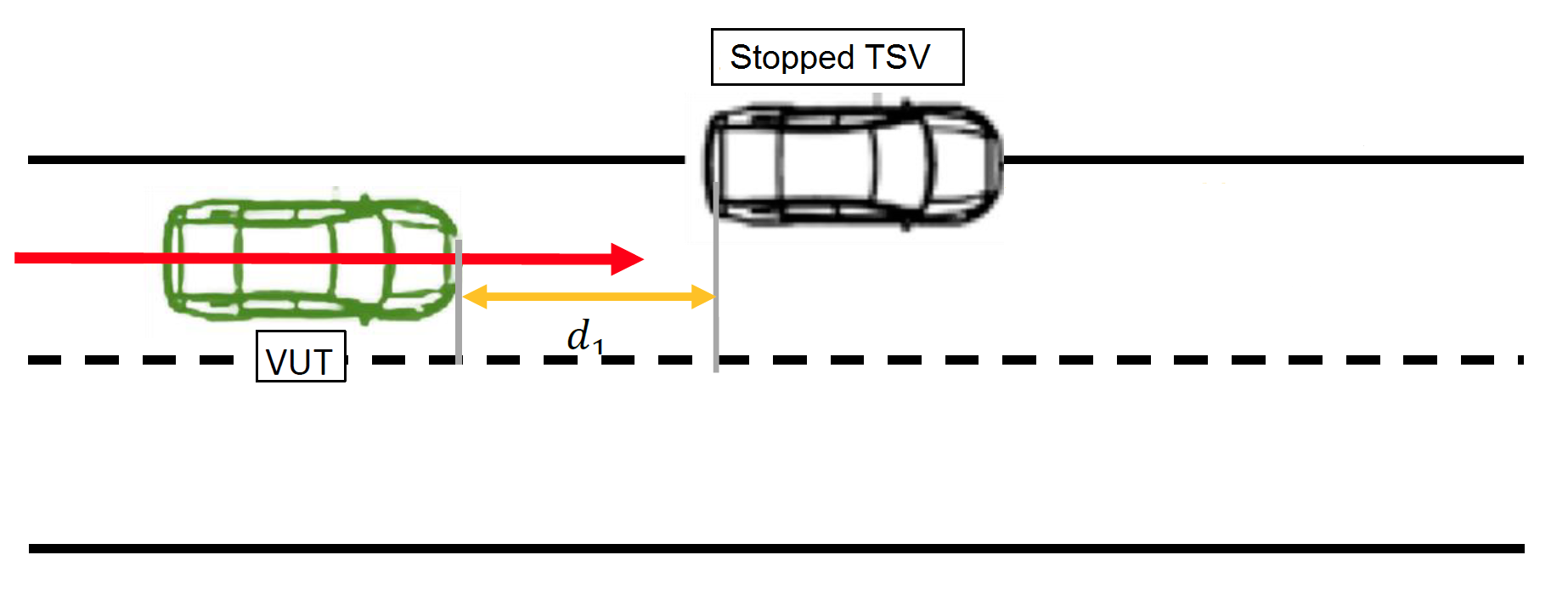}
	\caption{Test case parameters}
	\vspace{-3ex}
	\label{fig:test-case}
\end{figure}

\subsubsection{Parameters}
\begin{itemize}
	\item \textbf{Static environment}
	\begin{itemize}
		\item Location: CETRAN test track
		\item Single carriageway with two driving lanes in each direction; the two lanes are differently separated at various sections by broken centre lines, solid white lines, or solid double white lines.
		\item Width of the lane in each direction is 3.3 m
	\end{itemize}
\end{itemize} 
\begin{itemize}  
	\item \textbf{Dynamic objects}
	\begin{itemize}
		\item VUT
		\begin{itemize}
			\item Vehicle type: \ac{av} - according to actual vehicle to be tested
			\item Initial position: Refer to point A marked on the diagram.
			\item Mission / maneuvers: VUT follows path (A $\rightarrow$ B) as illustrated in the diagram, and also described in the sample \textit{ScenarioXML} file (Ref.\ref{fig:TC_eg_Path})
			\item Velocity: 0 - 40 km/h (the actual speed is up to the ADS to decide)
		\end{itemize}
		\item Stopped TSV
		\begin{itemize}
			\item Vehicle type: 4.4m long class 3 passenger vehicle (e.g, a passenger sedan car)
			\item Initial position: At the bus stop, illegally parked to the left side of the road; left side of the TSV is 0.5m from kerb
			\item Maneuver: None; remain stationary throughout
			\item Velocity: 0 km/h
		\end{itemize}
	\end{itemize}
	\item \textbf{Sensory and environmental conditions}
	\begin{itemize}
		\item Dry surface
		\item No precipitation
		\item Clear daylight and no cloud cover
	\end{itemize}
\end{itemize}

\begin{figure}[ht]
	\centering
	\includegraphics[width=\textwidth]{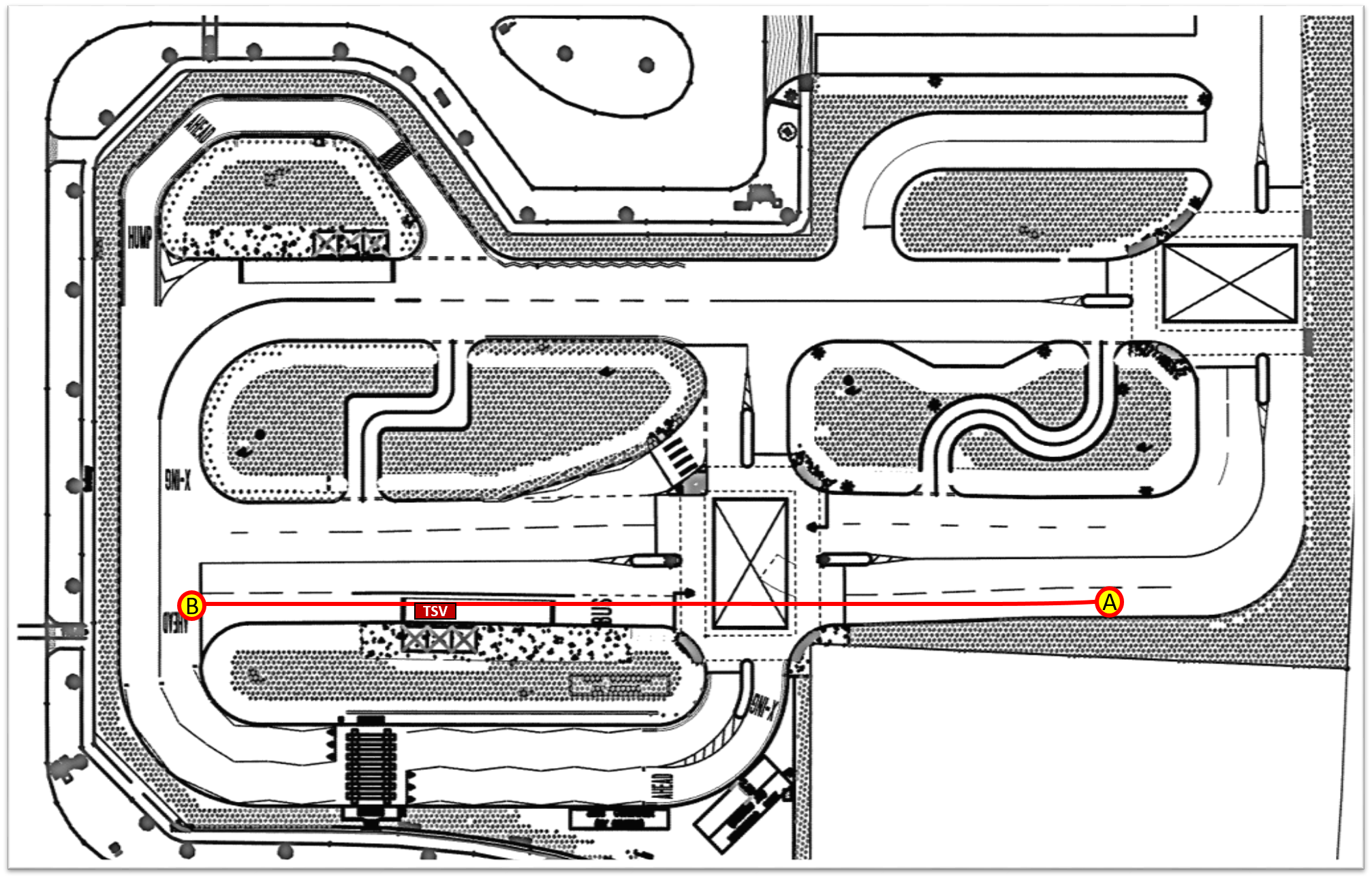}
	\caption{Example Scenario description diagram}
	\label{fig:TC_eg_Path}
\end{figure}

The static environment (road network) is represented in OpenDRIVE\textsuperscript{\textregistered} format in file \texttt{M2-CL4-S-TST-05.xodr}. 
The scenario is represented in the \textit{ScenarioXML} format in file \texttt{M2-CL4-S-TST-05-01.xml}.

\vfill

\subsection{Evaluation}
\label{app:ssec:evaluation}

Once the applicant has submitted their virtual testing results, the assessor has to analyze them as per the metrics discussed in \autoref{sec:eval}.
In this section, we separately analyze and evaluate three different examples of VUT behavior in this scenario. 
However, only one of them is considered safe and acceptable, whereas the other two are deemed unsafe and unacceptable, even though there are some similarities.

The results dataset for the 3 example cases used for this evaluation is available at \url{https://researchdata.ntu.edu.sg/dataverse/cetran_vista}.

\begin{figure*}[htb]
	\centering
	\captionsetup{justification=centering}	
	\subfloat[]{
		\label{fig:TC_eg_actual_behavior_case1}
		\includegraphics[width=0.48\textwidth]{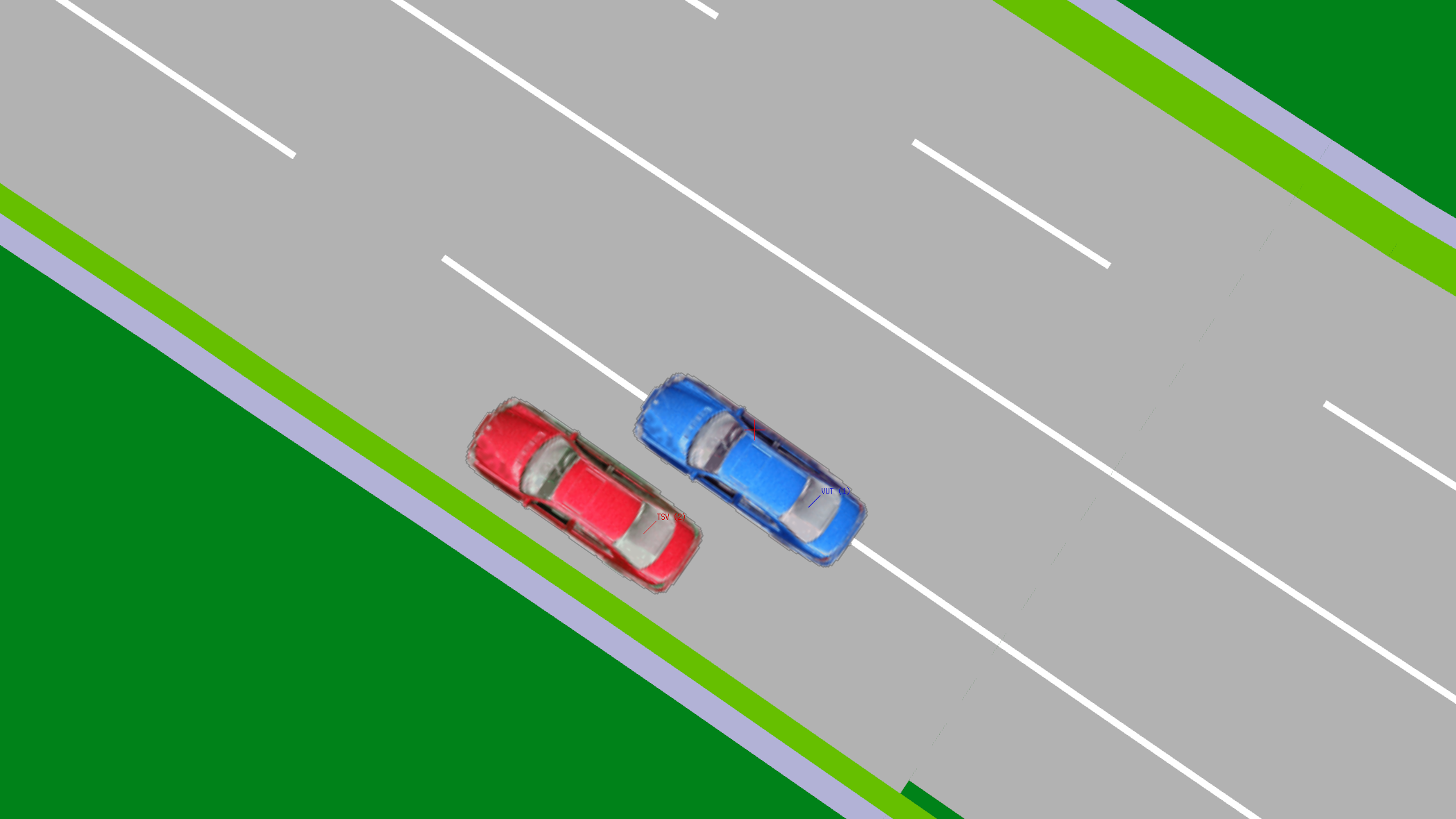}
		\includegraphics[width=0.48\textwidth]{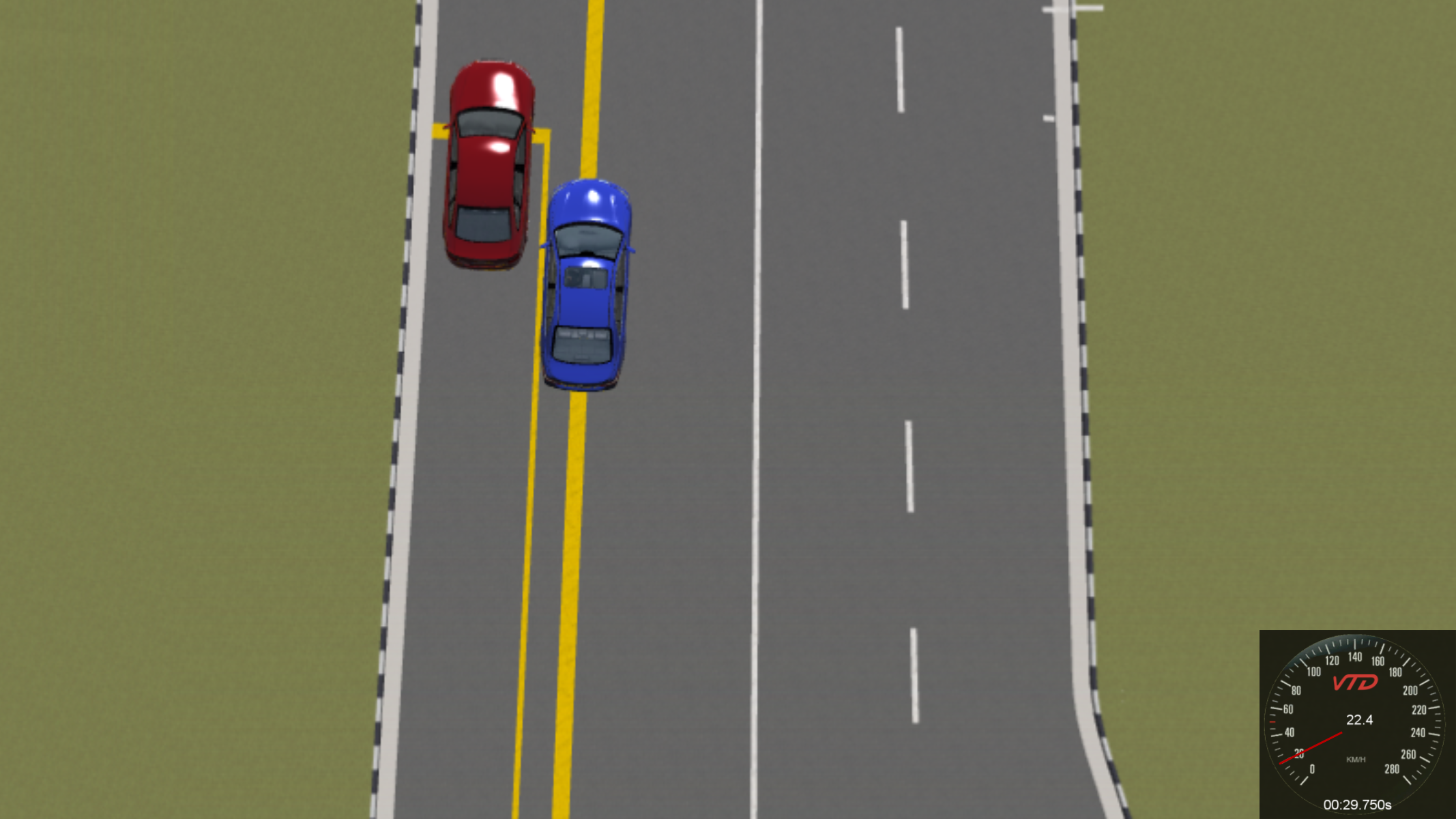}
	}
	\hspace{1pt}
	\subfloat[]{
		\label{fig:TC_eg_actual_behavior_case2}
		\includegraphics[width=0.48\textwidth]{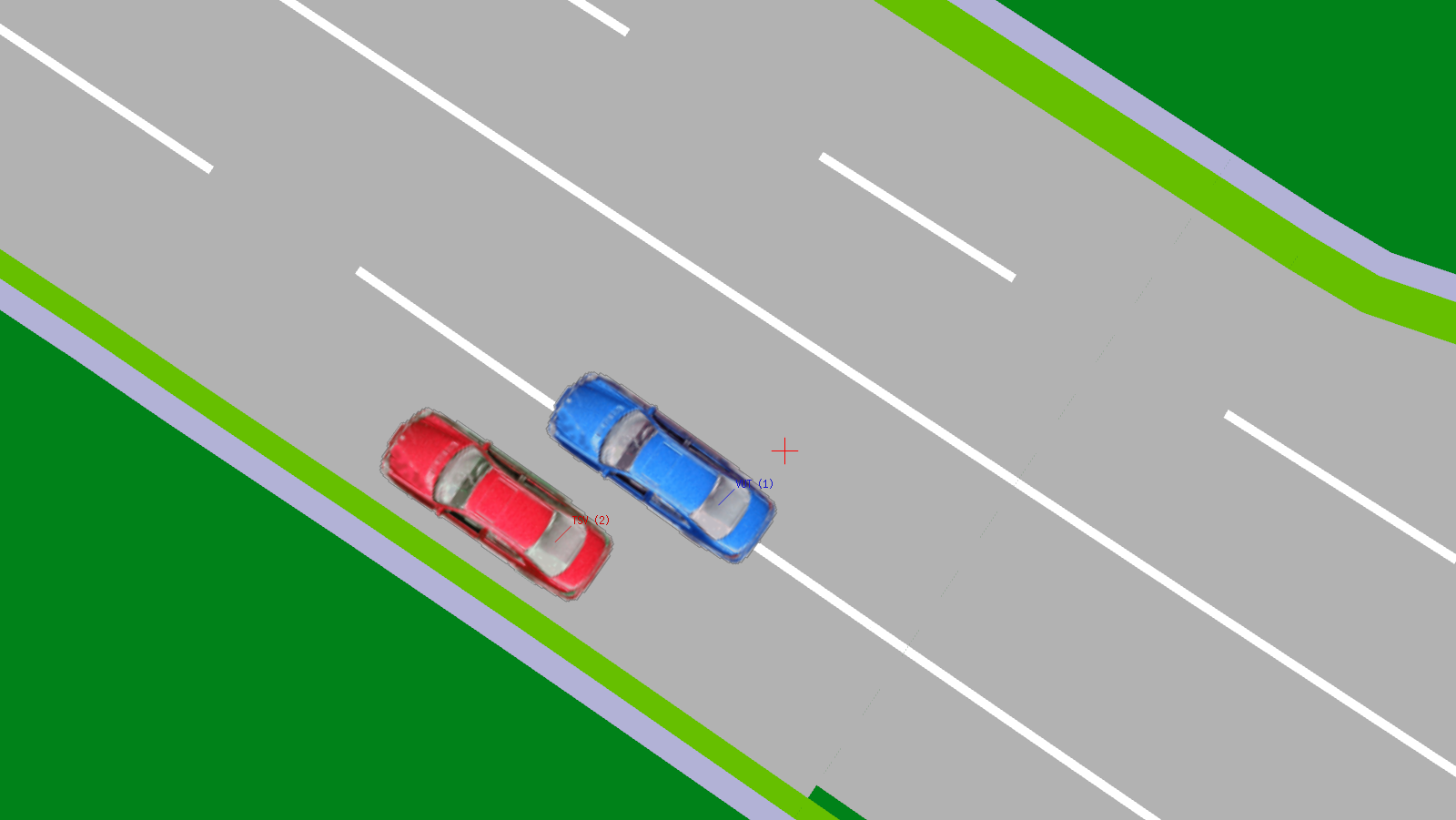}
		\includegraphics[width=0.48\textwidth]{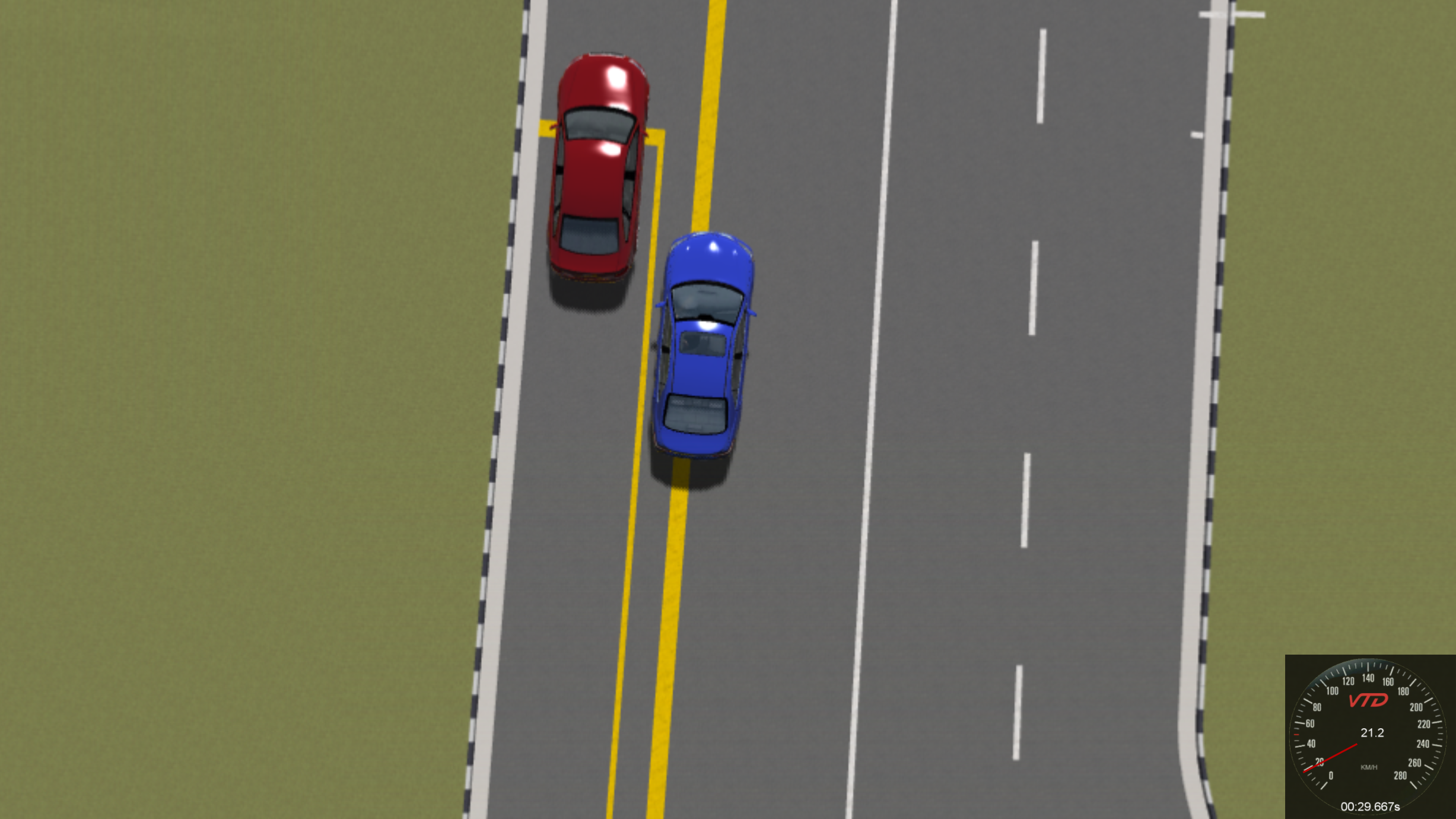}
	}
	\hspace{1pt}
	\subfloat[]{
		\label{fig:TC_eg_actual_behavior_case3}
		\includegraphics[width=0.48\textwidth]{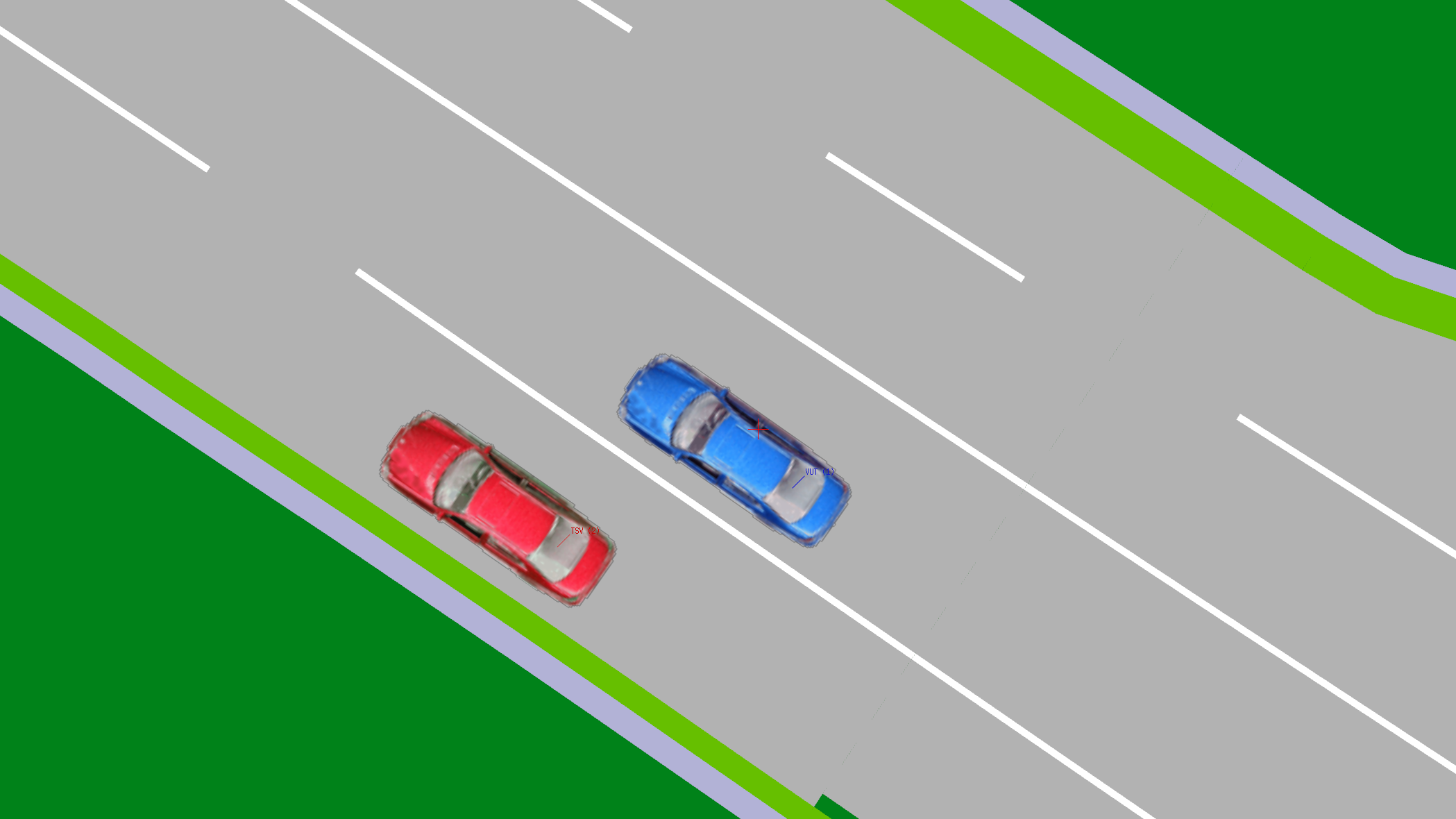}
		\includegraphics[width=0.48\textwidth]{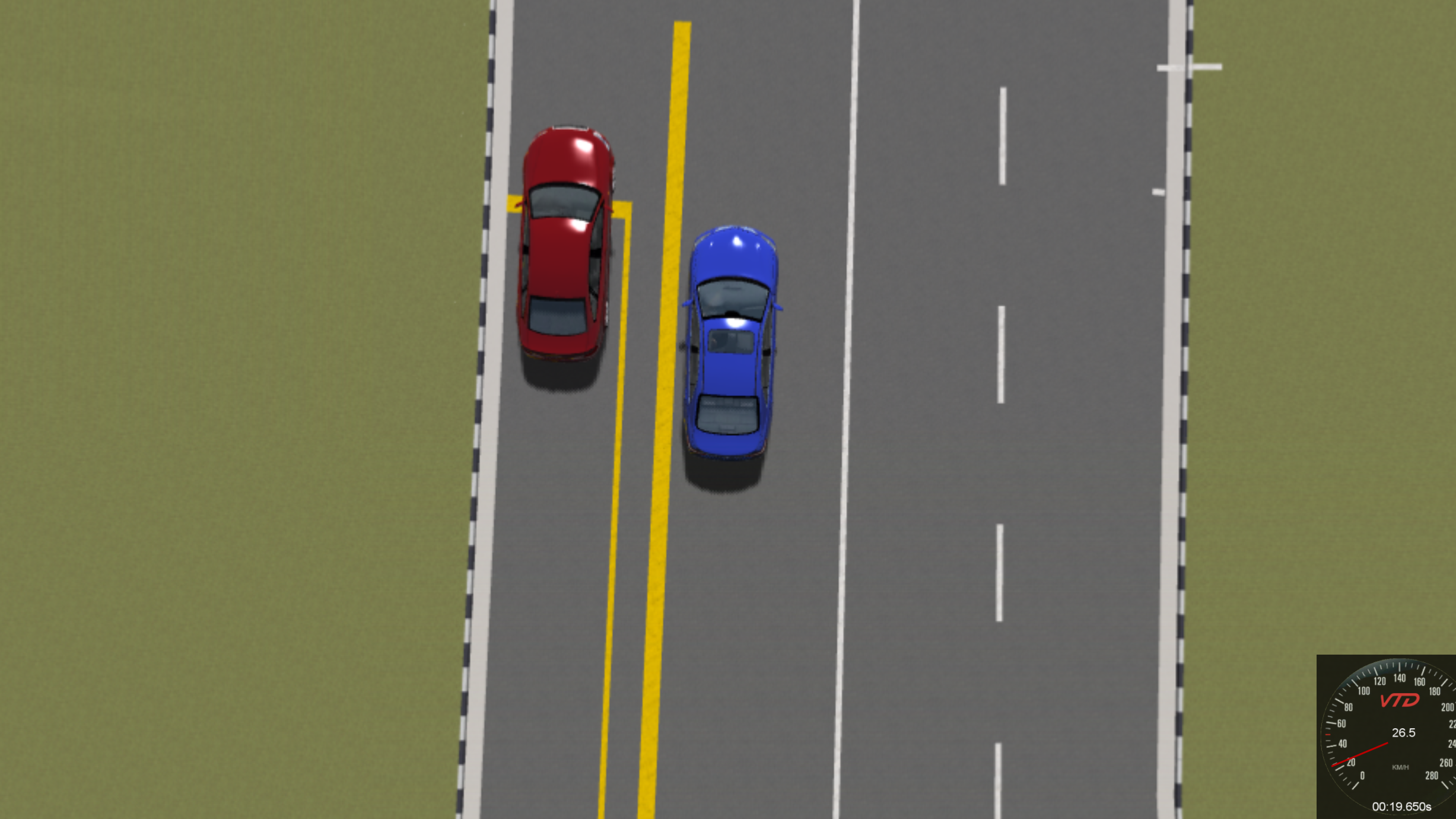}
	}
	
	\caption{
		Actual VUT behavior for the obstacle avoidance scenario, where the blue coloured car is the VUT and the red car is the TSV:
		\\ (a, b) Case 1 and Case 2: Unsafe VUT behavior
		\\ (c) Case 3: Safe VUT behavior
		\\ For each case, Scenario design editor view is on left and rendered 3D scene is on the right
	}
	\label{fig:TC_eg_actual_behavior}
\end{figure*}

\clearpage
\subsubsection{Evaluation example case 1: Unsafe VUT behavior}

As illustrated in \autoref{fig:TC_eg_actual_behavior_case1}, the VUT does not behaves safely in this scenario. 
When avoiding the stopped TSV, the VUT is not able to maintain the sufficient lateral clearance of $1 m$ from the stopped TSV, as recommended in \autoref{sec:eval}, and therefore comes spatially close to each other.
The nearest point is about $0.21 m$ away which is significantly less than the recommended threshold.
Even though technically there is no collision as such, this behavior could be a near miss. It could have potentially resulted in a collision if an occupant was exiting the stopped TSV by opening its right-side door.

Maintaining a sufficient lateral and longitudinal clearance from other road users is an essential aspect of good behavioral safety. 
Therefore, this test case outcome can be deemed as unacceptable and the test is considered a failure.

Furthermore, the decelerations are unacceptably high (e.g., $-8 m/s^2$ or stronger) that can potentially cause harm to occupants, especially if they are standing or not wearing seat belts.
However, this may also be due to poor fidelity of the vehicle dynamics models used in the virtual testing.
Therefore, an independent check of the vehicle dynamics fidelity is necessary.

\begin{figure*}[h]
	\centering
	\includegraphics[scale=0.5]{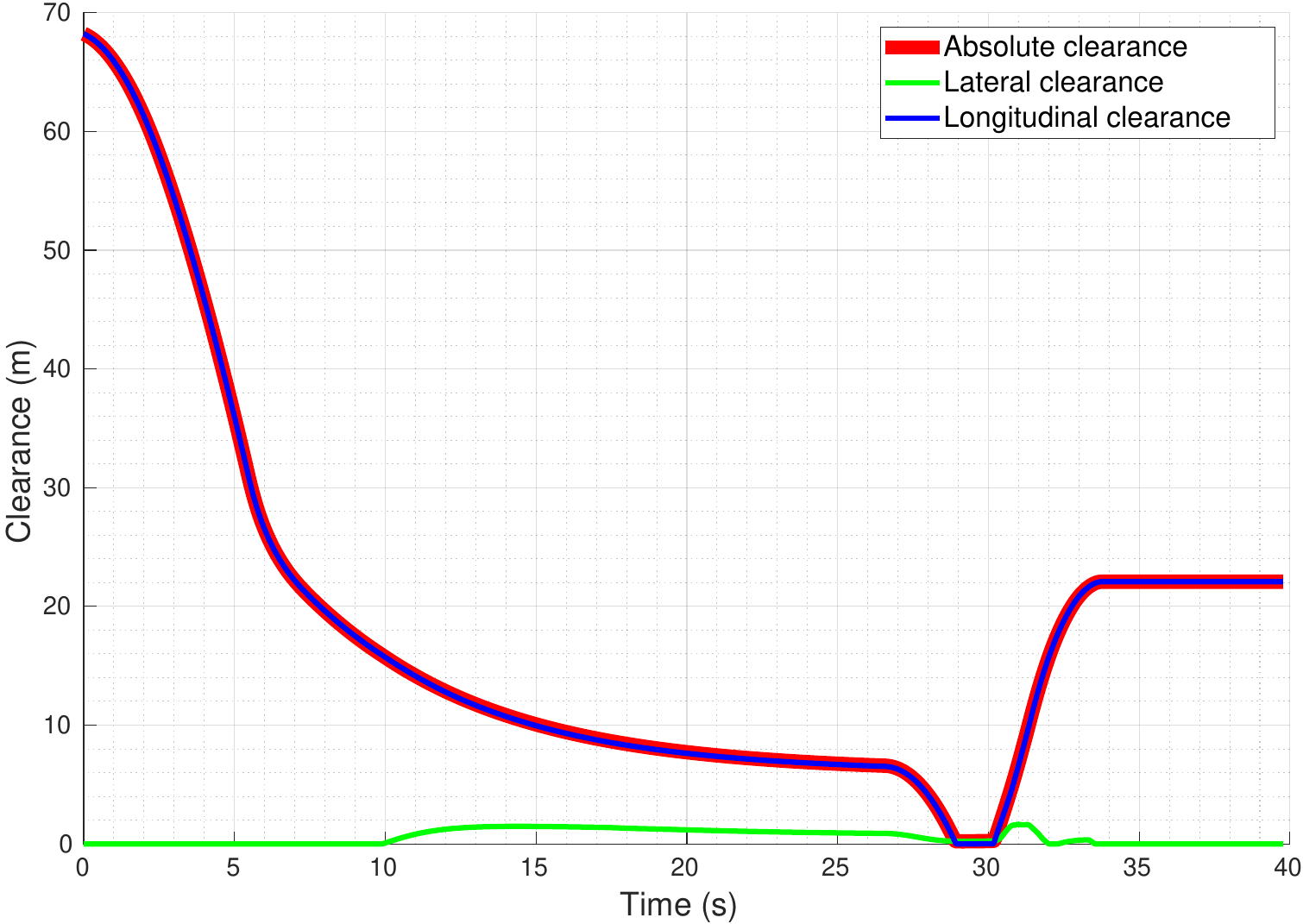}

	\caption{Clearance distances measured between VUT and Actor (stopped vehicle) across the duration of the drive, under case 1: Unsafe VUT behavior}
	\label{fig:TC_clearance_distances_case1}
\end{figure*}

\begin{figure*}[h]
	\centering
	\includegraphics[scale=0.5]{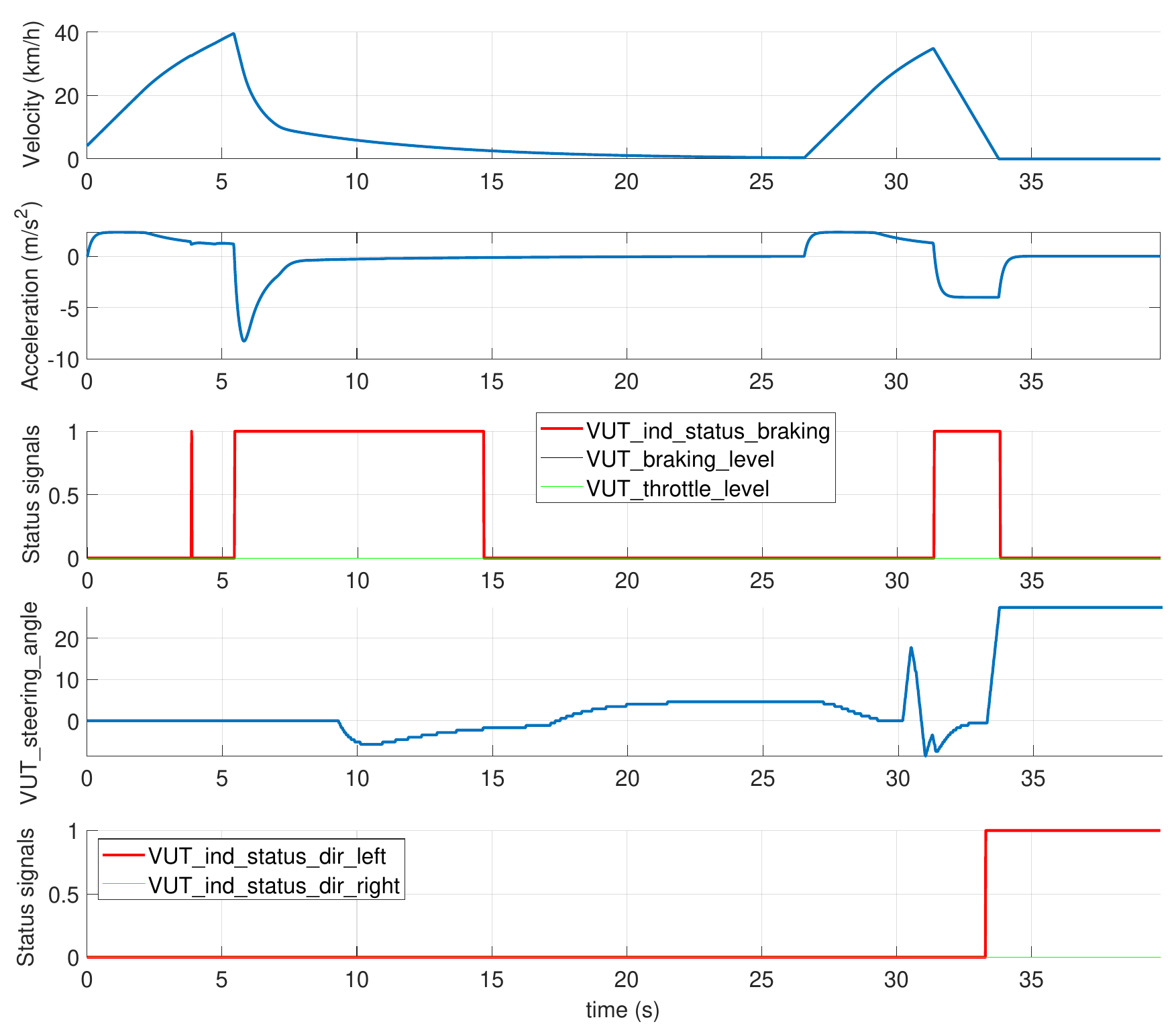}

	\caption{VUT Velocity/acceleration profile with steering angle and various control and status signals, during the obstacle avoidance scenario under case 1: Unsafe VUT behavior}
	\label{fig:TC_eg_velocity_profiles_case1}
\end{figure*}

\begin{figure*}[h]
	\begin{minipage}{\textwidth}
		\centering
		\captionsetup{justification=centering}
		\includegraphics[width=0.9\textwidth]{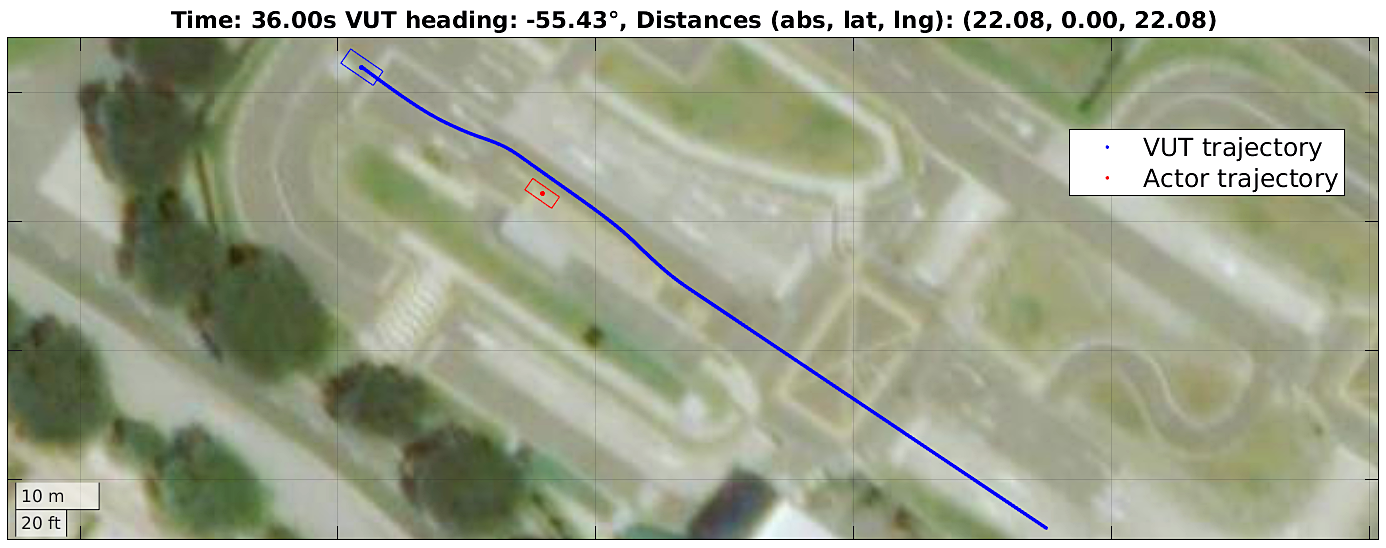}
		
		\caption{VUT trajectory\protect\footnotemark ~for the overtaking scenario, with respect to the actor (i.e., the stationary TSV) under case 1: Unsafe VUT behavior.}
		\label{fig:TC_egtrajectories_case1}
		\footnotetext{\footnotemark[\value{footnote}]Note that the positions depicted in this graph are in World coordinates, as WGS84. The applicant is expected to submit such positional data in WGS84 format and not as Northing and Easting.}		
	\end{minipage}
\end{figure*}

\clearpage

\subsubsection{Evaluation example case 2: Unsafe VUT behavior}

As illustrated in \autoref{fig:TC_eg_actual_behavior_case2}, the VUT maintains a lateral clearance from the stopped TSV. However, while the VUT is completing the overtaking, the lateral clearance goes below the threshold of $1 m$ and even reaches a minimum clearance value of $0.52 m$. 
It is evident that the VUT does maintain a non-zero clearance distance, stops behind the TSV first, proceeds further with an obstacle avoidance maneuver, and returns back to the original lane, and did not result in any collision as such. In spite of this, one can easily ascertain that the VUT still violates the exclusion zone and the lateral clearance thresholds applicable to be considered behaviorally safe in this scenario.

This behavior is unacceptable and therefore this test is also considered a failure.

Furthermore, the decelerations are unacceptably high (e.g., $-8 m/s^2$ or stronger) that can potentially cause harm to occupants, especially if they are standing or not wearing seat belts.
However, this may also be due to poor fidelity of the vehicle dynamics models used in the virtual testing.
Therefore, an independent check of the vehicle dynamics fidelity is necessary by conducting appropriate physical tests.

\begin{figure*}[h]
	\centering
	\includegraphics[scale=0.5]{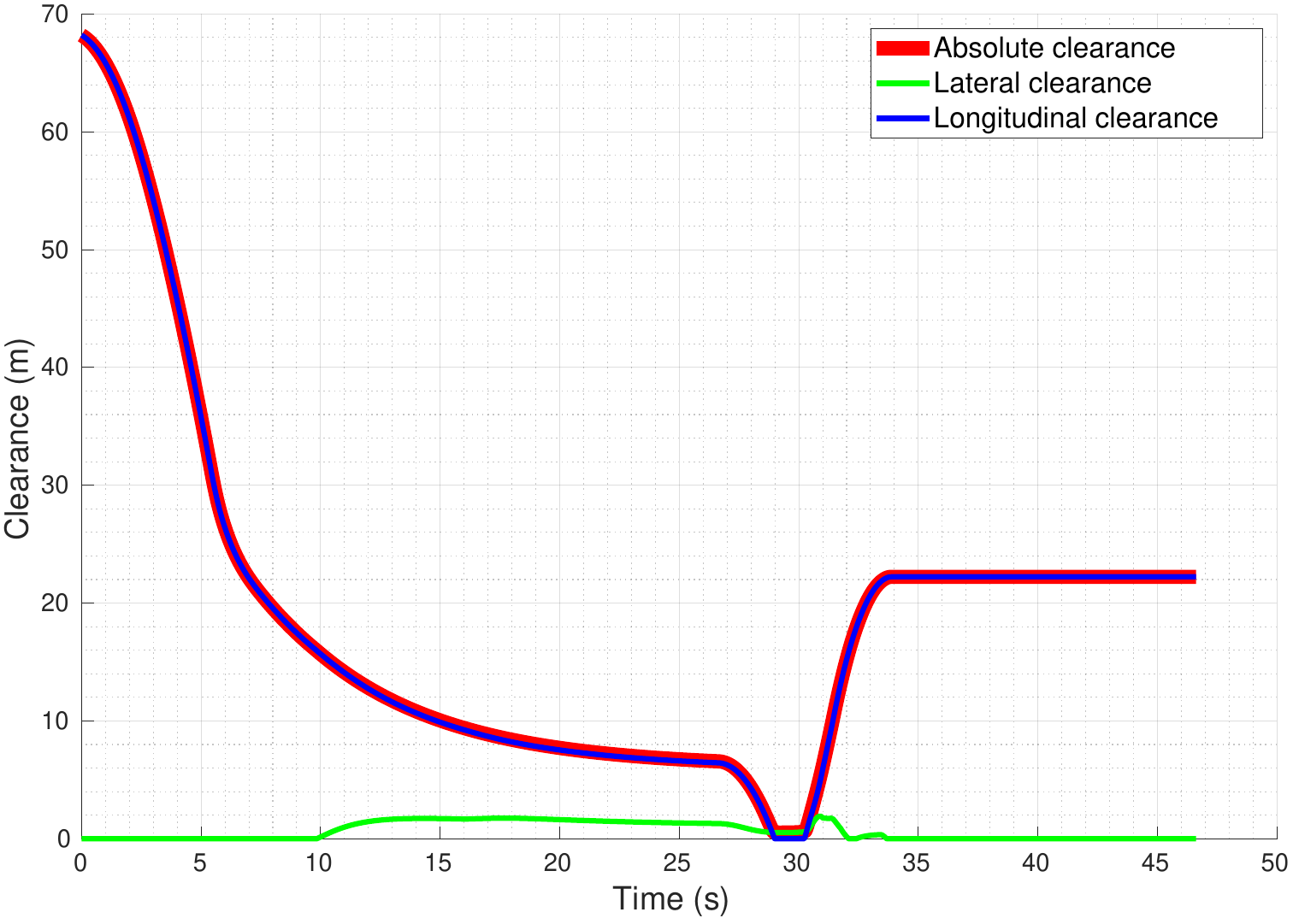}

	\caption{Clearance distances measured between VUT and Actor (stopped vehicle) across the duration of the drive, under case 2: Unsafe VUT behavior}
	\label{fig:TC_clearance_distances_case2}
\end{figure*}

\begin{figure*}[h]
	\centering
	\includegraphics[scale=0.5]{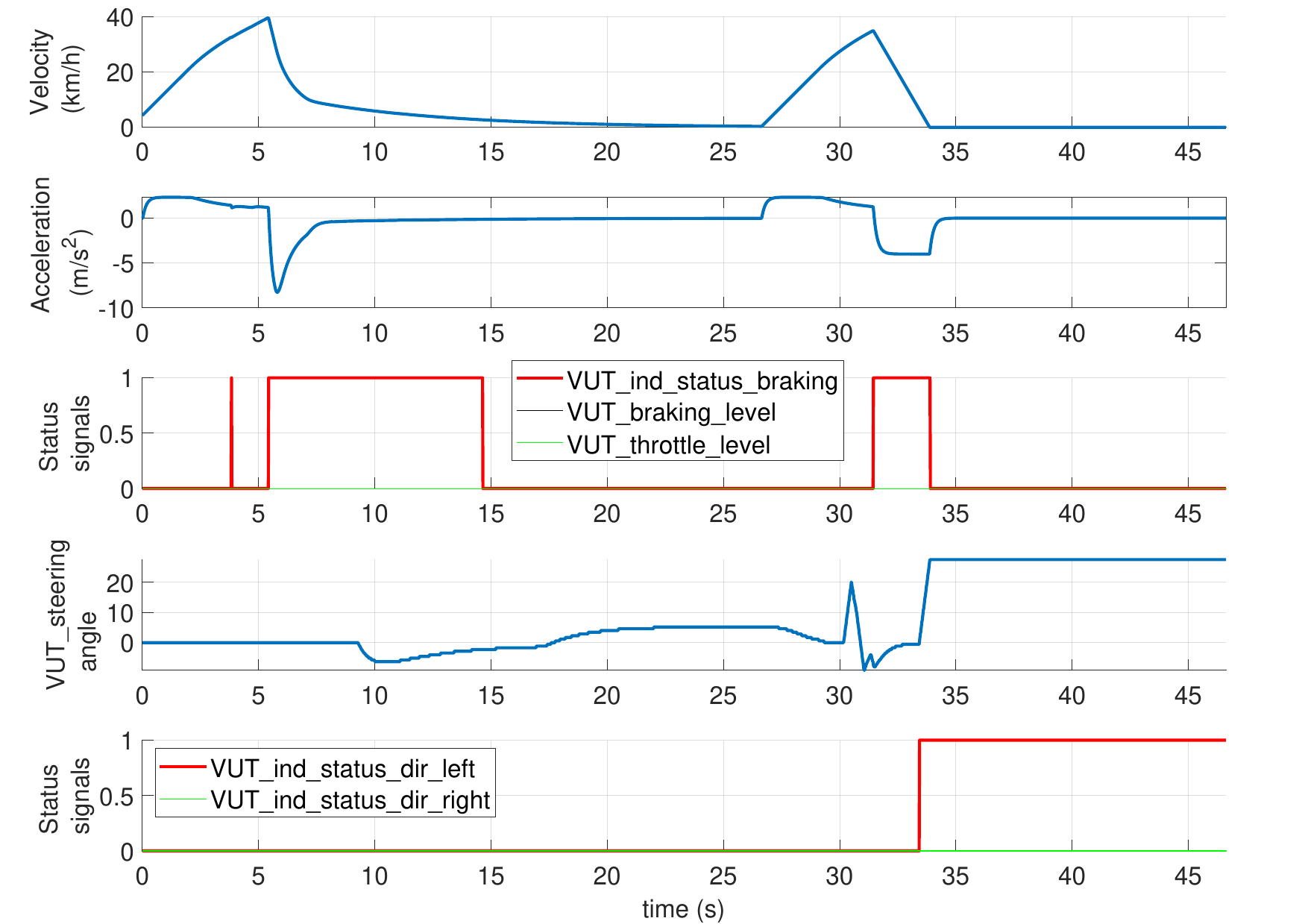}

	\caption{VUT Velocity/acceleration profile with steering angle and various control and status signals, during the overtaking scenario under case 2: Unsafe VUT behavior}
	\label{fig:TC_eg_velocity_profiles_case2}
\end{figure*}

\begin{figure*}[h]
	\begin{minipage}{\textwidth}
		\centering
		\captionsetup{justification=centering}
		\includegraphics[width=0.9\textwidth]{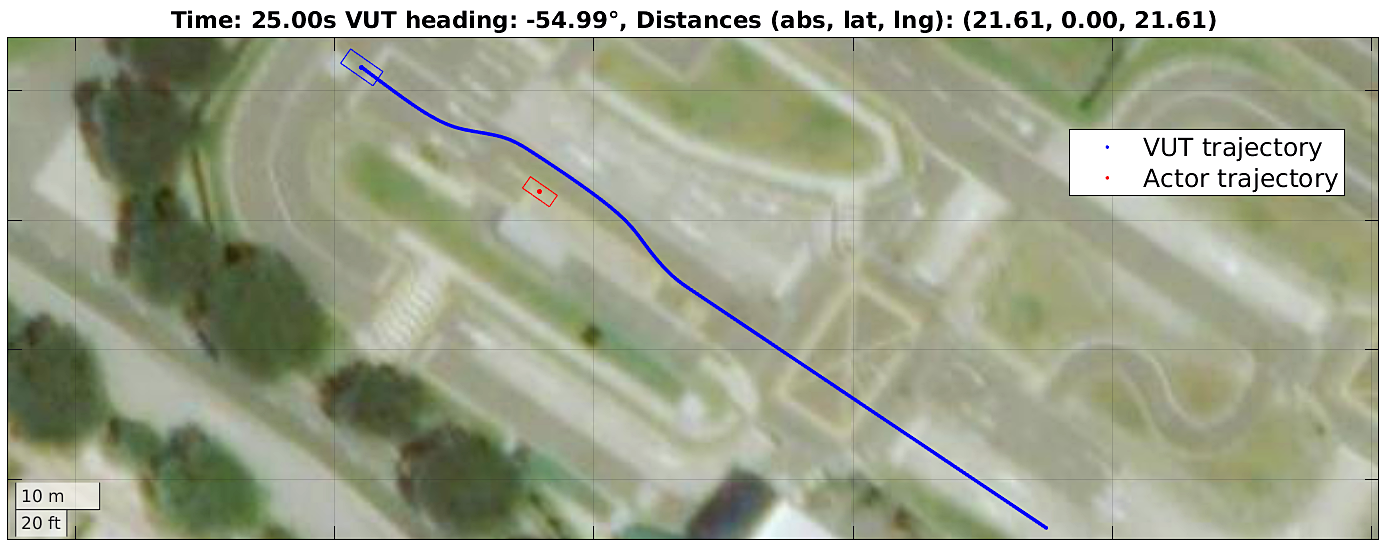}
		
		\caption{VUT trajectory\protect\footnotemark ~for the overtaking scenario, with respect to the actor (i.e., the stationary TSV) under case 2: Unsafe VUT behavior.}
		\label{fig:TC_egtrajectories_case2}
		\footnotetext{\footnotemark[\value{footnote}]Note that the positions depicted in this graph are in World coordinates, as WGS84. The applicant is expected to submit such positional data in WGS84 format and not as Northing and Easting.}		
	\end{minipage}
\end{figure*}

\clearpage

\subsubsection{Evaluation example case 3: Safe VUT behavior}

As illustrated in \autoref{fig:TC_eg_actual_behavior_case3}, the VUT maintains a lateral clearance from the stopped TSV, which is always above the threshold of 1 m, while completing the overtaking.
In this case, as the VUT performs the overtaking maneuver, it slightly crosses into the opposite lane and drives over the lane markings (broken lines). This is justifiable as this is a single carriageway and there is no oncoming traffic.
Furthermore, the VUT maintains a minimum safe clearance distance (the minimum observed being $1.53 m$) as per the exclusion zone defined for this case, and drives off and soon comes back to the original lane. 

In fact, the overtaking VUT behavior is quite similar to the previous example case 2 above.
However, the velocity profile (see \autoref{fig:TC_eg_velocity_profiles_case3}) indicates that the VUT has consistently kept its speed within the allowed road speed limit of 40 kmph (11.11 m/s) at the test track.
There has been no violations and the VUT behavior is generally found acceptable. 
Therefore, this test outcome may be deemed successful.

However, the decelerations are unusually and unacceptably high, e.g., $-8  m/s^2$ ($-0.8 g$) or stronger.
If this were occurring in the real world, this can potentially cause harm to the occupants of the vehicle. This is especially the case if there are occupants who are standing or not wearing seat belts, such as in an automated passenger bus or shuttle.
However, it is also highly likely that such decelerations could have occurred due to the poor fidelity of the vehicle dynamics models implemented and used for the virtual simulation, such as any simplistic kinematic models, e.g., a kinematic bicycle model \cite{}.
Therefore, an independent check of the vehicle dynamics fidelity will usually be necessary before the virtual testing results can be  considered representative of the actual \ac{av} operating in its actual physical \ac{odd}. 
This is relatively more important for heavy vehicles\footnote{such as Class 4 vehicles in Singapore \cite{road_traffic_act}} and for any vehicles that are expected to drive at relatively higher speeds (say, above 40 kmph) and/or with large lateral acceleration (say, above 2.5 $m/s^2$ or $0.5g$). 
Therefore, on a case-by-case basis, depending on the characteristics of the \ac{av} and its \ac{odd}, a demonstration of the applicant's on-road simulations may be requested to obtain more confidence on the fidelity and validity of the \ac{vtt} and the internal \ac{vnv} process adopted by the applicant.

\begin{figure*}[h]
	\centering
	\includegraphics[scale=0.5]{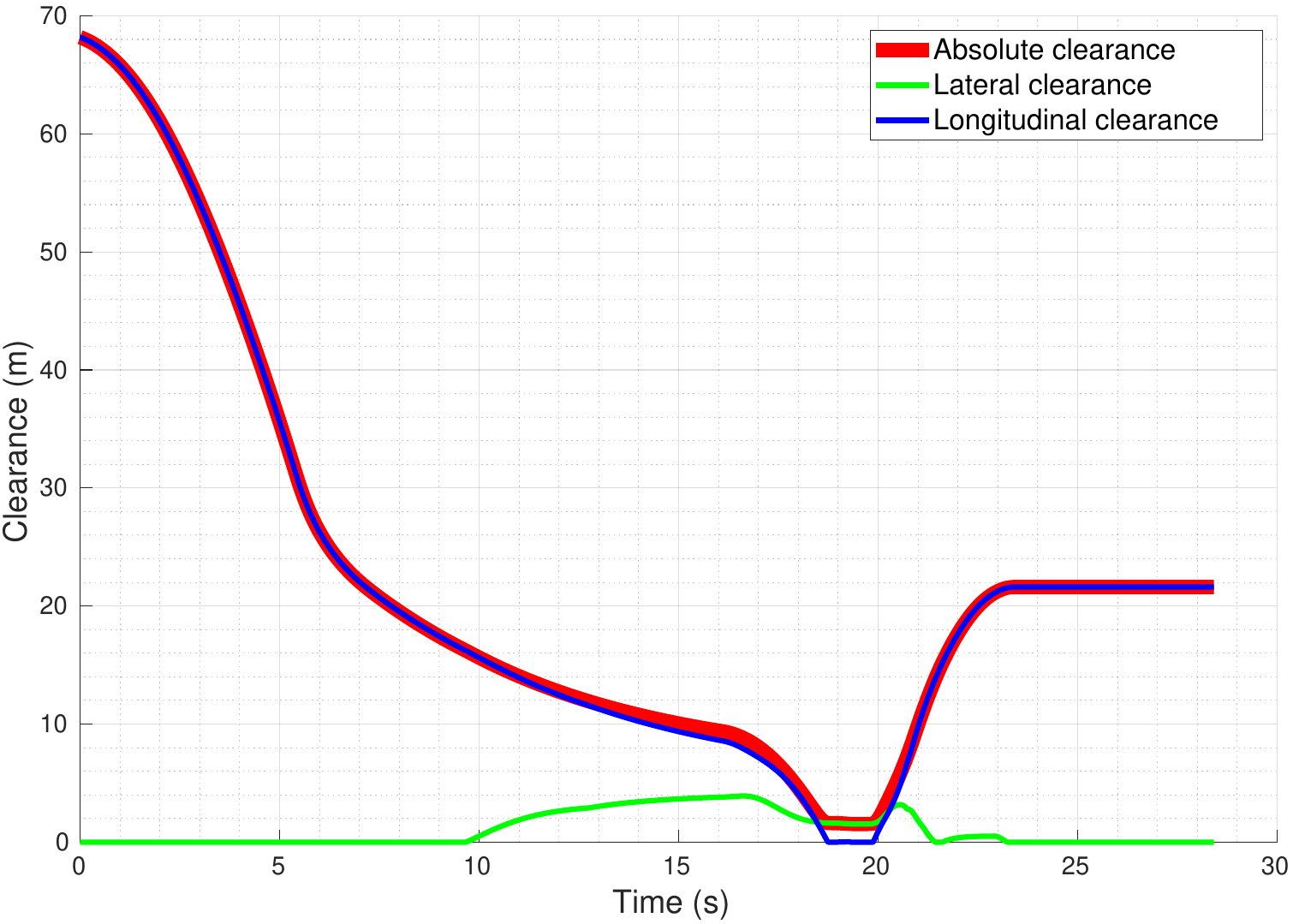}

	\caption{Clearance distances measured between VUT and Actor (stopped vehicle) across the duration of the drive, under case 3: Safe VUT behavior}
	\label{fig:TC_clearance_distances_case3}
\end{figure*}

\begin{figure*}[h]
	\centering
	\includegraphics[scale=0.5]{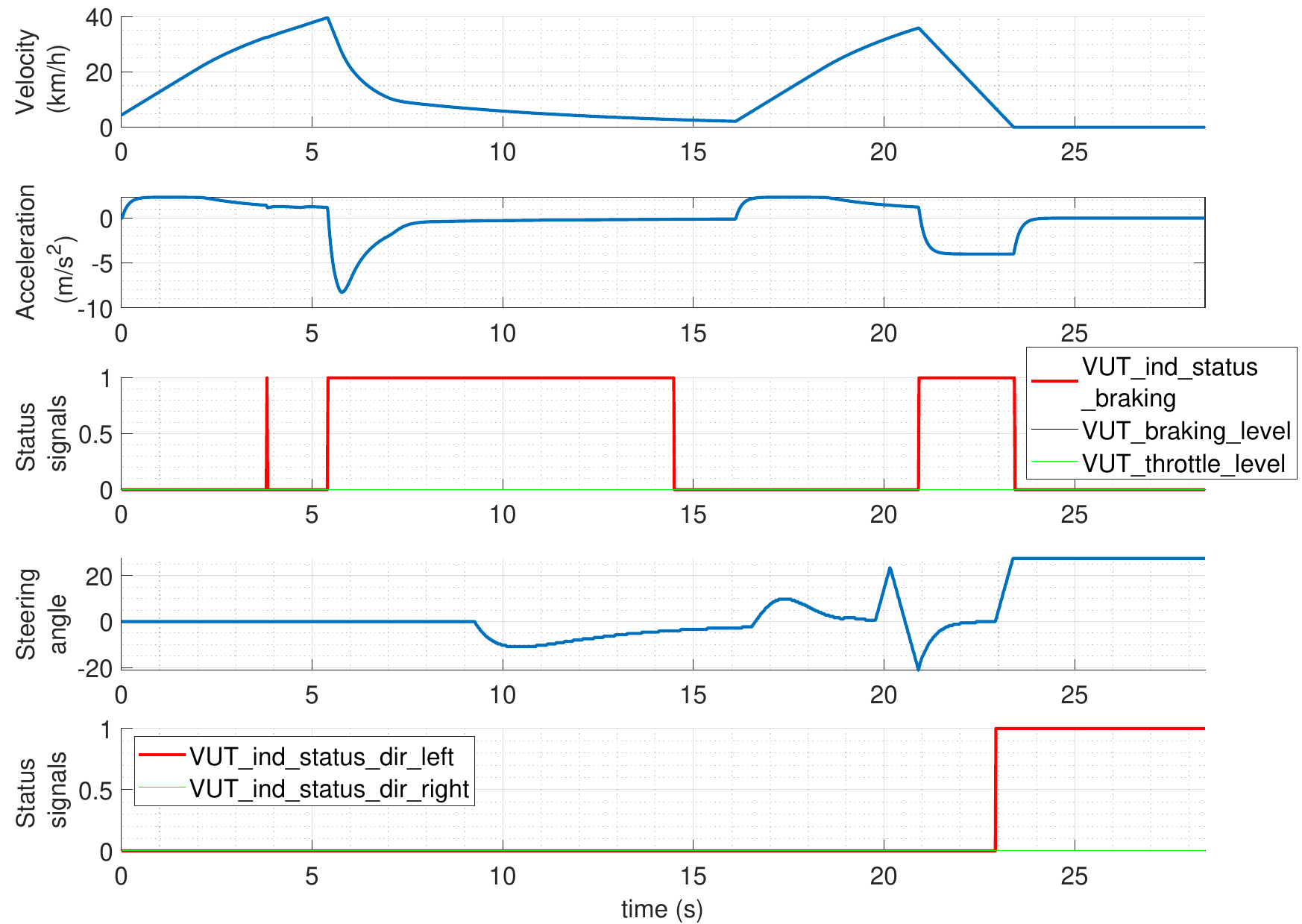}

	\caption{VUT Velocity/acceleration profile with steering angle and various control and status signals, during the overtaking scenario under case 3: Safe VUT behavior}
	\label{fig:TC_eg_velocity_profiles_case3}
\end{figure*}

\begin{figure*}[h]
	\begin{minipage}{\textwidth}
		\centering
		\captionsetup{justification=centering}
		\includegraphics[width=0.9\textwidth]{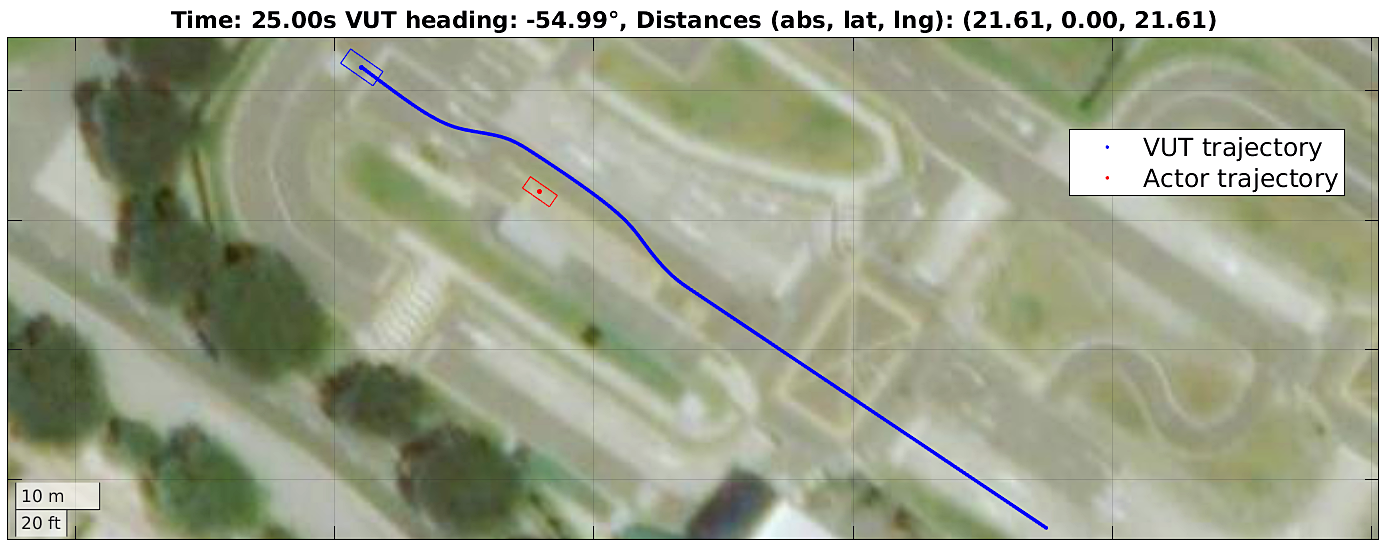}
		
		\caption{VUT trajectory\protect\footnotemark ~for the overtaking scenario, with respect to the actor (i.e., the stationary TSV) under case 3: Safe VUT behavior.}
		\label{fig:TC_egtrajectories_case3}

		\footnotetext{\footnotemark[\value{footnote}]Note that the positions depicted in this graph are in World coordinates, as WGS84. The applicant is expected to submit such positional data in WGS84 format and not as Northing and Easting.}		
	\end{minipage}
\end{figure*}

\clearpage

\section{ViSTA data format for recording virtual testing results}
\label{app:ssec:results_format}

The virtual simulation results are expected to be logged in the ViSTA\footnote{ViSTA stands for \textbf{Vi}rtual \textbf{S}cenario-based \textbf{T}esting of \textbf{A}utonomous Vehicles. A part of the project that focused on scenario-generation was demonstrated at the 2021 IEEE AV Test Challenge as part of the AITest 2021 conference. For details, refer the full paper at: \url{https://arxiv.org/abs/2109.02529} } (Virtual Scenario-based Testing of Autonomous Vehicles) results data format, as \texttt{.csv} files.

The\texttt{.csv} file shall contain various output parameters that expresses the state of the \ac{vut} and actors (such as \ac{vru} or \ac{tsv}) at each simulation step. 
This is most likely obtained through post-processing after the virtual tests are executed, and may therefore be done in two steps, namely, logging of virtual tests results into native proprietary data formats supported by the \ac{vtt}, and then, post-processing these to generate the ViSTA results data format for independent safety assessment.

This section describes the details of the contents, organization, sequence, data types and units applicable to the results data to be recorded.

\subsection{CSV file contents organization}
\label{app:sec:csv_file_contents}
In a nutshell, the \texttt{.csv} file shall contain the following:

\begin{itemize}
	\item current \textbf{timestamp} (starting at $t = 0.000 s$) and corresponding \textbf{simulation step} number
	\item current state of the \ac{vut}
	\item current state of the relevant stationary \textbf{obstacles} around the VUT
	\item current state of the relevant dynamic \textbf{actors} around the VUT	
	\item current state of the relevant \textbf{traffic light controllers} around the VUT
\end{itemize}

The type of entities involved in the data are illustrated in \ref{fig:vista_results_conceptual}.

The contents of the \texttt{.csv} files as a time series, are also illustrated in \ref{fig:vista_results_conceptual_timeseries}.

\begin{figure}[h]
	\centering
	\includegraphics[width=0.8\linewidth]{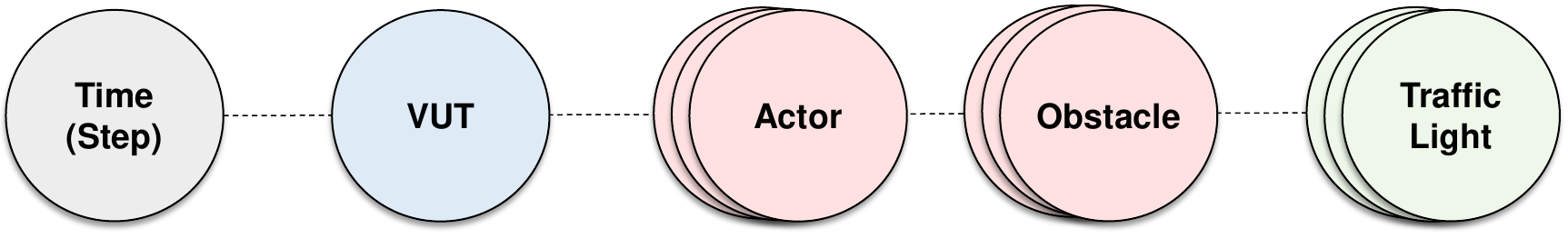}
	\caption{Illustration of how the virtual simulation results data is to be organized in the ViSTA results format}
	\label{fig:vista_results_conceptual}
\end{figure}

\begin{figure}[h!]
	\centering
	\includegraphics[width=0.9\linewidth]{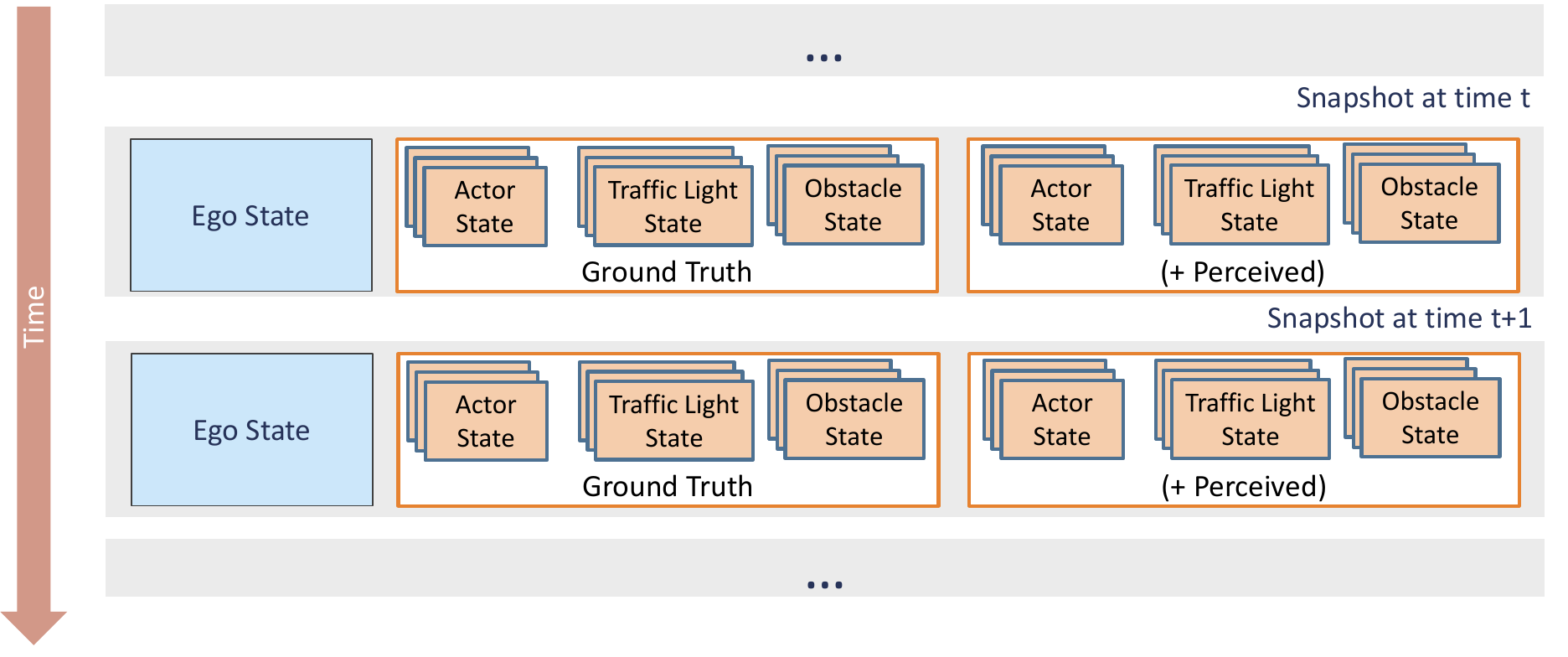}
	\caption{Illustration of how data is to organized in the ViSTA results format}
	\label{fig:vista_results_conceptual_timeseries}
\end{figure}

Note that stationary \textbf{obstacles} can have a simplified representation as described below. However, if applicant desires so, they may be also described as \textbf{actors} but with the corresponding motion parameters set to stationary values.

Note also that the ground truth on the state of all the actors, obstacles and traffic controllers (individual traffic lights need not be logged) that are relevant to the scenario or test case must be recorded at all times.
The only exception is for \textit{perceived} state, which really depends on the capabilities of the \ac{vut} for sensing, detection, and perception.

The \texttt{.csv} results data can be provided in either of the two formats as described in below sub-sections:

\begin{itemize}
	\item Single flat file format
	\item Distributed file format
\end{itemize}

\subsubsection{Single flat file format}
\label{app:ssec:csv_file_contents_flat}

In this case, all the virtual test results data is to be logged into a single flat \texttt{.csv} file.
The file is to be named after per the test case and run number (e.g., \texttt{M2-CL4-S-TST-05-01\_r09.csv}).

The first row shall contain a header with the name of each output parameter.
The results data shall be placed from the second row onwards and organized such that there is one row per simulation step and ordered as per monotonically increasing time.
The first data row shall always contain the \textit{initial} values when the simulation for the particular test case starts execution (time = 0).

The various tables listed under Section~\ref{app:ssec:data_types_units} illustrate the result parameters and their corresponding units. 
For a flat \texttt{.csv} file, the fields from different tables can be compiled together while preserving the sequence and avoiding any duplication of common fields such as \texttt{Time} or \texttt{Step\_number}.

A sample file is available in the file \texttt{results\_M2-CL4-S-TST-05-01\_r09.csv} that corresponds to the test case \texttt{M2-CL4-S-TST-05-01} and run number 09 (containing results from the 9\textsuperscript{th} time this simulation test case was executed under the same conditions).
Each row consists of a time-stamp, result parameters indicating the current state of the VUT, followed by result parameters indicating the current state of each actor included in the scenario (e.g., VRU, Oncoming TSV or Stopped TSV) with a unique id (as mentioned in the respective test case). 
Results parameters pertaining to a particular actor shall be included in \textit{consecutive} columns and shall always start with the column \textsf{Actor\_Id} and end with \textsf{Actor\_TTC}.
For each additional actor or obstacle or traffic light, the same fields (the full group) can be repeated with the same names, but with a new unique id value under \textsf{Actor\_Id} or \textsf{Obst\_Id} or or \textsf{Traffic\_Ctrl\_Id}.

Note that usage of this format may require duplication of data for some simulation steps, especially if the data sources for individual fields are operating at different data sample rates.

\subsubsection{Distributed file format}
\label{app:ssec:csv_file_contents_distributed}

Alternatively, instead of storing all the test results into a single file, they may also be distributed into \textit{separate} dedicated \texttt{.csv} \textit{files} with content and file names as listed below. 

These \texttt{.csv} files must be placed in an own \textit{folder} for each test case run, and the folder must be named in the similar style as above, viz., as per the test case and run number (e.g., \texttt{M2-CL4-S-TST-05-01\_r09}).
However, it is important that each entry across multiple files, must be synchronized with a simulation step (and implicitly, a common clock).
This is most likely achieved through post-processing after the virtual tests are executed.
In several practical systems, this distributed format can be more efficient in terms of data recording and management, compared to the flat file format.

Reference data files using this distributed \texttt{.csv} file format, for the 3 example cases discussed earlier in Section~\ref{app:ssec:evaluation}, is available at \url{https://researchdata.ntu.edu.sg/dataverse/cetran_vista}.

\begin{itemize}
	\item \texttt{VUT\_status.csv} - columns regarding VUT only, together with time and simulation step. The simulation step must match with those of the actors and traffic lights. 
	The detailed data format, types and units for applicable result parameters (fields) are specified in \ref{tab:results_csv_format_VUT}.
	\item \texttt{Environment\_actors\_true.csv} - contains columns regarding each actor (and optionally, each obstacle represented as a full actor) based on the ground truth, which is easy to acquire in virtual simulations. 
	The information on the state of each actor in a given time instant (simulation step) can be logged in a new line. However, simulation time and the current simulation step has also to be recorded and provided for each row. The simulation step for each row must match with that of the VUT data csv file (\texttt{VUT\_status.csv}).
	The detailed data format, types and units for applicable result parameters (fields) are specified in \ref{tab:results_csv_format_actors_true}.
	\item \texttt{Environment\_actors\_perceived.csv}	- contains columns regarding each actor (and optionally, each obstacle represented as a full actor) based on the actual sending and perception capabilities on the AV. 
	Each actor info can be logged in a new line. However, simulation time and the current simulation step has also to be recorded and provided for each row. The simulation step for each row must match with that of the VUT data csv file (\texttt{VUT\_status.csv}).
	The detailed data format, types and units for applicable result parameters (fields) are specified in \ref{tab:results_csv_format_actors_perceived}.	
	\item \texttt{Environment\_obstacles\_true.csv} - contains columns regarding each obstacle based on the ground truth, which is easy to acquire in virtual simulations. 
	The information on the state of each obstacle in a given time instant (simulation step) can be logged in a new line. However, simulation time and the current simulation step has also to be recorded and provided for each row. The simulation step for each row must match with that of the VUT data csv file (\texttt{VUT\_status.csv}).
	The detailed data format, types and units for applicable result parameters (fields) are specified in \ref{tab:results_csv_format_obstacles_true}.	
	\item \texttt{Environment\_obstacles\_perceived.csv}	- contains columns regarding each obstacle based on the actual sending and perception capabilities on the AV. 
	Each obstacle info can be logged in a new line. However, simulation time and the current simulation step has also to be recorded and provided for each row. The simulation step for each row must match with that of the VUT data csv file (\texttt{VUT\_status.csv}).
	The detailed data format, types and units for applicable result parameters (fields) are specified in \ref{tab:results_csv_format_obstacles_perceived}.		
	\item \texttt{TrafficLight\_true.csv} - 
	contains columns regarding the true status of each traffic light controller. 
	The values may be obtained from virtual simulation ground truth under SiL/HiL testing configurations. They may also be obtained from DSRC or equivalent V2X communication, in case of HiL or ViL testing configurations, where this is feasible and available.
	Each traffic light controller phase/timing info can be logged in a new line. 
	However, simulation time and the current simulation step has also to be recorded and provided for each row. The simulation step for each row must match with that of the VUT data csv file (\texttt{VUT\_status.csv}).
	The detailed data format, types and units for applicable result parameters (fields) are specified in \ref{tab:results_csv_format_traffic_lights_true}.		
	\item \texttt{TrafficLight\_perceived.csv} - 
	contains columns regarding the status of each traffic light controller (or individual traffic light) based on the actual sending and perception capabilities on the AV.	
	Since the ADS may not necessarily be able to perceive the state of the traffic light controller as well, the state of individual traffic lights, as and when they are perceived, may be sufficient. Even in this case, the result field names must be same as defined. E.g., \texttt{Traffic\_Ctrl\_Id} may be used to provide an id of an individual perceived traffic light also. This is because, in this particular case, we may assume that each perceived traffic light may have an exclusive traffic light controller, even though in reality a single controller may control multiple traffic lights.
	
	Each traffic light controller (or traffic light) phase/timing info can be logged in a new line. However, simulation time and the current simulation step has also to be recorded and provided for each row. The simulation step for each row must match with that of the VUT data csv file (\texttt{VUT\_status.csv}).
	The detailed data format, types and units for applicable result parameters (fields) are specified in \ref{tab:results_csv_format_traffic_lights_perceived}.		

\end{itemize}

\subsection{Data types, units, representation and sequence}
\label{app:ssec:data_types_units}

In this section, we describe the data types, units and sequence in which the result fields must be placed within the .csv file(s).
For each result field, the field name and unit are expected to be exactly the same as specified in the Tables below.
Furthermore, the column titled \textbf{Mandatory} indicates whether the field is necessary or optional.

The result fields may be created in either the flat .csv file format (in which case the result fields from different Tables can be compiled together while preserving the sequence and avoiding any duplication of common fields such as \texttt{Time} or \texttt{Step\_number}) or distributed into separate .csv files as described in Sections \ref{app:ssec:csv_file_contents_flat} and \ref{app:ssec:csv_file_contents_distributed} respectively.

The specific data format details for VUT and the environment (both for the true and perceived values) such as actors, obstacles and traffic lights can be found in the following Tables \ref{tab:results_csv_format_VUT}, \ref{tab:results_csv_format_actors_true}, \ref{tab:results_csv_format_actors_perceived}, \ref{tab:results_csv_format_obstacles_true}, \ref{tab:results_csv_format_obstacles_perceived}, \ref{tab:results_csv_format_traffic_lights_true} and \ref{tab:results_csv_format_traffic_lights_perceived} respectively.

\subsubsection{Use of Actor information fields to represent Obstacle information}

Note that if the applicant desires so, they can use the Actor information fields to store the information of stationary objects (such as static obstacles) as well, in order to have a common and consistent representation.
However, in such cases, some of the actor-specific fields may be superfluous (e.g., \textsf{Actor\_acc\_lat} or \textsf{Actor\_vel\_abs}) but they will still have to be recorded nonetheless to ensure consistent representation. 
The obstacle type (\textsf{Obst\_type}) field may be replaced with the Actor type (\textsf{Actor\_type}), while using the same values as were originally specified for obstacles, such as construction\_cones=100.

\subsubsection{Positions}
\label{typedef:pos}
The positions are, by default, expected to be recorded in WGS84 world coordinate system for \ac{vut} as well as actors, obstacles or any other objects. 
This is primarily for the ease of conducting a fidelity check by comparing virtual and physical test results.
Alternatively, the  vehicle coordinate system (VCS) as described below may also be used for all of these except the \ac{vut}.

\paragraph{WGS84 World Coordinate system}
\label{typedef:pos_wgs84}
Note the following aspects about the World coordinate system which is expected to be used for recording positions.

\begin{itemize}
	\item The positions are by default, expected to be recorded in a world coordinate frame, in the WGS84 latitude and longitude coordinates (and optionally, the elevation or Z coordinate), for the \ac{vut}, actors, obstacles and any other objects.
	\item The optional world position elevation (Z) values, if specified, must be \textit{always} calculated relative to a common reference plane and is applicable to the \ac{vut}, actors, obstacles and any other objects. For example, it could be the mean sea level (as is a common practice in GIS tools), or the height of common point on the ground as per the map used by the \ac{av}.
	\item Heading angles are always to be recorded in degrees, with $0^{\circ}$ pointing North and measured clock-wise.
	\item Optionally, the vehicle coordinate system (VCS), as described below, may also be used, but this only to describe the relative position of actors and obstacles with respect to the VUT's absolute world position at a given moment in time.
\end{itemize}


\paragraph{Vehicle Coordinate System (VCS)}
\label{typedef:pos_vcs}
\begin{figure}[htb]
	\centering
	\includegraphics[width=0.6\textwidth]{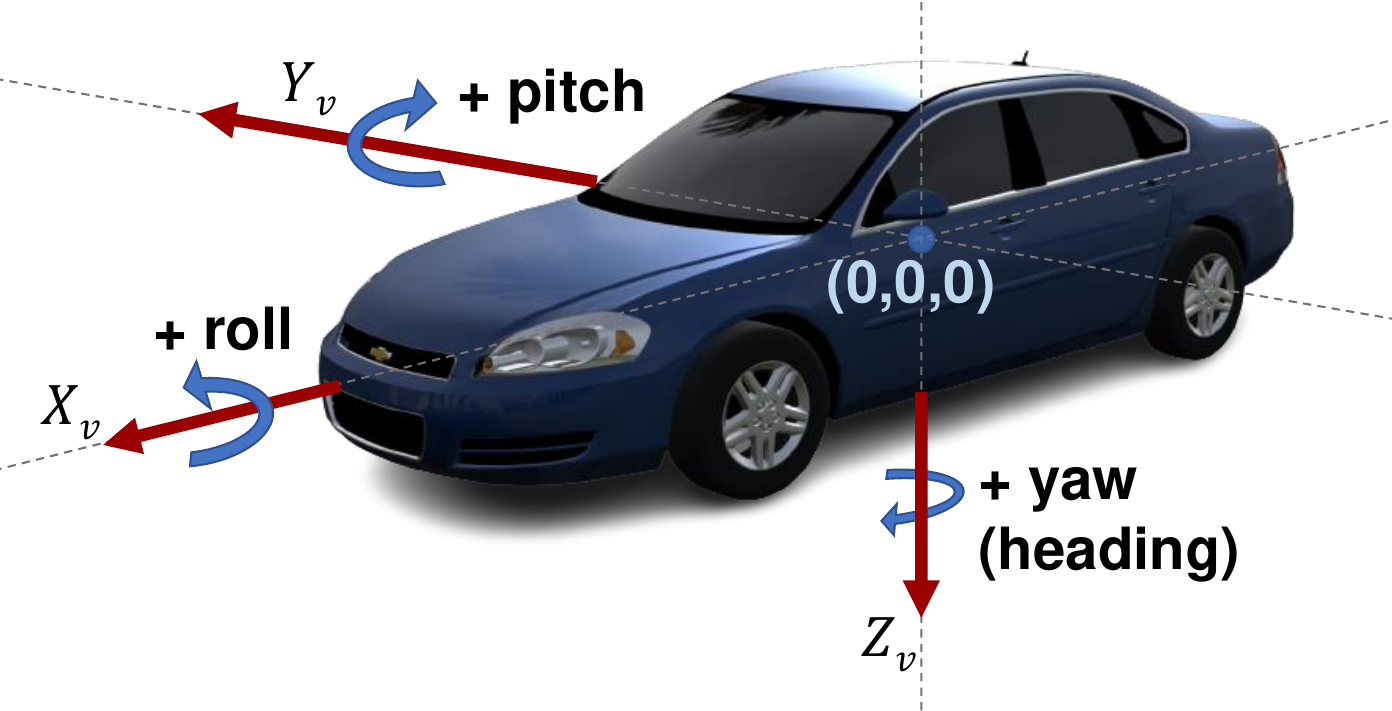}
	\caption{Illustration of the Vehicle Coordinate System (VCS) used for describing the relative position of actors or obstacles with respect to the absolute position of \ac{vut}}
	\label{fig:vcs_illustration}
\end{figure}

The vehicle coordinate system (VCS) may be used to additionally describe the position of actors, obstacles and other objects relative to the position of the \ac{vut}.
The VUT position, by itself, must be \textit{always} described as an absolute position, within the WGS84 world coordinate frame described earlier. 

By default, the positions of such actors or objects should be logged in World coordinates. However, VCS is provided as an alternative option, if the applicant prefers to describe relative positions instead of World position, especially for reasons such as improvements in precision and/or accuracy. 

Key aspects of this vehicle coordinate system are described below:

\begin{itemize}
	\item This is a Z-down coordinate frame, based on the Z-down orientation defined in SAE J670 (2008 edition) standard \cite{sae2008j670}
	\item The origin ($X_v=0, Y_v=0, Z_v=0$) or vehicle reference point \cite{sae2008j670}, shall be placed at the geometric centre of the \ac{vut}, which may be significantly different from the \acf{cg} of the \ac{vut}.
	\item This frame is attached to the \ac{vut} and rotates with it in all three axes with the heading, pitch and roll of the \ac{vut}.
	\item The $X_v$-axis is positively directed forward (longitudinally) from the \ac{vut}, and is parallel to the heading of the \ac{vut}.
	\item The $Y_v$-axis is perpendicular to the $X_v$-axis and is positively directed to the right (laterally) of the \ac{vut}.
	\item The $Z_v$-axis is perpendicular to the $X_v$ and $X_v$ axes, and is positively directed downwards from the \ac{vut}.
	\item The Euler angles, viz., the pitch, roll, and yaw (heading) angles are measured in a clockwise positive fashion, when looking in the positive 
	direction of the $X_v$, $Y_v$, and $Z_v$ axes respectively. 
	\begin{itemize}
		\item Specifically, if one is looking down at the \ac{vut} in a top-down perspective, then the yaw (heading) angle shall be measured as clockwise-positive.
		\item For calculating specific Euler angles, the knowledge of the following axes would also be required (as defined in SAE J670 \cite{sae2008j670})
		\begin{itemize}
			\item the ground-plane parallel Y-axis of the VUT
			\item the ground-plane parallel X-axis of the VUT (or the projection of $X_v$ on ground plane)
			\item the earth-fixed $Z_E$ axis (parallel to the gravitational vector) 
			\item the earth-fixed $X_E$ axis (parallel to ground plane and pointing in the geographic North direction)
		\end{itemize}
	\end{itemize}
	\item The relative position coordinates of the actors, obstacles or other objects have to be calculated in such a way that their own respective axis system be transformed to be aligned with that of the VUT. 
	For example, in the case of a TSV following the VUT on the adjacent right lane on a straight two-lane road, with the TSV's own geometric center placed 5m longitudinally and 4m laterally behind the VUT's vehicle reference point, its true position may be reported as $(-5, 4)$, i.e., $\textsf{Actor\_pos\_true\_x} = -5$ and $\textsf{Actor\_pos\_true\_y} =  4$. 
\end{itemize}

For the purpose of the independent simulation assessment as described in this document, only the X, Y and heading (yaw) values are generally sufficient to be reported in the virtual testing results data.
However, this does not exclude the possibility that there might be exceptional cases where the full information would be required and requested to perform a deeper analysis.

\subsubsection{Velocities and accelerations}
\label{typedef:vel_accln}

The lateral and longitudinal velocities and accelerations are to be recorded based on the object's own local frame of reference, irrespective of whether it is a VUT, actor or obstacle.

\subsubsection{Bounding polygons and Serialized list of positions}
\label{typedef:bpoly_array_position}
\paragraph{\texttt{Array<Position>}}
The \texttt{Array<Position>} data structure can be used to conveniently represent a list of geographic positions such as for bounding polygons or paths within a .csv file, by serializing the list into a single formatted string.
Note the following salient points with regard to this data structure, together with the illustration provided below.

\begin{itemize}
	\item The entire \texttt{Array<Position>} data structure has to be serialized into a single formatted string without commas and recorded into a single csv cell.
	E.g., a polygon with 5 points (positions) can be represented by the following string (which has been split into multiple lines below, only for ease of illustration).
	\item Each \texttt{Position} instance is serialized as \texttt{|lat lng Z|} or \texttt{|X Y Z|}, for WGS84 or VCS coordinates respectively, depending on which coordinate frame is used.
	\item This uses the \texttt{|} symbol to delimit consecutive positions, and a \texttt{whitespace} to delimit the individual latitude and longitude and elevation (Z) values for WGS84 coordinates, or the X and Y and Z values for VCS coordinates as applicable. 
	\item The Z values are, by default, optional for the independent simulation assessment and may therefore be omitted usually (unless if explicitly requested) as illustrated in the below example.		
	
\end{itemize}

\vspace{-2ex}
\begin{center}
	\begin{tabular}{ l }
		\texttt{< 5 }\\
		\texttt{| 103.6957499292194 1.354088453458461} \\
		\texttt{| 103.6956478 1.3540848} \\
		\texttt{| 103.6956508073799 1.354060261353194} \\ 
		\texttt{| 103.6957503367588 1.354064076671374} \\
		\texttt{| 103.6957499292194 1.354088453458461} \\
		\texttt{>} 
	\end{tabular}
\end{center}



\newpage

\begin{adjustwidth}{-2cm}{}
	\setlength\arrayrulewidth{0.01pt}
	\begin{longtable}{|l|c|c|c|}
		\caption{Name and unit of result parameters: VUT state}
		\label{tab:results_csv_format_VUT}
		\\ \hline			
		\bfseries Field & \bfseries Unit & \bfseries Description & \bfseries Mandatory
		\begin{filecontents*}{resultsmetadata/resultsmetadata_VUT.csv}
fieldname,unit,description,mandatory
Time,s,\makecell{current timestamp\\ (Time=0.000s \\at start of simulation)},yes
Step\_number,integer $>=0$,\makecell{current Simulation step\\ (Step\_number=0 \\at start of simulation,\\monotonically increasing)},yes
VUT\_pos\_lat,WGS84 decimal,\makecell{current world position\\ latitude of VUT},yes
VUT\_pos\_lng,WGS84 decimal,\makecell{current world position\\ longitude of VUT},yes
VUT\_pos\_z,$m$,\makecell{estimated world position\\ height above ground (Z) of VUT},yes
VUT\_heading,\makecell{$\deg$ $[0^{\circ},360^{\circ})$},\makecell{current yaw (heading) angle\\ (measured \textit{clockwise} positive\\ about $Z_E$ or gravitational vector, \\from north to the projection of \\ $X_v$ on horizontal ground plane\\ as per SAE J670 \cite{sae2008j670};\\ $0^{\circ} \implies$ north)},yes
VUT\_pitch,\makecell{$\deg$ $[-90^{\circ},90^{\circ}]$},\makecell{current pitch angle\\ (measured \textit{clockwise} positive\\ about the ground-parallel Y-axis,\\ from ground projection of $X_v$\\ to the $X_v$ axis itself,\\ as per SAE J670 \cite{sae2008j670})},no
VUT\_roll,\makecell{$\deg$ $[-180^{\circ},180^{\circ}]$},\makecell{current roll angle\\ (measured \textit{clockwise} positive\\ about the $X_v$ axis\\ from the ground-parallel Y-axis\\ to the $Y_v$ axis of VUT,\\ as per SAE J670 \cite{sae2008j670})},no
VUT\_yaw\_rate,$\deg/s$,yaw rate,yes
VUT\_jerk\_lat,$m/s^3$,lateral jerk,yes
VUT\_jerk\_lng,$m/s^3$,longitudinal jerk,yes
VUT\_accl\_lat,$m/s^2$,lateral acceleration,yes
VUT\_accl\_lng,$m/s^2$,longitudinal acceleration,yes
VUT\_vel\_lat,$m/s$,lateral velocity,no
VUT\_vel\_lng,$m/s$,longitudinal velocity,no
VUT\_vel\_abs,$m/s$,\makecell{absolute velocity\\ (speed)},yes
VUT\_travelled,$m$,\makecell{distance travelled \\from start},yes
VUT\_ind\_st\_dir\_left,\makecell{boolean},indicator light status,yes
VUT\_ind\_st\_dir\_right,\makecell{boolean},indicator light status,yes
VUT\_ind\_st\_hazard,\makecell{boolean},indicator light status,yes
VUT\_ind\_st\_reverse,\makecell{boolean},indicator light status,yes
VUT\_ind\_st\_braking,\makecell{boolean},indicator light status,yes
VUT\_throttle\_level,{\makecell{percentage $[0,100]$}},throttle level,yes
VUT\_braking\_level,{\makecell{percentage $[0,100]$}},braking level,yes
VUT\_steering\_angle,\makecell{$\deg$},steering angle\footnote{Exact vehicle dimensions and steering ratio (\url{https://en.wikipedia.org/wiki/Steering\_ratio}) of the AV may be provided separately to the assessor for simulation assessment, if they are not explicitly mentioned in the application form.},no
{VUT\_steering\_angle\_ \\ percentage},{\makecell{percentage $[0,100]$}},{\makecell{steering angle\footnotemark[\value{footnote}]\\as a percentage}},yes
VUT\_AV\_drive\_status,\makecell{(autonomous=0;\\ manual=1;\\ tele-operation=2)},AV driving status\footnote{If the VUT operation cannot be fully described by these 3 modes alone, then the AV developer is expected to propose additional modes and inform the assessor before providing the simulation results.},yes
{VUT\_special\_operation\_\\status},\makecell{(normal\_\\operation\_mode=0;\\ environmental\_\\service\_mode=1;\\ any\_other\_special\_\\operation\_mode=99)},\makecell{status that indicates \\whether VUT is driving \\normally as any standard \\vehicle, or as per any\\ special operation mode\\ (e.g., the environmental\\ service mode for an AESV)}\footnote{If the VUT has a special operation mode but this status cannot be logged from VTT, then the applicant (AV developer) is expected to propose any alternative and get the agreement from assessor, before submitting the simulation results.},yes
Number\_of\_obstacles\_true,integer $\ge 0$,\makecell{Number of obstacles,\\ based on ground truth},yes
{Number\_of\_obstacles\_\\ perceived},integer $\ge 0$,\makecell{Number of actual\\ perceived obstacles,\\ based on perception},yes
Number\_of\_Actors\_true,integer $\ge 0$,\makecell{Number of actors,\\ based on ground truth},yes
Number\_of\_Actors\_perceived,integer $\ge 0$,\makecell{Number of actual\\ perceived actors,\\ based on perception},yes
Number\_of\_Traffic\_Ctrl\_true,integer $\ge 0$,\makecell{Number of relevant traffic\\ controllers (ground truth)\\ relevant to the current test},yes
{Number\_of\_Traffic\_Ctrl\_\\ perceived},integer $\ge 0$,\makecell{Number of actual perceived \\ traffic controllers \\ relevant to the current test \\ based on perception},yes
\end{filecontents*}
\csvreader[head to column names]{resultsmetadata/resultsmetadata_VUT.csv}{}
		{\\ \hline \textsf{\makecell{\fieldname}} & \unit & \description & \mandatory} 
		\\ \hline
	\end{longtable}
\end{adjustwidth}
\newpage

\begin{adjustwidth}{-2cm}{}
	\setlength\arrayrulewidth{0.01pt}
	\begin{longtable}{|l|c|c|c|}
		\caption{Name and unit of result parameters: Actor states (ground truth)}
		\label{tab:results_csv_format_actors_true}
		\\ \hline			
		\bfseries Field & \bfseries Unit & \bfseries Description & \bfseries Mandatory
		\begin{filecontents*}{resultsmetadata/resultsmetadata_env_actors_true.csv}
fieldname,unit,description,mandatory
Time,s,\makecell{current timestamp\\ (Time=0.000s \\at start of simulation)},yes
Step\_number,integer $>=0$,\makecell{current Simulation step\\ (Step\_number=0 \\at start of simulation,\\monotonically increasing)},yes
Number\_of\_Actors\_true,integer $\ge 0$,\makecell{Number of actors,\\ based on ground truth,\\ in current simulation step},yes
Actor\_Id,alphanumeric,identifier for actor,yes
Actor\_type\_true,\makecell{(pedestrian=0; \acs{pmd}=1;\\ cyclist=2; animal=3;\\ passenger vehicle=4;\\ motorcycle (motorbike)=5;\\ fire truck=6; ambulance=7;\\ van = 8; trailer = 9; \\ truck= 10; bus=11;\\ others=99;)},type of actor,yes
Actor\_pos\_true\_lat,WGS84 decimal,\makecell{true 2D world position\\ latitude of actor},yes
Actor\_pos\_true\_lng,WGS84 decimal,\makecell{true 2D world position\\ longitude of actor},yes
Actor\_heading\_true,$\deg$,\makecell{true world heading angle\\ (measured \textit{clockwise};\\ $0^{\circ} \implies$ north)},yes
Actor\_pos\_true\_x,\makecell{$m$, as per VCS},\makecell{true 2D $x$ position\\ of actor relative from VUT \\ along $X_v$ axis of VUT},no
Actor\_pos\_true\_y,\makecell{$m$, as per VCS},\makecell{true 2D $y$ position\\ of actor relative from VUT \\ along $Y_v$ axis of VUT},no
Actor\_yaw\_true,\makecell{$\deg$, as per VCS},\makecell{true yaw angle of actor \\ between x-axis of actor\\ and $X_v$ axis of VUT\\(measured \textit{clockwise} along\\ the $Z_v$ axis of VUT;\\ $0^{\circ} \implies$ parallel to $X_v$)},no
Actor\_acc\_lat\_true,$m/s^2$,\makecell{true lateral\\ acceleration of actor},yes
Actor\_acc\_lng\_true,$m/s^2$,\makecell{true longitudinal\\ acceleration of actor},yes
Actor\_vel\_lat\_true,$m/s$,\makecell{true lateral\\ velocity of actor},yes
Actor\_vel\_lng\_true,$m/s$,\makecell{true longitudinal\\ velocity of actor},yes
Actor\_vel\_abs\_true,$m/s$,\makecell{true absolute velocity\\ (speed) of actor},yes
Actor\_bpoly\_true,\makecell{\texttt{Array<Position>} \\serialized into a string \\(refer definition \ref{typedef:bpoly_array_position})},\makecell{bounding polygon\\ of actor in WGS84 or VCS\\ based on ground truth},yes
\end{filecontents*}
\csvreader[head to column names]{resultsmetadata/resultsmetadata_env_actors_true.csv}{}
		{\\ \hline \textsf{\makecell{\fieldname}} & \unit & \description & \mandatory} 
		\\ \hline
	\end{longtable}
\end{adjustwidth}

\begin{adjustwidth}{-2cm}{}
	\setlength\arrayrulewidth{0.01pt}
	\begin{longtable}{|l|c|c|c|}
		\caption{Name and unit of result parameters: Actor states (perceived)}
		\label{tab:results_csv_format_actors_perceived}
		\\ \hline			
		\bfseries Field & \bfseries Unit & \bfseries Description & \bfseries Mandatory
		\begin{filecontents*}{resultsmetadata/resultsmetadata_env_actors_perceived.csv}
fieldname,unit,description,mandatory
Time,s,\makecell{current timestamp\\ (Time=0.000s \\at start of simulation)},yes
Step\_number,integer $>=0$,\makecell{current Simulation step\\ (Step\_number=0 \\at start of simulation,\\monotonically increasing)},yes
Number\_of\_Actors\_perceived,integer $\ge 0$,\makecell{Number of actual\\ perceived actors,\\ based on perception,\\ in current simulation step},yes
Actor\_Id,alphanumeric,identifier for actor,yes
Actor\_type\_perceived,\makecell{(pedestrian=0; \acs{pmd}=1;\\ cyclist=2; animal=3;\\ passenger vehicle=4;\\ motorcycle (motorbike)=5;\\ fire truck=6; ambulance=7;\\ van = 8; trailer = 9; \\ truck= 10; bus=11;\\ others=99;)},\makecell{type of actor,\\based on perception},yes
Actor\_pos\_perceived\_lat,WGS84 decimal,\makecell{estimated 2D world position\\ latitude of actor\\ based on perception},yes
Actor\_pos\_perceived\_lng,WGS84 decimal,\makecell{estimated 2D world position\\ longitude of actor\\ based on perception},yes
Actor\_heading\_perceived,$\deg$,\makecell{estimated world heading angle\\ (measured \textit{clockwise};\\ $0^{\circ} \implies$ north)\\ based on perception},yes
Actor\_pos\_perceived\_x,\makecell{$m$, as per VCS},\makecell{estimated 2D $x$ position\\ of actor relative from VUT \\ along $X_v$ axis of VUT,\\ based on perception},no
Actor\_pos\_perceived\_y,\makecell{$m$, as per VCS},\makecell{estimated 2D $y$ position\\ of actor relative from VUT \\ along $Y_v$ axis of VUT,\\ based on perception},no
Actor\_yaw\_perceived,\makecell{$\deg$, as per VCS},\makecell{estimated yaw angle of actor \\ between x-axis of actor\\ and $X_v$ axis of VUT\\(measured \textit{clockwise} along\\ the $Z_v$ axis of VUT;\\ $0^{\circ} \implies$ parallel to $X_v$),\\ based on perception},no
Actor\_bpoly\_perceived,\makecell{\texttt{Array<Position>} \\serialized into a string \\(refer definition \ref{typedef:bpoly_array_position})},\makecell{estimated bounding polygon\\ of actor in WGS84 or VCS,\\ based on perception},yes
Actor\_temporal\_distance,s,\makecell{estimated minimal temporal\\ distance (e.g., \acs{ttc}, if they \\ are on a collision course)\\ between VUT and actor, \\ based on perception},yes
\end{filecontents*}
\csvreader[head to column names]{resultsmetadata/resultsmetadata_env_actors_perceived.csv}{}
		{\\ \hline \textsf{\makecell{\fieldname}} & \unit & \description & \mandatory} 
		\\ \hline
	\end{longtable}
\end{adjustwidth}
\newpage

\begin{adjustwidth}{-2cm}{}
	\setlength\arrayrulewidth{0.01pt}
	\begin{longtable}{|l|c|c|c|}
		\caption{Name and unit of result parameters: Obstacle states (ground truth)}
		\label{tab:results_csv_format_obstacles_true}
		\\ \hline			
		\bfseries Field & \bfseries Unit & \bfseries Description & \bfseries Mandatory
		\begin{filecontents*}{resultsmetadata/resultsmetadata_env_obstacles_true.csv}
fieldname,unit,description,mandatory
Time,s,\makecell{current timestamp\\ (Time=0.000s \\at start of simulation)},yes
Step\_number,integer $>=0$,\makecell{current Simulation step\\ (Step\_number=0 \\at start of simulation,\\monotonically increasing)},yes
Number\_of\_obstacles\_true,integer $\ge 0$,\makecell{Number of actors,\\ based on ground truth,\\ in current simulation step},yes
Obst\_Id,alphanumeric,identifier for obstacle,yes
Obst\_type\_true,\makecell{(construction\_cones=100; \\passable\_object=101;\\ others=199;)},type of obstacle,yes
Obst\_pos\_true\_lat,WGS84 decimal,\makecell{true 2D world position\\ latitude of geometric center \\ of the obstacle},yes
Obst\_pos\_true\_lng,WGS84 decimal,\makecell{true 2D world position\\ longitude of geometric center \\ of the obstacle},yes
Obst\_pos\_true\_x,\makecell{$m$, as per VCS},\makecell{true 2D $x$ position\\ of geometric center of obstacle \\ relative from VUT \\ along $X_v$ axis of VUT},no
Obst\_pos\_true\_y,\makecell{$m$, as per VCS},\makecell{true 2D $y$ position\\ of geometric center of obstacle \\ relative from VUT \\ along $Y_v$ axis of VUT},no
Obst\_bpoly\_true,\makecell{\texttt{Array<Position>} \\serialized into a string \\(refer definition \ref{typedef:bpoly_array_position})},\makecell{bounding polygon\\ of obstacle in WGS84 or VCS\\ based on ground truth},yes
\end{filecontents*}
\csvreader[head to column names]{resultsmetadata/resultsmetadata_env_obstacles_true.csv}{}
		{\\ \hline \textsf{\makecell{\fieldname}} & \unit & \description & \mandatory} 
		\\ \hline
	\end{longtable}
\end{adjustwidth}

\newpage

\begin{adjustwidth}{-2cm}{}
	\setlength\arrayrulewidth{0.01pt}
	\begin{longtable}{|l|c|c|c|}
		\caption{Name and unit of result parameters: Obstacle states (perceived)}
		\label{tab:results_csv_format_obstacles_perceived}
		\\ \hline			
		\bfseries Field & \bfseries Unit & \bfseries Description & \bfseries Mandatory
		\begin{filecontents*}{resultsmetadata/resultsmetadata_env_obstacles_perceived.csv}
fieldname,unit,description,mandatory
Time,s,\makecell{current timestamp\\ (Time=0.000s \\at start of simulation)},yes
Step\_number,integer $>=0$,\makecell{current Simulation step\\ (Step\_number=0 \\at start of simulation,\\monotonically increasing)},yes
Number\_of\_obstacles\_true,integer $\ge 0$,\makecell{Number of actors,\\ based on ground truth,\\ in current simulation step},yes
Obst\_Id,alphanumeric,identifier for obstacle,yes
Obst\_type\_perceived,\makecell{(construction\_cones=100; \\passable\_object=101;\\ others=199;)},\makecell{type of obstacle, \\based on perception},yes
Obst\_pos\_perceived\_lat,WGS84 decimal,\makecell{estimated 2D world position\\ latitude of geometric center \\ of the obstacle,\\ based on perception},yes
Obst\_pos\_perceived\_lng,WGS84 decimal,\makecell{estimated 2D world position\\ longitude of geometric center \\ of the obstacle,\\ based on perception},yes
Obst\_bpoly\_perceived,\makecell{\texttt{Array<Position>} \\serialized into a string \\(refer definition \ref{typedef:bpoly_array_position})},\makecell{estimated bounding polygon\\ of obstacle in WGS84 or VCS,\\ based on perception},yes
Obst\_pos\_perceived\_x,\makecell{$m$, as per VCS},\makecell{true 2D $x$ position\\ of geometric center of obstacle \\ relative from VUT \\ along $X_v$ axis of VUT,\\ based on perception},no
Obst\_pos\_perceived\_y,\makecell{$m$, as per VCS},\makecell{true 2D $y$ position\\ of geometric center of obstacle \\ relative from VUT \\ along $Y_v$ axis of VUT,\\ based on perception},no
Obst\_temporal\_distance,s,\makecell{estimated minimal temporal\\ distance (e.g., \acs{ttc}, if they \\ are on a collision course)\\ between VUT and obstacle, \\ based on perception},yes
\end{filecontents*}
\csvreader[head to column names]{resultsmetadata/resultsmetadata_env_obstacles_perceived.csv}{}
		{\\ \hline \textsf{\makecell{\fieldname}} & \unit & \description & \mandatory} 
		\\ \hline
	\end{longtable}
\end{adjustwidth}

\newpage

\begin{adjustwidth}{-2cm}{}
	\setlength\arrayrulewidth{0.01pt}
	\begin{longtable}{|l|c|c|c|}
		\caption{Name and unit of result parameters: Traffic light controller states (ground truth)}
		\label{tab:results_csv_format_traffic_lights_true}
		\\ \hline			
		\bfseries Field & \bfseries Unit & \bfseries Description & \bfseries Mandatory
		\begin{filecontents*}{resultsmetadata/resultsmetadata_traffic_light_true.csv}
fieldname,unit,description,mandatory
Time,s,\makecell{current timestamp\\ (Time=0.000s \\at start of simulation)},yes
Step\_number,integer $>=0$,\makecell{current Simulation step\\ (Step\_number=0 \\at start of simulation,\\monotonically increasing)},yes
Number\_of\_Traffic\_Ctrl\_true,integer $\ge 0$,\makecell{Number of relevant traffic\\ controllers (ground truth)\\ relevant to the current test},yes
Traffic\_Ctrl\_Id,alphanumeric,\makecell{identifier for the\\ traffic controller)},yes
Traffic\_Ctrl\_Phase\_true,\makecell{(\texttt{go}=0;\\ \texttt{go\_exclusive}=1;\\ \texttt{attention}=2;\\ \texttt{stop}=3;\\ \texttt{blink}=4;\\ \texttt{others}=99;)},\makecell{Current phase of \\ the traffic controller \\ (as per ground truth) \\ \texttt{attention} $\Rightarrow$  amber light \\ \texttt{go} $\Rightarrow$  solid green light \\ \texttt{go\_exclusive} $\Rightarrow$  turn arrow },yes
\end{filecontents*}
\csvreader[head to column names]{resultsmetadata/resultsmetadata_traffic_light_true.csv}{}
		{\\ \hline \textsf{\makecell{\fieldname}} & \unit & \description & \mandatory} 
		\\ \hline
	\end{longtable}
\end{adjustwidth}

\begin{adjustwidth}{-2cm}{}
	\setlength\arrayrulewidth{0.01pt}
	\begin{longtable}{|l|c|c|c|}
		\caption{Name and unit of result parameters: Traffic light controller states (perceived)}
		\label{tab:results_csv_format_traffic_lights_perceived}
		\\ \hline			
		\bfseries Field & \bfseries Unit & \bfseries Description & \bfseries Mandatory
		\begin{filecontents*}{resultsmetadata/resultsmetadata_traffic_light_perceived.csv}
fieldname,unit,description,mandatory
Time,s,\makecell{current timestamp\\ (Time=0.000s \\at start of simulation)},yes
Step\_number,integer $>=0$,\makecell{current Simulation step\\ (Step\_number=0 \\at start of simulation,\\monotonically increasing)},yes
Number\_of\_Traffic\_Ctrl\_\\ perceived,integer $\ge 0$,\makecell{Number of relevant traffic\\ controllers (as perceived)\\ relevant to the current test},yes
Traffic\_Ctrl\_Id,alphanumeric,\makecell{identifier for the\\ traffic controller)},yes
Traffic\_Ctrl\_Phase\_\\ perceived,\makecell{(\texttt{go}=0;\\ \texttt{go\_exclusive}=1;\\ \texttt{attention}=2;\\ \texttt{stop}=3;\\ \texttt{blink}=4;\\ \texttt{others}=99;)},\makecell{Current phase of \\ the traffic controller \\ (as perceived) \\ \texttt{attention} $\Rightarrow$  amber light \\ \texttt{go} $\Rightarrow$  solid green light \\ \texttt{go\_exclusive} $\Rightarrow$  turn arrow },yes
\end{filecontents*}
\csvreader[head to column names]{resultsmetadata/resultsmetadata_traffic_light_perceived.csv}{}
		{\\ \hline \textsf{\makecell{\fieldname}} & \unit & \description & \mandatory} 
		\\ \hline
	\end{longtable}
\end{adjustwidth}


\clearpage
\newpage

\end{document}